\documentclass[10pt]{article} 
\usepackage[accepted]{tmlr}
\usepackage{amssymb}

\usepackage[noend]{algorithm2e}

\newcommand\blfootnote[1]{%
  \begin{NoHyper}%
  \renewcommand\thefootnote{}\footnote{#1}%
  \addtocounter{footnote}{-1}%
  \end{NoHyper}
}
\SetCommentSty{mycommfont}
\RestyleAlgo{ruled}
\usepackage[pdftex]{graphicx} 
\usepackage[colorlinks=true, allcolors=blue]{hyperref}     
\usepackage{url}            
\usepackage{wrapfig}
\usepackage{amsmath}
\usepackage{amssymb}
\usepackage{caption} 
\usepackage{makecell}
\usepackage{float}
\usepackage{subcaption}
\usepackage{booktabs}
\usepackage[noabbrev,capitalize]{cleveref}
\usepackage{multirow}
\usepackage{array}
\usepackage[inline,shortlabels]{enumitem}
\usepackage{etoolbox}

\makeatletter
\patchcmd{\@algocf@start}
  {-1.5em}
  {0pt}
  {}{}
\makeatother

\usepackage{siunitx}
\newcommand{\eg}{e.g.\@\xspace}

\newcommand{\ie}{i.e.\@\xspace}

\newcommand{\wrt}{w.r.t.\@\xspace}

\newcommand{\cf}{cf.\@\xspace}

\definecolor{rebuttal_color}{rgb}{0.5, 0.0, 0.5}

\makeatletter
\g@addto@macro{\endtabular}{\rowfont{}}
\makeatother
\newcommand{\rowfonttype}{}
\newcommand{\rowfont}[1]{
\gdef\rowfonttype{#1}#1\ignorespaces%
}
\makeatother

\title{Time Series Continuous Modeling for Imputation and Forecasting with Implicit Neural Representations}

\author{\name Etienne Le Naour\textsuperscript{*} \email etienne.le-naour@edf.fr \\
      \addr EDF R\&D, Palaiseau, France\\
      Sorbonne Université, CNRS, ISIR, 75005 Paris, France
      \AND
      \name Louis Serrano\textsuperscript{*} \email louis.serrano@sorbonne-universite.fr \\
      \addr Sorbonne Université, CNRS, ISIR, 75005 Paris, France
      \AND
      \name Léon Migus\textsuperscript{*} \email leon.migus@sorbonne-universite.fr\\
      \addr Sorbonne Université, CNRS, ISIR, 75005 Paris, France\\
      Sorbonne Université, CNRS, Laboratoire Jacques-Louis Lions, 75005 Paris, France
      \AND
      \name Yuan Yin \email yuan.yin@sorbonne-universite.fr \\
      \addr Sorbonne Université, CNRS, ISIR, 75005 Paris, France
      \AND
      \name Ghislain Agoua \email ghislain.agoua@edf.fr \\
      \addr EDF R\&D, Palaiseau, France\\
      \AND
      \name Nicolas Baskiotis \email nicolas.baskiotis@sorbonne-universite.fr \\
      \addr Sorbonne Université, CNRS, ISIR, 75005 Paris, France
      \AND
      \name Patrick Gallinari \email patrick.gallinari@sorbonne-universite.fr \\
      \addr Sorbonne Université, CNRS, ISIR, 75005 Paris, France \\
      Criteo AI Lab, Paris, France
      \AND
      \name Vincent Guigue \email vincent.guigue@sorbonne-universite.fr \\
      \addr AgroParisTech, Palaiseau, France\\
      }

\def\month{04}  
\def\year{2024} 
\def\openreview{\url{https://openreview.net/forum?id=P1vzXDklar}} 

\begin{document}

\maketitle
\blfootnote{* Equal contribution}

\begin{abstract}
We introduce a novel modeling approach for time series imputation and forecasting, tailored to address the challenges often encountered in real-world data, such as irregular samples, missing data, or unaligned measurements from multiple sensors. Our method relies on a continuous-time-dependent model of the series' evolution dynamics. It leverages adaptations of conditional, implicit neural representations for sequential data. A modulation mechanism, driven by a meta-learning algorithm, allows adaptation to unseen samples and extrapolation beyond observed time-windows for long-term predictions. The model provides a highly flexible and unified framework for imputation and forecasting tasks across a wide range of challenging scenarios. It achieves state-of-the-art performance on classical benchmarks and outperforms alternative time-continuous models.
\end{abstract}

\section{Introduction}
Time series analysis and modeling are ubiquitous in a wide range of fields, including industry, medicine, and climate science. The variety, heterogeneity and increasing number of deployed sensors, raise new challenges when dealing with real-world problems for which current methods often fail. For example, data are frequently irregularly sampled, contain missing values, or are unaligned when collected from distributed sensors \citep{schulz1997spectrum, clark2004population}.
Recent advancements in deep learning have significantly improved state-of-the-art performance in both data imputation \citep{cao2018brits,du2023saits} and forecasting tasks \citep{Dlinear,PatchTST}. Many state-of-the-art models, such as transformers, have been primarily designed for dense and regular grids \citep{AutoFormer,PatchTST,du2023saits}. They struggle to handle irregular data and often suffer from significant performance degradation \citep{chen2001study,kim2019analysis}.

Our objective is to explore alternatives to state-of-the-art (SOTA) transformers able to handle, in a unified framework, imputation and forecasting tasks for irregularly, arbitrarily sampled, and unaligned time series sources. Time-dependent continuous models \citep{RasmussenW06,GarneloRMRSSTRE18,corr/abs-1907-03907} offer such an alternative. However, until now, their performance has lagged significantly behind that of models designed for regular discrete grids.
A few years ago, implicit neural representations (INRs) emerged as a powerful tool for representing images as continuous functions of spatial coordinates \citep{SIREN, tancik2020fourier} with recent new applications such as image generation \citep{Functa} or even  modeling dynamical systems \citep{DiNO}.

In this work, we leverage the potential of conditional INR models within a meta-learning approach to introduce TimeFlow: a unified framework designed for modeling continuous time series and addressing imputation and forecasting tasks with irregular and unaligned observations. Our key contributions are: 

\begin{itemize}
    \item We propose a novel framework that excels in modeling time series as continuous functions of time, accepting arbitrary time step inputs, thus enabling the handling of irregular and unaligned time series for both imputation and forecasting tasks. 
    This is one of the very first attempts to adapt INRs that enables efficient handling of both imputation and forecasting tasks within a unified framework. The methodology which leverages the synergy between the model components, evidenced in the context of this application, is a pioneering contribution to the field.
    
    \item We conducted an extensive comparison with state-of-the-art continuous and discrete models. It demonstrates that our approach outperforms continuous and discrete SOTA deep learning approaches for imputation. As for long-term forecasting, it outperforms existing continuous models both on regular and irregular samples. It is on par with SOTA discrete models on regularly sampled time series while allowing for a much greater flexibility for irregular samplings, allowing to cope with situations where discrete models fail. Furthermore, we prove that our method effortlessly handles previously unseen time series and new time windows, making it well-suited for real-world applications.
\end{itemize}

\section{Related work}
\label{RelatedContent}

\paragraph{Discrete methods for time series imputation and forecasting.}
\looseness=-1
Recently, Deep Learning (DL) methods have been widely used for both time series imputation and forecasting. For imputation, BRITS \citep{cao2018brits} uses a bidirectional recurrent neural network (RNN). Alternative frameworks were later explored, \eg, GAN-based \citep{luo2018multivariate,luo2019e2gan,liu2019naomi}, VAE-based \citep{fortuin2020gp}, diffusion-based \citep{tashiro2021csdi}, matrix factorization-based (TIDER, \citealp{TIDER}) and transformer-based (SAITS, \citealp{du2023saits}) approaches. These methods cannot handle irregular time series. In situations involving multiple sensors, such as those placed at different locations, incorporating new sensors necessitates retraining the entire model, thereby limiting their usability.
 For forecasting, most recent DL SOTA models are based on transformers. Initial approaches apply plain transformers directly to the series, each token being a series element  \citep{Informer,PyraFormer,AutoFormer,FedFormer}. These transformers may underperform linear models as shown in \citep{Dlinear}. 
PatchTST \citep{PatchTST} significantly improved transformers SOTA performance by considering sub-series as tokens of the series. However, all these models cannot handle properly irregularly sampled look-back windows. 

\paragraph{Continuous methods for time series.} 
Gaussian Processes \citep{RasmussenW06} have been a popular family of methods for modeling time series as continuous functions. They require choosing an appropriate kernel \citep{GPpriorneeded} and may suffer limitations in large dimensions settings. Neural Processes (NPs) \citep{GarneloRMRSSTRE18, kim2019analysis} parameterize Gaussian processes through an encoder-decoder architecture leading to more computationally efficient implementations. NPs have been used to model simple signals for imputation and forecasting tasks, but struggle with more complex signals. \cite{BilosRSNG23} parameterizes a Gaussian Process through a diffusion model, but the model has difficulty adapting to a large number of timestamps.  
Other approaches such as \cite{BrouwerSAM19} and \cite{corr/abs-1907-03907} model time series continuously with latent ordinary differential equations. mTAN \citep{ShuklaM21}, a transformer model, uses an attention mechanism to impute irregular time series.
While these approaches have shown significant progress in continuous modeling for time series, we observed that their performances on imputation and forecasting tasks are inferior compared to the aforementioned discrete models (\cref{tab:impu_classic}, \cref{tab:forecast_time_shift_electricity}).

\paragraph{Implicit neural representations.} The recent development of implicit neural representations (INRs) has led to impressive results in computer vision \citep{SIREN,tancik2020fourier,MFN,mildenhall2021nerf}. INRs can represent data as a continuous function, which can be queried at any coordinate. 
While they have been applied in other fields such as physics \citep{DiNO} and meteorology \citep{Huang2023}, there has been limited research on INRs for time series analysis. Prior works \citep{HyperTime,INRAD} focused on time series generation for data augmentation and on time series encoding for reconstruction but are limited by their fixed grid input requirement. DeepTime \citep{DeepTime} is the closest work to our contribution. DeepTime learns a set of basis INR functions from a training set of multiple time series and combines them using a Ridge regressor. This regressor allows it to adapt to new time series. It has been designed for forecasting only. The original version cannot handle imputation properly and was adapted to do so for our comparisons. In our experiments, we will demonstrate that TimeFlow significantly outperforms DeepTime in imputation and also in forecasting tasks when dealing with missing values in the look-back window. TimeFlow also shows a slight advantage over DeepTime in forecasting regularly sampled series.

\section{The TimeFlow framework}
\label{sec:method}

\subsection{Problem setting}
\looseness=-1
We aim to develop a unified framework for time series imputation and forecasting that reduces dependency on a fixed sampling scheme for time series. We introduce the following notations for both tasks. During training, in the imputation setting, we have access to time series in an observed temporal grid denoted as $\mathcal{T}_{in}$, which is a subset of the dense temporal support $\mathcal{T}$. In the forecasting setting, we observe time series within a limited past time grid, referred to as the 'look-back window' and denoted as $\mathcal{T}_{in}$ (a subset of $\mathcal{T}$), as well as a future grid, the 'horizon', denoted as $\mathcal{T}_{out}$ (also a subset of $\mathcal{T}$). At test time, in both cases, and given observed values in a temporal grid $\mathcal{T}_{in}^{*}$ included in a possibly new temporal window $\mathcal{T}^{*}$, our objective is to infer the time series values within $\mathcal{T}^{*}$. $\mathcal{T}^{*} = \mathcal{T}$ if we infer values in the training temporal support (e.g. in the classical imputation scenario, see \cref{sec:imputation}), $\mathcal{T}^{*} \neq \mathcal{T}$ if we infer for a new temporal support(e.g. in the forecasting setting, see \cref{sec:expe_forecasting}). 

\subsection{Key components}

Our framework is articulated around three key components:
\begin{enumerate}[(i)] 
\item \textbf{INR-based time-continuous functions}: a discrete time series $x = (x_{t_1}, x_{t_2}, \ldots, x_{t_k})$ can be represented  by an underlying time-continuous function $\textbf{x} \colon t \in \mathbb{R}_+ \to x_t \in \mathbb{R}^d$ (in our experiments $d=1$). We want to approximate  the ground-truth $\textbf{x}$ by employing implicit neural representations (INRs), which are neural networks capable of learning a parameterized continuous function $f_\theta$ from discrete data by minimizing the reconstruction loss between observed data and network's outputs.

\item \textbf{Conditional INRs with modulations}: An INR can represent only one function, whether it's an image or a time series. To effectively represent a collection of time series $(x^{(j)})_j$ using INRs, we improve their encoding by incorporating per-sample modulations, which we denote as $\psi^{(j)}$.  These modulations condition the parameters $\theta$ of the INRs. We use the notation $f_{\theta, \psi^{(j)}}$ to refer to the conditioned INR with the modulations $\psi^{(j)}$.

\item \textbf{Optimization-based encoding}: the conditioning modulation parameters $\psi^{(j)}$ are calculated as a function of codes $z^{(j)}$ that represent the individual sample series. We acquire these codes $z^{(j)}$ through a meta-learning optimization process using an auto-decoding strategy. Notably, auto-decoding has been found to be more efficient for this purpose than set encoders \citep{kim2019analysis}.
\end{enumerate}
In the following sections, we will elaborate on each component of our method. Given that the choices made for each component and the methodology developed to enhance their synergy are essential aspects, we provide a discussion of the various choices involved in \cref{discussion}.

\paragraph{INR-based time-continuous functions.}
We implement our INR with Fourier features and a feed-forward network (FFN) with ReLU activations, \ie for a time coordinate $t\in \mathcal{T}$, the output of the INR $f_\theta$ is given by $f_\theta(t) = \text{FFN}(\gamma(t))$. The Fourier Features $\gamma(\cdot)$ are a frequency embedding of the time coordinates used to capture high-frequencies  \citep{tancik2020fourier, mildenhall2021nerf}. In our case, we chose $\gamma(t) := (\sin(\pi t), \cos(\pi t), \cdots, \sin(2^{N-1}\pi t), \cos(2^{N-1}\pi t))$, with $N$ the number of fixed frequencies. For an INR with $L$ layers, the output is computed as follows: \begin{enumerate*}[(i)] \item we get the frequency embedding $\phi_0 = \gamma(t)$, \item we update the hidden states according to $\phi_{l} = \text{ReLU}(\theta_{l} \phi_{l-1} + b_{l})$ for $l = 1, \ldots, L$, \item we project onto the output space $f_\theta(t) = \theta_{L+1} \phi_{L} + b_{L+1}$.
\end{enumerate*}

 \begin{wrapfigure}[21]{R}{0.55\linewidth}
\vspace{-1.2em}
    \centering
    \includegraphics[width=0.99\linewidth]{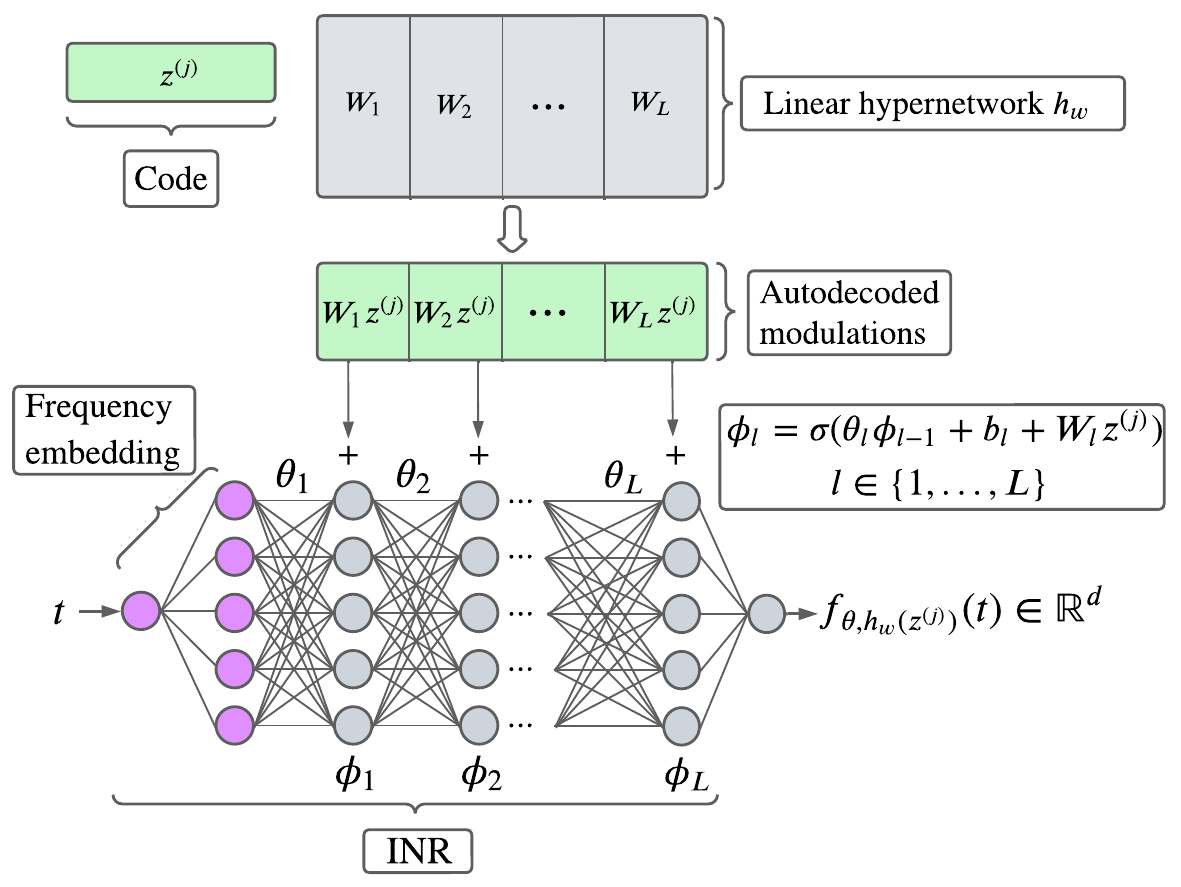}
    \caption{Overview of TimeFlow architecture. Forward pass to approximate the time series $x^{(j)}$. $\sigma$ stands for the ReLU activation function.}
    \label{fig:INR-scheme}
\end{wrapfigure}

\paragraph{Conditional INRs with modulations.} As indicated, sample conditioning of the INR is performed through modulations of its parameters. In order to adapt rapidly the model to new samples, the conditioning should rely only on a small number of the INR parameters. This is achieved by modifying only the biases of the INR through the introduction of  an additional bias term $\psi_l^{(j)}$ for each layer $l$, also known as \textit{shift modulation}. To further limit the versatility of the conditioning, we generate the instance modulations $\psi^{(j)}$ from compact codes $z^{(j)}$ through a linear hypernetwork $h$ with parameters $w$, \ie,  $\psi^{(j)} = h_w(z^{(j)})$. Consequently, the approximation of a time series ${x}^{(j)}$, denoted globally as $f_{\theta, h_w(z^{(j)})}$, will depend on shared parameters $\theta$ and $w$ that are common among all the INRs involved in modeling the series family and on the code $z^{(j)}$ specific to series $x^{(j)}$. The output of the $l$-th layer of the modulated INR is given by $\phi_{l} = \text{ReLU}(\theta_l \phi_{l-1} + b_l + \psi_l^{(j)})$, where $\psi_l^{(j)} = W_l z^{(j)}$, and $w:=(W_l)_{l=1}^{L}$ are the parameters of the hypernetwork $h_w$. This design enables gathering information across samples into the common parameters of the INR and hypernetwork, while the codes contain only specific information about their respective time-series samples. The architecture is illustrated in \cref{fig:INR-scheme}.

\paragraph{Optimization-based encoding.} 
We condition the INR using the data from $\mathcal{T}_{in}$, and learn the shared INR and hypernetwork parameters $\theta$ and $w$ using $\mathcal{T}_{in}$ for both imputation and forecasting, and $\mathcal{T}_{out}$ for forecasting only. We achieve the conditioning on $\mathcal{T}_{in}$ by optimizing the codes $z^{(j)}$ through gradient descent. The joint optimization of the codes and common parameters is challenging.
In TimeFlow, it is achieved through a meta-learning approach, adapted from \cite{Functa} and \cite{CAVIA}.  The objective is to learn shared parameters so that the code $z^{(j)}$ can be adapted in just a few gradient steps for a new series $x^{(j)}$. For training, we  perform parameter optimization at two levels: the inner-loop and the outer-loop. The inner-loop adapts the code $z^{(j)}$ to condition the network on the set $\mathcal{T}_{in}^{(j)}$, while the outer-loop updates the common parameters using $\mathcal{T}_{in}^{(j)}$ and also $\mathcal{T}_{out}^{(j)}$ for forecasting (see \cref{intuition_metalearning} for more detailed intuition.). We present our training optimization in \Cref{alg:TimeFlow}. At each training epoch and for each batch of data $\mathcal{B}$ composed of time series $x^{(j)}$ sampled from the training set, we first update individually the codes $z^{(j)}$ in the inner loop, before updating the common parameters in the outer loop using a loss over the whole batch. We introduce a parameter $\lambda$ to weight the importance of the loss over $\mathcal{T}_{out}$ \wrt the loss over $\mathcal{T}_{in}$ for the outer-loop. In practice, when $\mathcal{T}_{out}$ exists, \ie for forecasting, we set $\lambda=1$ and $\lambda=0$ otherwise. We use an MSE loss over the observations grid $\mathcal{L}_{\mathcal{T}}(x_t, \tilde{x_t}) := \mathbb{E}_{t \sim \mathcal{T}}[(x_t - \tilde{x_t})^2]$. We denote $\alpha$ and $\eta$ the learning rates of the inner-loop and outer-loop. Using $K=3$ steps for training and testing is sufficient for our experiments thanks to the use of second-order meta-learning as explained in \cref{discussion}.

\begin{algorithm}[H]
\caption{TimeFlow Training}\label{alg:TimeFlow}
\SetKwComment{Comment}{/* }{ */}
\While{no convergence}{
    Sample batch $\mathcal{B}$ of data $({x}^{(j)})_{j\in  \mathcal{B}}$\;
    Set codes to zero $z^{(j)} \gets 0, \forall j \in \mathcal{B}$ \; 
    \tcp*[h]{inner loop for encoding:} \\
    \For{$j \in \mathcal{B}$ and step $\in \{1, ..., K\}$}{
        $z^{(j)}\!\gets z^{(j)} - \alpha \nabla_{z^{(j)}} \mathcal{L}_{\mathcal{T}_{in}^{(j)}}(f_{\theta, h_{w}(z^{(j)})}, x^{(j)})$\; 
    }
    \tcp*[h]{outer loop step:} \\
    $[\theta, w] \gets [\theta, w] - \eta \nabla_{[\theta, w]} \frac{1}{|\mathcal{B}|} \sum_{j \in \mathcal{B}} [\mathcal{L}_{\mathcal{T}_{in}^{(j)}}(f_{\theta, h_{w}(z^{(j)})}, x^{(j)}) + \lambda \mathcal{L}_{\mathcal{T}_{out}^{(j)}}(f_{\theta, h_{w}(z^{(j)})}, x^{(j)}) ]$ \;
}
\end{algorithm}

\subsection{TimeFlow inference}
\looseness=-1
During the inference process, we aim to infer the time series value for each timestamp in the dense grid $\mathcal{T}^{* (j)}$ based on the partial observation grid $\mathcal{T}^{* (j)}_{in} \subset \mathcal{T}^{* (j)}$. We can encounter two scenarios: \begin{enumerate*}[(i)] \item One where we observe the same time window as during training ($\mathcal{T}^{* (j)} = \mathcal{T}^{(j)}$) as in the imputation setting in \cref{sec:imputation}. \item One, where we are dealing with a newly observed time window ($\mathcal{T}^{* (j)} \neq \mathcal{T}^{(j)}$), as in the forecasting setting in \cref{sec:expe_forecasting}. \end{enumerate*} At inference, the parameters $\theta$ and $w$ are kept fixed to their final training values. We optimize the individual parameters $z^{* (j)}$ based on the newly observed grid $\mathcal{T}^{* (j)}_{in}$ using the $K$ inner-steps of the meta-learning algorithm as described in \cref{alg:TimeFlow_inference}. We are then in position to query $f_{\theta, h_{w}(z^{* (j)})}(t)$ for any given timestamp $t \in \mathcal{T}^{* (j)}$.

\begin{algorithm}[H]
\caption{TimeFlow Inference with trained $\theta,w$ }\label{alg:TimeFlow_inference}
\SetKwComment{Comment}{/* }{ */}
    For the $j$-th series $({x}^{(j)})$, set code to zero $z^{* (j)} \gets 0$\;
    \For{step $\in \{1, ..., K\}$}{
        $z^{* (j)} \gets z^{* (j)} - \alpha \nabla_{z^{* (j)}} \mathcal{L}_{\mathcal{T}^{{*}(j)}_{in}}(f_{\theta, h_{w}(z^{* (j)})}, x_t)$
    }

    Query $f_{\theta, h_{w}(z^{* (j)})}(t)$ for any $t \in \mathcal{T}^{{*}(j)}$
\end{algorithm}

\subsection{Discussion on implementation choices} 
\label{discussion}

As indicated before, adapting the components and enhancing their synergy for the tasks of imputation and forecasting is not trivial and requires careful choices. We conducted several ablation studies to provide a comprehensive examination of key implementation choices of our framework. 

Our findings can be summarized as follows:
\begin{itemize}
    \item \textbf{Choice of INR}: An FFN with Fourier Features outperformed other popular INRs for the tasks considered in this study. Unlike SIREN \citep{SIREN}, which does not explicitly incorporate frequencies but uses sine activation functions, the Fourier features network can more effectively capture a wider range of frequencies, especially at low sampling rates. This is crucial for accurately capturing high frequencies in sparsely observed time series. Our experiments, detailed in Section \ref{section:fourier-vs-siren} and Table \ref{tab:sirenVSfourier}, demonstrate this superiority across various datasets.
    \item \textbf{Choice of encoding / meta-learning}: TimeFlow with a set encoder for learning the compact conditioning codes $z$ in place of the auto-decoding strategy used here, proved much less effective on complex datasets. This is further elaborated in Section \ref{section:timeflow-other-metalearning} and Table \ref{tab:forecast_abla_set_encoder}. Additionally, replacing the 2nd-order meta-learning optimization for a 1st-order one, such as REPTILE \citep{nichol2018first}, led to unstable training, as shown in Table \ref{tab:impu_abla_first_order_meta_learning}.
    \item \textbf{Choice of modulations}:  Complexifying the modulation by introducing scaling parameters in addition to shift parameters did not provide performance gains. Our experiments on the \textit{Electricity} dataset, detailed in Section \ref{section:modulation-choice} and Table \ref{tab:ablation_modulations}, indicate that shift-only modulation is more efficient.
\end{itemize}

For TimeFlow, across all experiments, we used a code dimension of 128, an FFN with a depth of 5 and a width of 256, and 64 Fourier features. We used 3 inner steps and a learning rate of 0.01 for the inner-loop, and a learning rate of $5 \times 10^{-4}$ for the outer-loop. We performed a comprehensive analysis to understand notably the \textbf{influence of the $z$ dimension}: a latent code dimension of 128 was suitable for our tasks; this is supported by results in Section \ref{section:dim-z} and Table \ref{tab:inference_dim_z} - and the \textbf{influence of the number of inner steps}: using 3 inner steps for training and inference struck a favorable balance between reconstruction capabilities and computational efficiency, as detailed in Section \ref{section:gradient_steps}.

\section{Experiments}
We conducted a comprehensive evaluation of our TimeFlow framework across three different tasks, comparing its performance to state-of-the-art continuous and discrete baseline methods. In \Cref{sec:imputation}, we assess TimeFlow's capabilities to impute sparsely observed time series under various sampling rates. \Cref{sec:expe_forecasting} focuses on long-term forecasting, where we evaluate TimeFlow over standard long-term forecasting horizons. 
In \Cref{sec:impute_forecast}, we tackle a challenging task forecasting with incomplete look-back windows, thus combining the challenges of imputation and forecasting. The code for the experiments is available at \href{https://github.com/EtienneLnr/TimeFlow.git}{this link}.

\paragraph{Datasets.}{
We tested our framework on three extensive multivariate datasets where a single phenomenon is measured at multiple locations over time, namely \textit{Electricity}, \textit{Traffic} and \textit{Solar}.
 The \textit{Electricity} dataset comprises hourly electricity load curves of 321 customers in Portugal, spanning the years 2012 to 2014. The \textit{Traffic} dataset is composed of hourly road occupancy rates from 862 locations in San Francisco during 2015 and 2016. Lastly, the \textit{Solar} dataset contains measurements of solar power production from 137 photovoltaic plants in Alabama, recorded at 10-minute intervals in 2006. Additionally, we have created an hourly version, \textit{SolarH}, for the sake of consistency in the forecasting section.
These datasets exhibit diversity in various characteristics: 
\begin{itemize*}
    \item They exhibit diverse temporal frequencies, including daily and weekly seasonality observed in the \textit{Traffic} and \textit{Electricity} datasets, while the \textit{Solar} dataset possesses only daily frequency.
    \item There is individual variability across data samples and more pronounced trends in the \textit{Electricity} dataset compared to the \textit{Traffic} and \textit{Solar} datasets.
\end{itemize*}}

\subsection{Imputation}
\label{sec:imputation} 
We consider the classical imputation setting where $n$ time series are partially observed over a given time window. Using our approach, we can predict for each time series the value at any timestamp $t$ in that time window based on partial observations.

\begin{figure*}[!htb]
    \centering
    \includegraphics[width=0.95\linewidth]{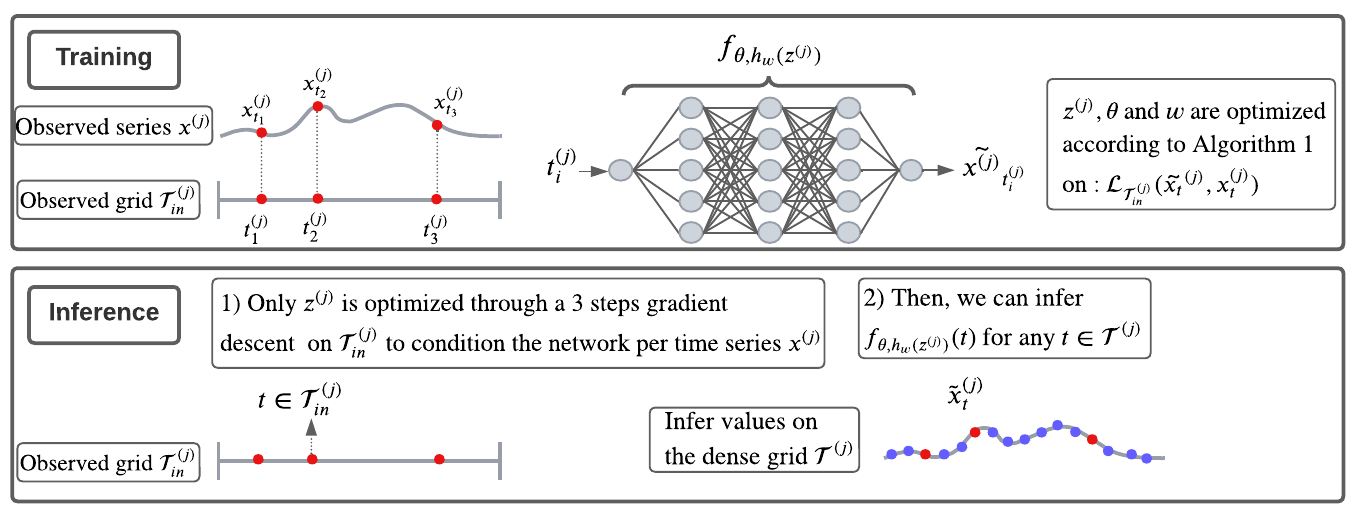}
    \caption[foo bar]{Training and inference procedures of TimeFlow for imputation. \begin{enumerate*}[(i)] \item During training, for each time series $x^{(j)}$, our observations (red dots \textcolor{red}{\scalebox{1.1}{$\bullet$}}) are restricted to the sparsely sampled grid, denoted as $\mathcal{T}^{(j)}_{in}$. \item During inference, our objective is to infer the values over the dense grids $\mathcal{T}^{(j)}$, on the unobserved data points (such as the blue dots \textcolor{blue}{\scalebox{1.1}{$\bullet$}} on the figure).\end{enumerate*}}
    \label{fig:imputation_procedure}
\end{figure*}

\looseness=-1
\paragraph{Setting.} For a time series $x^{(j)}$, we denote the set of observed points as $\mathcal{T}_{in}^{{(j)}}$ and the ground truth set of points as $\mathcal{T}^{{(j)}}$. The observed time grids may be irregularly spaced and may differ across the different time series ($\mathcal{T}_{in}^{{(j_1)}} \neq \mathcal{T}_{in}^{{(j_2)}}, \forall j_1\neq j_2$). The model is trained for each $x^{(j)}$ following \cref{alg:TimeFlow}. Then, we aim to infer for any unobserved $t \in \mathcal{T}^{{(j)}}$ the missing value $x_t^{(j)}$ conditioned on $\mathcal{T}^{(j)}_{in}$ according to \cref{alg:TimeFlow_inference}.
For this imputation task, the TimeFlow training and inference procedures are detailed in \cref{sec:method} and illustrated in \cref{fig:imputation_procedure}.
For comparison with the SOTA imputation baselines, we assume that the ground truth time grid is the same for each sample. The subsampling rate $\tau$ is define as the rate of observed values. 

\paragraph{Baselines.} We compare TimeFlow with various baselines, including discrete imputation methods, such as CSDI \citep{tashiro2021csdi}, SAITS \citep{du2023saits}, BRITS \citep{cao2018brits}, and TIDER \citep{TIDER}, and continuous ones, such as Neural Process (NP, \citealp{GarneloRMRSSTRE18}), mTAN \citep{ShuklaM21}, and DeepTime with slight adjustments \citep{DeepTime} (details \cf \cref{sec:deeptime-adaptation}). See \cref{baseline-training-sec} for the baseline training procedure and hyperparameter selection. For each dataset, we divide the series into five independent time windows (consisting of 2000 timestamps for \textit{Electricity} and \textit{Traffic}, and 10,000 timestamps for \textit{Solar}), perform imputation on each time window and average the performance to obtain robust results. We evaluate the quality of the models for different subsampling rates, from the easiest $\tau=0.5$ to the most difficult $\tau=0.05$. All the scores presented in the experiments are reported as Mean Absolute Error (MAE).

\begin{table}[!t]
\setlength{\tabcolsep}{4pt}
\caption{Mean MAE imputation results on the missing grid only. Each time series is divided into 5 time windows onto which imputation is performed, and the performances are averaged over the 5 windows. In the table, $\tau$ stands for the subsampling rate, i.e. the proportion of observed points considered for each time window. Bold results are best, underlined results are second best. TimeFlow improvement represents the overall percentage improvement achieved by TimeFlow compared to the specific method being considered.}
\label{tab:impu_classic}
\begin{center}
\resizebox{\textwidth}{!}{%
\begin{tabular}{ccccccccccc}
\toprule

  &        & \multicolumn{4}{c}{Continuous methods} & \multicolumn{4}{c}{Discrete methods} \\
  
  \cmidrule(r){3-6} \cmidrule(r){7-10} 

  & $\tau$ &  TimeFlow & DeepTime & mTAN & Neural Process & CSDI & SAITS & BRITS & TIDER\\
\midrule

        & 0.05 &  \textbf{0.324 $\pm$ 0.013} & 0.379 $\pm$ 0.037 & 0.575 $\pm$ 0.039 & 0.357 $\pm$ 0.015 & 0.462 $\pm$ 0.021 & 0.384 $\pm$ 0.019 & \underline{0.329 $\pm$ 0.015} & 0.427 $\pm$ 0.010  \\

        & 0.10 &  \textbf{0.250 $\pm$ 0.010} & 0.333 $\pm$ 0.034 & 0.412 $\pm$ 0.047 & 0.417 $\pm$ 0.057 & 0.398 $\pm$ 0.072  & 0.308 $\pm$ 0.011 & \underline{0.287 $\pm$ 0.015} & 0.399 $\pm$ 0.009\\

        Electricity & 0.20 &  \textbf{0.225 $\pm$ 0.008}  & \underline{0.244 $\pm$ 0.013} & 0.342 $\pm$ 0.014  & 0.320 $\pm$ 0.017 & 0.341 $\pm$ 0.068 & 0.261 $\pm$ 0.008 &  0.245 $\pm$ 0.011 & 0.391 $\pm$ 0.010\\

        & 0.30 &  \textbf{0.212 $\pm$ 0.007}  & 0.240 $\pm$ 0.014  &  0.335 $\pm$ 0.015  & 0.300 $\pm$ 0.022 & 0.277 $\pm$ 0.059 & 0.236 $\pm$ 0.008  &  \underline{0.221 $\pm$ 0.008} & 0.384 $\pm$ 0.009\\

        & 0.50 &  0.194 $\pm$ 0.007  & 0.227 $\pm$ 0.012 & 0.340 $\pm$ 0.022 & 0.297 $\pm$ 0.016 & \textbf{0.168 $\pm$ 0.003}  & 0.209  $\pm$ 0.008 &  \underline{0.193 $\pm$ 0.008} & 0.386 $\pm$ 0.009 \\ 
        
        \midrule

        & 0.05 &  \textbf{0.095 $\pm$ 0.015}  & 0.190 $\pm$ 0.020  & 0.241 $\pm$ 0.102 & \underline{0.115 $\pm$ 0.015} & 0.374 $\pm$ 0.033 & 0.142 $\pm$ 0.016    &  0.165 $\pm$ 0.014 & 0.291 $\pm$ 0.009 \\

        & 0.10 &  \textbf{0.083 $\pm$ 0.015}  & 0.159 $\pm$ 0.013 & 0.251 $\pm$ 0.081 &  \underline{0.114 $\pm$ 0.014} & 0.375 $\pm$ 0.038 & 0.124  $\pm$ 0.018  &  0.132 $\pm$ 0.015 & 0.276 $\pm$ 0.010 \\

        Solar & 0.20 &   \textbf{0.072 $\pm$ 0.015}  & 0.149 $\pm$ 0.020 & 0.314 $\pm$ 0.035 & 0.109 $\pm$ 0.016 & 0.217 $\pm$ 0.023 &   \underline{0.108  $\pm$ 0.014} &  0.109 $\pm$ 0.012 & 0.270 $\pm$ 0.010\\

        & 0.30 &   \textbf{0.061 $\pm$ 0.012}   & 0.135 $\pm$ 0.014  & 0.338 $\pm$ 0.05 & 0.108 $\pm$ 0.016 & 0.156 $\pm$ 0.002  & 0.100 $\pm$ 0.015   &  \underline{0.098 $\pm$ 0.012} & 0.266 $\pm$ 0.010\\

        & 0.50 &   \textbf{0.054 $\pm$ 0.013}  & 0.098 $\pm$ 0.013 & 0.315 $\pm$ 0.080 & 0.107 $\pm$ 0.015 & \underline{0.079 $\pm$ 0.011} & 0.094 $\pm$ 0.013   &  0.088 $\pm$ 0.013 & 0.262 $\pm$ 0.009\\

        \midrule

        & 0.05 &  0.283 $\pm$ 0.016   & \textbf{0.246 $\pm$ 0.010} & 0.406 $\pm$ 0.074  & 0.318 $\pm$ 0.014 & 0.337 $\pm$ 0.045 & 0.293 $\pm$ 0.007   & \underline{0.261 $\pm$ 0.010} & 0.363 $\pm$ 0.007 \\

        & 0.10 &  \textbf{0.211 $\pm$ 0.012}   & \underline{0.214 $\pm$ 0.007} & 0.319 $\pm$ 0.025 & 0.288 $\pm$ 0.018 & 0.288 $\pm$ 0.017 & 0.237 $\pm$ 0.006  &  0.245 $\pm$ 0.009 & 0.362 $\pm$ 0.006 \\

        Traffic & 0.20 &  \textbf{0.168 $\pm$ 0.006} & 0.216 $\pm$ 0.006 & 0.270 $\pm$ 0.012 &  0.271 $\pm$ 0.011 & 0.269 $\pm$ 0.017  &  \underline{0.197 $\pm$ 0.005}  & 0.224 $\pm$ 0.008 & 0.361 $\pm$ 0.006\\

        & 0.30 &  \textbf{0.151 $\pm$ 0.007}   & \underline{0.172 $\pm$ 0.008} & 0.251 $\pm$ 0.006  & 0.259 $\pm$ 0.012 & 0.240 $\pm$ 0.037 & 0.180  $\pm$ 0.006   & 0.197 $\pm$ 0.007 &  0.355 $\pm$ 0.006 
        \\

        & 0.50 &  \textbf{0.139 $\pm$ 0.007} & 0.171 $\pm$ 0.005 & 0.278 $\pm$ 0.040 & 0.240 $\pm$  0.021 & \underline{0.144 $\pm$ 0.022}  & 0.160  $\pm$ 0.008  &  0.161 $\pm$ 0.060 & 0.354 $\pm$ 0.007\\

        \midrule
         TimeFlow improvement & & / &  24.14 $\%$ &  50.53 $\%$ &  31.61 $\%$ &  36.12 $\%$ &  20.33  $\%$ &  18.90 $\%$ &  53.40 $\%$ \\

        \bottomrule

\end{tabular}}
\end{center}
\end{table}

\begin{figure}[!htb]
    \centering    \includegraphics[width=0.90\linewidth]{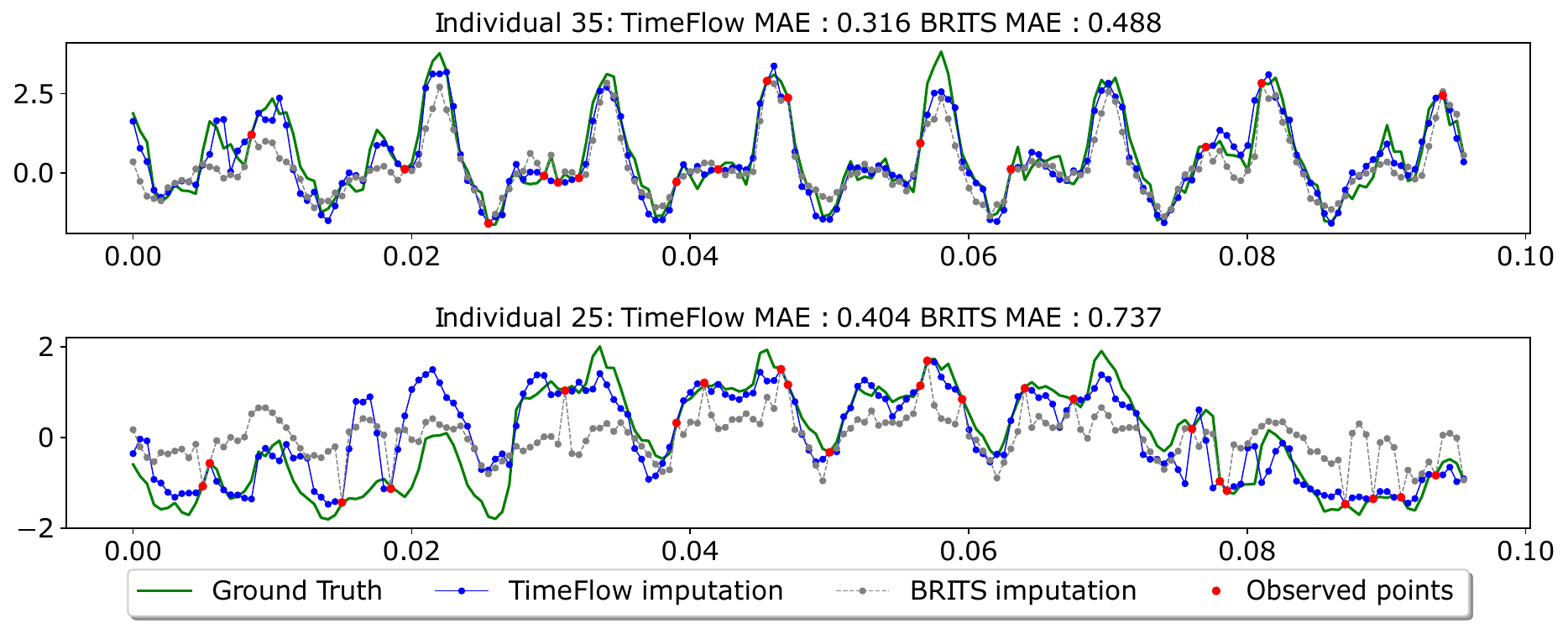}
    \caption{\textit{Electricity dataset}. TimeFlow imputation (blue line) and BRITS imputation (gray line) with 10$\%$ of known point (red points) on the eight first days of samples 35 (top) and 25 (bottom).}
    \label{fig:imputation_core}
\end{figure}

\paragraph{Results.}{
We show in \cref{tab:impu_classic} that TimeFlow outperforms both discrete and continuous models across almost all $\tau$'s for the given datasets. The relative improvements of TimeFlow, as defined in \cref{sec:dataset_prez_norm}, over baselines are significant, ranging from 15\% to 50\%. Especially for the lowest sampling rate $\tau=0.05$, TimeFlow outperforms all discrete baselines, demonstrating the advantages of continuous modeling. Additionally, it achieves lower imputation errors compared to continuous models in all but one cases. Qualitatively, we see on example series in \cref{fig:imputation_core} that our model shows significant imputation capabilities, with a subsampling rate at $\tau=0.1$ on the \textit{Electricity} dataset. It captures well different frequencies and amplitudes in a challenging case (sample 35), although it underestimates the amplitude of some peaks. In a more challenging scenario (sample 25), where the series exhibit additional trend changes and frequency variations within the data, TimeFlow correctly imputes most timestamps, outperforming BRITS, which is the best-performing method for the \textit{Electricity} dataset.}

\paragraph{Imputation on previously unseen time series.} In more practical scenarios, such as cases involving the installation of new sensors, we often encounter new time series originating from the same underlying phenomenon. In such instances, it becomes crucial to make inferences for these previously unseen time series.
Thanks to efficient adaptation in latent space, our model can easily be applied to these new time series (as shown in \cref{imputation_unseen_series}, \cref{tab:imputation_gene_train_test}), contrasting with SOTA methods like SAITS and BRITS, which require full model retraining on the whole set of time series.

\subsection{Forecasting}
\label{sec:expe_forecasting}

\subsubsection{Forecasting for known time series}
\label{subsubsec:classic_setting}

In this section, we are interested in the conventional long-term forecasting scenario. It consists in predicting the phenomenon in a specific future period, the horizon, based on the history of a limited past period, the look-back window. The forecaster is trained on a set of $n$ observed time series for a given time window (train period) and tested on new distinct time windows.

\begin{figure*}[htb!]
    \centering
    \includegraphics[width=0.95\linewidth]{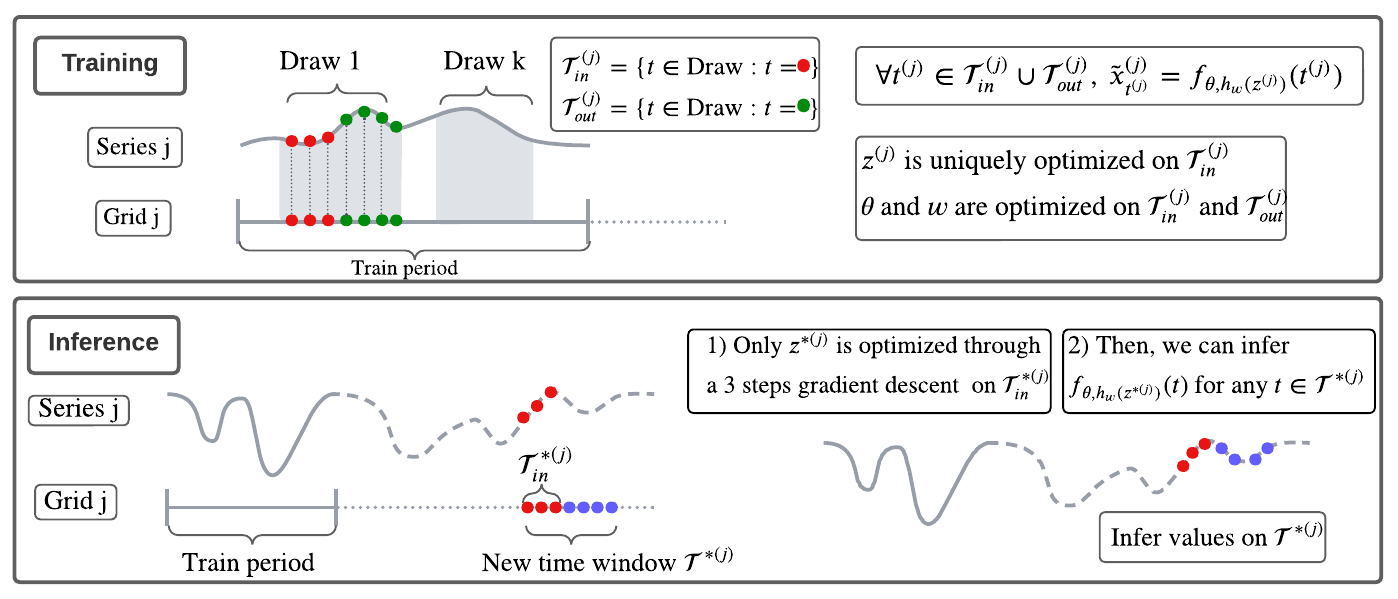}
    \caption[foo bar]{Training and inference procedure of TimeFlow for forecasting. \begin{enumerate*}[(i)] \item During training (top-figure), for each time series $x^{(j)}$, we observe some look-back window/horizon drawing pairs in the trained period. TimeFlow is trained with \cref{alg:TimeFlow} to predict all observed timestamps in this drawing pairs while being conditioned by the observed look-back window. \item Once TimeFlow is optimized, the objective during inference (bottom-figure) is to infer the horizon over new time windows (blue dots \textcolor{blue}{\scalebox{1}{$\bullet$}}) while being conditioned by the newly observed look-back window (red dots \textcolor{red}{\scalebox{1}{$\bullet$}}). \end{enumerate*}}
    \label{fig:forecasting_procedure}
\end{figure*}

\paragraph{Setting.}{
For a given time series $x^{(j)}$, $\mathcal{T}^{(j)}_{in}$ denotes the look-back window 
and $\mathcal{T}^{(j)}_{out}$ the horizon of $H$ points. During training, at each epoch, we train $f_{\theta, h_w(z^{(j)})}$ following \cref{alg:TimeFlow} with randomly drawn pairs of look-back window and horizon $(\mathcal{T}^{(j)}_{in} \cup \mathcal{T}^{(j)}_{out})_{j \in \mathcal{B}}$ within the observed train period. Then, for a distinct new time window $\mathcal{T}^{*(j)}$, given a look-back window $\mathcal{T}_{in}^{* (j)}$ we forecast future values any $t \in \mathcal{T}^{* (j)}$, the horizon interval, following \cref{alg:TimeFlow_inference}. We illustrate the training and inference of TimeFlow for the forecasting task in \cref{fig:forecasting_procedure}. For further insight into the training window and inference periods, as well as additional experiments conducted under different inference scenarios, see \cref{sec:diff_adjacent_new_time_windows_appendix}.}

\paragraph{Baselines.} To evaluate the quality of our model in long-term forecasting, we compare it to the discrete baselines PatchTST \citep{PatchTST}, DLinear \citep{Dlinear}, AutoFormer \citep{AutoFormer}, and Informer \citep{Informer}. We also include continuous baselines DeepTime and Neural Process (NP). See \cref{baseline-forecast-training} for the baseline training procedure and hyperparameter selection. In \cref{tab:forecast_time_shift_electricity}, we present the forecasting results for standard horizons in long-term forecasting: $H \in \{96, 192, 336, 720\}$. The look-back window length is fixed to 512.

\begin{table}[htbp]
\caption{Mean MAE forecast results averaged over different time windows. Each time, the model is trained on one time window and tested on the others (there are 2 windows for \textit{SolarH} and 5 for \textit{Electricity} and \textit{Traffic}). $H$ stands for the horizon. Bold results are best, and underlined results are second best. TimeFlow improvement represents the overall percentage improvement achieved by TimeFlow compared to the specific method being considered.}
\label{tab:forecast_time_shift_electricity}
\begin{center}
\resizebox{\textwidth}{!}{%
\begin{tabular}{ccccccccc}

\toprule

  &        & \multicolumn{3}{c}{Continuous methods} & \multicolumn{4}{c}{Discrete methods} \\
  
  \cmidrule(r){3-5} \cmidrule(r){6-9} 
  & $H$ &  TimeFlow & DeepTime & Neural Process & Patch-TST & DLinear & AutoFormer & Informer\\

        \midrule

        \multirow{4}{*}{Electricity} & 96 & \underline{0.228 $\pm$ 0.028} & 0.244 $\pm$ 0.026 & 0.392 $\pm$ 0.045 & \textbf{0.221 $\pm$ 0.023}  & 0.241 $\pm$  0.030 & 0.546 $\pm$ 0.277 & 0.603 $\pm$ 0.255 \\

        & 192  & \underline{0.238 $\pm$ 0.020} & 0.252 $\pm$ 0.019 & 0.401 $\pm$ 0.046 & \textbf{0.229 $\pm$ 0.020} & 0.252 $\pm$ 0.025 & 0.500 $\pm$ 0.190  & 0.690 $\pm$ 0.291\\

        & 336 & \underline{0.270 $\pm$ 0.031} & 0.284 $\pm$ 0.034 & 0.434 $\pm$ 0.076 & \textbf{0.251 $\pm$ 0.027} & 0.288 $\pm$ 0.038 & 0.523 $\pm$ 0.188 & 0.736 $\pm$ 0.271 \\

        & 720 & \underline{0.316 $\pm$ 0.055} & 0.359 $\pm$ 0.051 & 0.607 $\pm$ 0.150 &\textbf{0.297 $\pm$ 0.039} & 0.365 $\pm$ 0.059 & 0.631 $\pm$ 0.237 & 0.746 $\pm$ 0.265\\

        \midrule

        \multirow{4}{*}{SolarH} & 96 & \textbf{0.190 $\pm$ 0.013} &  \textbf{0.190 $\pm$ 0.020} & 0.221 $\pm$ 0.048 & 0.262 $\pm$ 0.070 & 0.208 $\pm$ 0.014 & 0.245 $\pm$ 0.045 & 0.248 $\pm$ 0.022 \\

        & 192 & \textbf{0.202 $\pm$ 0.020} & \underline{0.204 $\pm$ 0.028} & 0.244 $\pm$ 0.048 & 0.253 $\pm$ 0.051 & 0.217 $\pm$ 0.022 & 0.333 $\pm$ 0.107 & 0.270 $\pm$ 0.031 \\

        & 336 & \underline{0.209 $\pm$ 0.017} & \textbf{0.199 $\pm$ 0.026} & 0.240 $\pm$ 0.006 & 0.259 $\pm$ 0.071 & 0.217 $\pm$ 0.026 & 0.334 $\pm$ 0.079 & 0.328 $\pm$ 0.048 \\

        & 720 & \textbf{0.218 $\pm$ 0.041} & \underline{0.229 $\pm$ 0.024} & 0.403 $\pm$ 0.147 & 0.267 $\pm$ 0.064 & 0.249 $\pm$ 0.034 & 0.351 $\pm$ 0.055 & 0.337 $\pm$ 0.037\\
                
        \midrule

        \multirow{4}{*}{Traffic} & 96 & \underline{0.217 $\pm$ 0.032} & 0.228 $\pm$ 0.032 & 0.283 $\pm$ 0.027 & \textbf{0.203 $\pm$ 0.037} & 0.228 $\pm$ 0.033 & 0.319 $\pm$ 0.059 & 0.372 $\pm$ 0.078 \\

        & 192 & \underline{0.212 $\pm$ 0.028} & 0.220 $\pm$ 0.022 & 0.292 $\pm$ 0.024 &\textbf{0.197 $\pm$ 0.030} & 0.221 $\pm$ 0.023 & 0.368 $\pm$ 0.057 & 0.511 $\pm$ 0.247 \\

        & 336 & \underline{0.238 $\pm$ 0.034} & 0.245 $\pm$ 0.038 & 0.305 $\pm$ 0.039 & \textbf{0.222 $\pm$ 0.039} & 0.250 $\pm$ 0.040 & 0.434 $\pm$ 0.061 & 0.561 $\pm$ 0.263 \\

        & 720 & \underline{0.279 $\pm$ 0.050} & 0.290 $\pm$ 0.052 & 0.339 $\pm$ 0.038 & \textbf{0.269 $\pm$ 0.057} & 0.300 $\pm$  0.057 & 0.462 $\pm$ 0.062 & 0.638 $\pm$ 0.067 \\

        \midrule
         TimeFlow improvement & & / &  3.74 $\%$ &  29.06 $\%$ &  3.23 $\%$ &  6.92 $\%$ &   42.09 $\%$ &  48.57 $\%$ \\
        
        \bottomrule
\end{tabular}}
\end{center}
\end{table}



\paragraph{Results.}
The results in \cref{tab:forecast_time_shift_electricity} show that our approach ranks in the top two across all datasets and horizons and is the overall best continuous method. TimeFlow's performance is comparable to the current SOTA model PatchTST, with only 2\% relative difference. Moreover, TimeFlow shows consistent results across the three datasets, whereas the other best discrete and continuous baselines, \ie PatchTST and DeepTime, performance drops for some datasets. We also note that, despite the great performance of the SOTA PatchTST, other transformer-based baselines (discrete methods in \cref{tab:forecast_time_shift_electricity}) perform poorly. We provide a detailed insight on these results in \cref{sec:diff_adjacent_new_time_windows_appendix}. 
Overall, although this evaluation setting favors discrete methods because the time series are observed at evenly distributed time steps, TimeFlow consistently performs as well as PatchTST and outperforms all the other methods, whether discrete or continuous. It is the first time that a continuous model has achieved the same level of performance as discrete methods within their specific setting.

\subsubsection{Forecasting on previously unseen time series.}
\label{subsubsec:new_samples_setting}

This section discusses how TimeFlow adapts to unseen time series, which is critical in forecasting. Indeed, in many real-world applications, forecasters are trained on a limited subset of available samples and applied to a wider range of samples during inference. Informer, AutoFormer, or DLinear original architectures directly model the relationships between time series (channel-dependence), limiting their adaptability to new samples. In contrast, TimeFlow takes a different approach by considering the observed series at different locations as distinct samples, similar to PatchTST, Neural Process, and DeepTime. This independence allows TimeFlow to effectively generalize to previously unseen time series of the same phenomenon. 

\begin{figure}[H]
    \centering
    \includegraphics[width=0.99\linewidth]{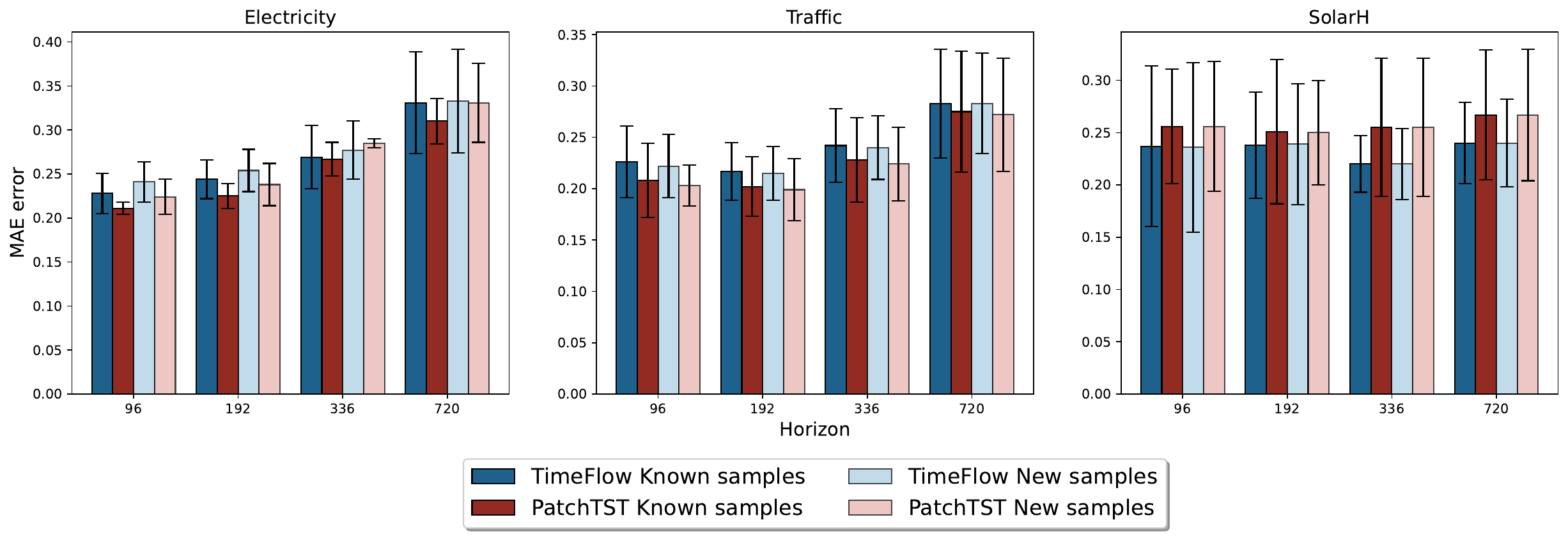}
    \caption{Mean MAE forecasting task results over different horizons in the context of generalization to new time series. Comparison of TimeFlow and PatchTST performances on the \textit{Electricity}, \textit{Traffic} and \textit{SolarH} datasets.}
    \label{fig:barplot_new_samples}
\end{figure}

\paragraph{Setting.}In this setting, we propose to evaluate how TimeFlow performs on previously unseen time series. We compare it to the best forecaster, PatchTST. We train TimeFlow and PatchTST on 50 \% of the samples and consider the remaining 50 \% as the new time series. The training procedure is the same as described in \cref{fig:forecasting_procedure}. In \cref{fig:barplot_new_samples}, we present the results of TimeFlow and PatchTST for both known and new samples (for periods outside the training window).

\paragraph{Results.}{The results in \cref{fig:barplot_new_samples} highlight two key observations. First, both approaches show robust adaptability to new samples, as evidenced by the minimal difference in mean absolute error between known and new samples at inference. Second, TimeFlow and PatchTST exhibit comparable performance in this context, with negligible differences across horizons and datasets.}

\clearpage

\subsection{Challenging task: Forecast while imputing incomplete look-back windows}
\label{sec:impute_forecast}

In real-world scenarios, it is common to encounter missing or irregularly sampled series when making predictions on new time windows \citep{DBLP:journals/ijon/CinarMGGA18, DBLP:conf/aaai/TangYSAMW20}. Continuous methods can handle these cases, as they are designed to accommodate irregular sampling within the look-back window. In this section, we formulate a task to simulate these real-world scenarios. 
It's worth noting that this task is often encountered in practice but is rarely considered in the DL literature.

\looseness=-1
\paragraph{Setting and baselines.}{This scenario is similar to the forecast setting in \cref{sec:expe_forecasting} and illustrated in \cref{fig:forecasting_procedure}. The difference is that during inference, the look-back window is subsampled at a rate $\tau$ smaller than the one used for the training phase. This simulates a situation with missing observations in the look back window. Consequently, two distinct tasks emerge during the inference phase: imputing missing points within the sparsely observed look-back window, and forecasting over the horizon with this degraded context. In \cref{tab:flexibility_two_datasets}, we compare to the two other continuous baselines, DeepTime and NP on \textit{Electricity} and \textit{Traffic} for different $\tau$'s and horizons.
}

\begin{table}[htb]
\caption{MAE results for forecasting with missing values in the look-back window. $\tau$ stands for the percentage of observed values in the look-back window. Best results are in bold.}
\label{tab:flexibility_two_datasets}
\begin{center}
\scalebox{0.63}{
\begin{tabular}{ccccccccc}
\toprule
  &   &        & \multicolumn{2}{c}{TimeFlow} & \multicolumn{2}{c}{DeepTime} & \multicolumn{2}{c}{Neural Process} \\
  
  \cmidrule(r){4-5} \cmidrule(r){6-7} \cmidrule(r){8-9}                                       
  
  & $H$ & $\tau$ & Imputation error & Forecast error & Imputation error & Forecast error & Imputation error & Forecast error\\
  
        \midrule
        \multirow{6}{*}{Electricity} &  \multirow{3}{*}{96}  & 0.5 & \textbf{0.151 $\pm$ 0.003} & \textbf{0.239 $\pm$ 0.013} & 0.209 $\pm$ 0.004 & 0.270 $\pm$ 0.019 & 0.460 $\pm$ 0.048 & 0.486 $\pm$ 0.078\\
                                      &                      & 0.2 & \textbf{0.208 $\pm$ 0.006} & \textbf{0.260 $\pm$ 0.015} & 0.249 $\pm$ 0.006 & 0.296 $\pm$ 0.023 & 0.644 $\pm$ 0.079 & 0.650 $\pm$ 0.095\\
                                      &                      & 0.1 & \textbf{0.272 $\pm$ 0.006} & \textbf{0.295 $\pm$ 0.016} & 0.284 $\pm$ 0.007  & 0.324 $\pm$ 0.026 & 0.740 $\pm$ 0.083 & 0.737 $\pm$ 0.106\\

                            \cmidrule(r){2-9}
                    
                                       & \multirow{3}{*}{192} & 0.5 & \textbf{0.149 $\pm$ 0.004} & \textbf{0.235 $\pm$ 0.011} & 0.204 $\pm$ 0.004 & 0.265 $\pm$ 0.018 & 0.461 $\pm$ 0.045 & 0.498 $\pm$ 0.070 \\
                                       &                      & 0.2 & \textbf{0.209 $\pm$ 0.006} & \textbf{0.257 $\pm$ 0.013} & 0.244 $\pm$ 0.007 & 0.290 $\pm$ 0.023 & 0.601 $\pm$ 0.075 & 0.626 $\pm$ 0.101 \\
                                       &                      & 0.1 & \textbf{0.274 $\pm$ 0.010} & \textbf{0.289 $\pm$ 0.016} & 0.282 $\pm$ 0.007 &  0.315 $\pm$ 0.025 & 0.461 $\pm$ 0.045 & 0.724 $\pm$ 0.090\\

        \midrule

\multirow{6}{*}{Traffic} & \multirow{3}{*}{96} & 0.5 & \textbf{0.180 $\pm$ 0.016} & \textbf{0.219 $\pm$ 0.026} & 0.272 $\pm$ 0.028 & 0.243 $\pm$ 0.030 & 0.436 $\pm$ 0.025 & 0.444 $\pm$ 0.047\\
                                    &          & 0.2 & \textbf{0.239 $\pm$ 0.019} & \textbf{0.243 $\pm$ 0.027} & 0.335 $\pm$ 0.026 & 0.293 $\pm$ 0.027 & 0.596 $\pm$ 0.049 & 0.597 $\pm$ 0.075\\
                                          &    & 0.1 & \textbf{0.312 $\pm$ 0.020} & \textbf{0.290 $\pm$ 0.027} & 0.385 $\pm$ 0.025 & 0.344 $\pm$ 0.027 & 0.734 $\pm$ 0.102 & 0.731 $\pm$ 0.132\\

                    \cmidrule(r){2-9}
                    
              &  \multirow{3}{*}{192}   & 0.5 & \textbf{0.176 $\pm$ 0.014} & \textbf{0.217 $\pm$ 0.017} & 0.241 $\pm$ 0.027 & 0.234 $\pm$ 0.021 & 0.477 $\pm$ 0.042 & 0.476 $\pm$ 0.043 \\
                                  &     & 0.2 & \textbf{0.233 $\pm$ 0.017} & \textbf{0.236 $\pm$ 0.021} & 0.286 $\pm$ 0.027 & 0.276 $\pm$ 0.020 & 0.685 $\pm$ 0.109 & 0.678 $\pm$ 0.108 \\
                                  &     & 0.1 & \textbf{0.304 $\pm$ 0.019} & \textbf{0.277 $\pm$ 0.021} & 0.331 $\pm$ 0.025 & 0.324 $\pm$ 0.021 & 0.888 $\pm$ 0.178 & 0.877 $\pm$ 0.174 \\

            \midrule
            
             TimeFlow improvement & & & / & / &  18.97 $\%$ &  11.87 $\%$ &  61.88 $\%$ &  58.41 $\%$ \\
        
        \bottomrule

\end{tabular}}
\end{center}
\end{table}





\begin{figure}[H]
    \centering
    \includegraphics[width=0.78\linewidth]{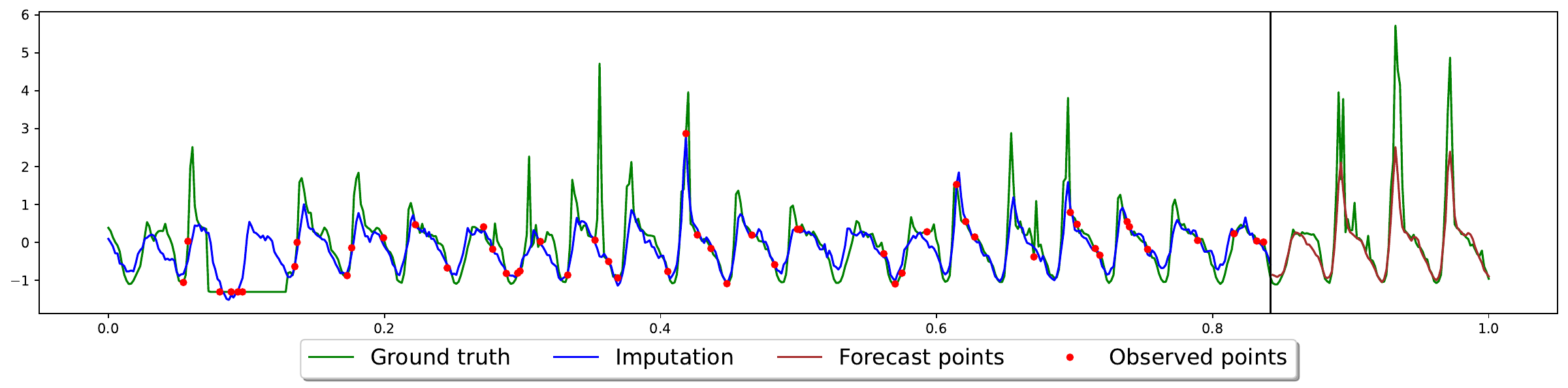}
    \caption{\textit{Traffic dataset, sample 95}. In this figure, TimeFlow simultaneously imputes and forecasts at horizon 96 with a 10\% partially observed look-back window of length 512.}
    \label{fig:imputation_forecast_traffic}
\end{figure}

\paragraph{Results.}{In \cref{tab:flexibility_two_datasets}, the results show that TimeFlow consistently outperforms other methods in imputation and forecasting for every scenarios. When comparing with the complete look-back windows observations scenario from \cref{tab:forecast_time_shift_electricity}, one observes that at a 0.5 sampling rate, TimeFlow presents only a slight reduction in performance, whereas other baseline methods experience more significant drops. For instance, when we compare forecast results between a complete window and a $\tau=0.5$ subsampled window for \textit{Electricity} with a forecasting horizon of $H=96$, TimeFlow's error increases by a mere 4.6\% (from 0.228 to 0.239). In contrast, DeepTime's error grows by over 10\% (from 0.244 to 0.270), and NP experiences a rise of around 25\% (from 0.392 to 0.486). For lower sampling rates, TimeFlow still delivers correct predictions. Qualitatively, we see on the series example in \cref{fig:imputation_forecast_traffic} that despite observing only 10\% of the look-back window, the model can correctly infer both the complete look-back window and the horizon. 
Both quantitative and qualitative results show the robustness and efficiency of TimeFlow on this particularly challenging setting.}

\section{Limitations}

While TimeFlow shows promising performance across various tasks and settings, it is important to recognize some limitations. First, due to its auto-decoding process, TimeFlow tends to be significantly slower at inference time compared to other baselines by one to two orders of magnitude (\cref{tab:inference_time}). In addition, although TimeFlow effectively handles sets of homogeneous time series, additional mechanisms are required to handle heterogeneous time series with different frequencies effectively. The per-context shift modulation mechanism does not allow TimeFlow to fit time series with drastically different structures. This also explains why TimeFlow is particularly well adapted to regular frequency patterns.In the experiment section, all considered datasets exhibit pronounced periodicities. Finally, it should be noted that effective training of TimeFlow requires a relatively large number of samples (typically $\ge$ 100) to allow the model to accurately distinguish between individual patterns and shared information.

\section{Conclusion}
We have introduced a unified framework for continuous time series modeling leveraging conditional INR and meta-learning. Our experiments have demonstrated superior performance compared to other continuous methods, and better or comparable results to SOTA discrete methods. One of the standout features of our framework is its inherent continuity and the ability to modulate the INR parameters. This unique flexibility lets TimeFlow effectively tackle a wide array of challenges, including forecasting in the presence of missing values, accommodating irregular time steps, and extending the trained model's applicability to previously unseen time series and new time windows. Our empirical results have shown TimeFlow's effectiveness in handling homogeneous multivariate time series. As a logical next step, extending TimeFlow's capabilities to address heterogeneous multivariate phenomena represents a promising direction for future research.

\section*{Acknowledgment}

We would like to thank Tahar Nabil for the valuable discussions on this project.

\bibliographystyle{abbrvnat}
\bibliography{biblio}

\begin{thebibliography}{44}
\providecommand{\natexlab}[1]{#1}
\providecommand{\url}[1]{\texttt{#1}}
\expandafter\ifx\csname urlstyle\endcsname\relax
  \providecommand{\doi}[1]{doi: #1}\else
  \providecommand{\doi}{doi: \begingroup \urlstyle{rm}\Url}\fi

\bibitem[Bilos et~al.(2023)Bilos, Rasul, Schneider, Nevmyvaka, and G{\"{u}}nnemann]{BilosRSNG23}
M.~Bilos, K.~Rasul, A.~Schneider, Y.~Nevmyvaka, and S.~G{\"{u}}nnemann.
\newblock Modeling temporal data as continuous functions with stochastic process diffusion.
\newblock In A.~Krause, E.~Brunskill, K.~Cho, B.~Engelhardt, S.~Sabato, and J.~Scarlett, editors, \emph{International Conference on Machine Learning, {ICML}}, volume 202 of \emph{Proceedings of Machine Learning Research}, pages 2452--2470. {PMLR}, 2023.

\bibitem[Brouwer et~al.(2019)Brouwer, Simm, Arany, and Moreau]{BrouwerSAM19}
E.~D. Brouwer, J.~Simm, A.~Arany, and Y.~Moreau.
\newblock Gru-ode-bayes: continuous modeling of sporadically-observed time series.
\newblock In \emph{Proceedings of the 33rd International Conference on Neural Information Processing Systems}, pages 7379--7390, 2019.

\bibitem[Cao et~al.(2018)Cao, Wang, Li, Zhou, Li, and Li]{cao2018brits}
W.~Cao, D.~Wang, J.~Li, H.~Zhou, Y.~Li, and L.~Li.
\newblock Brits: bidirectional recurrent imputation for time series.
\newblock In \emph{Proceedings of the 32nd International Conference on Neural Information Processing Systems}, pages 6776--6786, 2018.

\bibitem[Chen et~al.(2001)Chen, Grant-Muller, Mussone, and Montgomery]{chen2001study}
H.~Chen, S.~Grant-Muller, L.~Mussone, and F.~Montgomery.
\newblock A study of hybrid neural network approaches and the effects of missing data on traffic forecasting.
\newblock \emph{Neural Computing \& Applications}, 10:\penalty0 277--286, 2001.

\bibitem[Chen and Guestrin(2016)]{chen2016xgboost}
T.~Chen and C.~Guestrin.
\newblock Xgboost: A scalable tree boosting system.
\newblock In \emph{Proceedings of the 22nd acm sigkdd international conference on knowledge discovery and data mining}, pages 785--794, 2016.

\bibitem[Cinar et~al.(2018)Cinar, Mirisaee, Goswami, Gaussier, and A{\"{\i}}t{-}Bachir]{DBLP:journals/ijon/CinarMGGA18}
Y.~G. Cinar, H.~Mirisaee, P.~Goswami, {\'{E}}.~Gaussier, and A.~A{\"{\i}}t{-}Bachir.
\newblock Period-aware content attention rnns for time series forecasting with missing values.
\newblock \emph{Neurocomputing}, 312:\penalty0 177--186, 2018.

\bibitem[Clark and Bj{\o}rnstad(2004)]{clark2004population}
J.~S. Clark and O.~N. Bj{\o}rnstad.
\newblock Population time series: process variability, observation errors, missing values, lags, and hidden states.
\newblock \emph{Ecology}, 85\penalty0 (11):\penalty0 3140--3150, 2004.

\bibitem[Corani et~al.(2021)Corani, Benavoli, and Zaffalon]{GPpriorneeded}
G.~Corani, A.~Benavoli, and M.~Zaffalon.
\newblock Time series forecasting with gaussian processes needs priors.
\newblock In \emph{Machine Learning and Knowledge Discovery in Databases. Applied Data Science Track: European Conference, ECML PKDD 2021, Bilbao, Spain, September 13--17, 2021, Proceedings, Part IV 21}, pages 103--117. Springer, 2021.

\bibitem[Du et~al.(2023)Du, C{\^o}t{\'e}, and Liu]{du2023saits}
W.~Du, D.~C{\^o}t{\'e}, and Y.~Liu.
\newblock Saits: Self-attention-based imputation for time series.
\newblock \emph{Expert Systems with Applications}, 219:\penalty0 119619, 2023.

\bibitem[Dupont et~al.(2022)Dupont, Kim, Eslami, Rezende, and Rosenbaum]{Functa}
E.~Dupont, H.~Kim, S.~M.~A. Eslami, D.~J. Rezende, and D.~Rosenbaum.
\newblock From data to functa: Your data point is a function and you can treat it like one.
\newblock In \emph{International Conference on Machine Learning, {ICML} 2022, 17-23 July 2022, Baltimore, Maryland, {USA}}, volume 162 of \emph{Proceedings of Machine Learning Research}, pages 5694--5725. {PMLR}, 2022.

\bibitem[Fathony et~al.(2021)Fathony, Sahu, Willmott, and Kolter]{MFN}
R.~Fathony, A.~K. Sahu, D.~Willmott, and J.~Z. Kolter.
\newblock Multiplicative filter networks.
\newblock In \emph{International Conference on Learning Representations}, 2021.

\bibitem[Finn et~al.(2017)Finn, Abbeel, and Levine]{finn2017model}
C.~Finn, P.~Abbeel, and S.~Levine.
\newblock Model-agnostic meta-learning for fast adaptation of deep networks.
\newblock In \emph{International conference on machine learning}, pages 1126--1135. PMLR, 2017.

\bibitem[Fons et~al.(2022)Fons, Sztrajman, El{-}Laham, Iosifidis, and Vyetrenko]{HyperTime}
E.~Fons, A.~Sztrajman, Y.~El{-}Laham, A.~Iosifidis, and S.~Vyetrenko.
\newblock Hypertime: Implicit neural representation for time series.
\newblock \emph{CoRR}, abs/2208.05836, 2022.

\bibitem[Fortuin et~al.(2020)Fortuin, Baranchuk, R{\"a}tsch, and Mandt]{fortuin2020gp}
V.~Fortuin, D.~Baranchuk, G.~R{\"a}tsch, and S.~Mandt.
\newblock Gp-vae: Deep probabilistic time series imputation.
\newblock In \emph{International conference on artificial intelligence and statistics}, pages 1651--1661. PMLR, 2020.

\bibitem[Garnelo et~al.(2018)Garnelo, Rosenbaum, Maddison, Ramalho, Saxton, Shanahan, Teh, Rezende, and Eslami]{GarneloRMRSSTRE18}
M.~Garnelo, D.~Rosenbaum, C.~Maddison, T.~Ramalho, D.~Saxton, M.~Shanahan, Y.~W. Teh, D.~J. Rezende, and S.~M.~A. Eslami.
\newblock Conditional neural processes.
\newblock In \emph{Proceedings of the 35th International Conference on Machine Learning, {ICML}}, volume~80, pages 1690--1699. {PMLR}, 2018.

\bibitem[Hastie(2017)]{hastie2017generalized}
T.~J. Hastie.
\newblock Generalized additive models.
\newblock In \emph{Statistical models in S}, pages 249--307. Routledge, 2017.

\bibitem[Huang and Hoefler(2023)]{Huang2023}
L.~Huang and T.~Hoefler.
\newblock Compressing multidimensional weather and climate data into neural networks.
\newblock In \emph{International Conference on Learning Representations, ICLR}, 2023.

\bibitem[Hyndman and Athanasopoulos(2018)]{hyndman2018forecasting}
R.~J. Hyndman and G.~Athanasopoulos.
\newblock \emph{Forecasting: principles and practice}.
\newblock OTexts, 2018.

\bibitem[Jeong and Shin(2022)]{INRAD}
K.~Jeong and Y.~Shin.
\newblock Time-series anomaly detection with implicit neural representation.
\newblock \emph{CoRR}, abs/2201.11950, 2022.

\bibitem[Kim et~al.(2019)Kim, Ko, and Kim]{kim2019analysis}
T.~Kim, W.~Ko, and J.~Kim.
\newblock Analysis and impact evaluation of missing data imputation in day-ahead pv generation forecasting.
\newblock \emph{Applied Sciences}, 9\penalty0 (1):\penalty0 204, 2019.

\bibitem[Liu et~al.(2022)Liu, Yu, Liao, Li, Lin, Liu, and Dustdar]{PyraFormer}
S.~Liu, H.~Yu, C.~Liao, J.~Li, W.~Lin, A.~X. Liu, and S.~Dustdar.
\newblock Pyraformer: Low-complexity pyramidal attention for long-range time series modeling and forecasting.
\newblock In \emph{The Tenth International Conference on Learning Representations, {ICLR} 2022, Virtual Event, April 25-29, 2022}, 2022.

\bibitem[Liu et~al.(2023)Liu, Li, Cong, Chen, and Jiang]{TIDER}
S.~Liu, X.~Li, G.~Cong, Y.~Chen, and Y.~Jiang.
\newblock Multivariate time-series imputation with disentangled temporal representations.
\newblock In \emph{The Eleventh International Conference on Learning Representations, ICLR}, 2023.

\bibitem[Liu et~al.(2019)Liu, Yu, Zheng, Zhan, and Yue]{liu2019naomi}
Y.~Liu, R.~Yu, S.~Zheng, E.~Zhan, and Y.~Yue.
\newblock Naomi: Non-autoregressive multiresolution sequence imputation.
\newblock \emph{Advances in neural information processing systems}, 32, 2019.

\bibitem[Luo et~al.(2018)Luo, Cai, Zhang, Xu, et~al.]{luo2018multivariate}
Y.~Luo, X.~Cai, Y.~Zhang, J.~Xu, et~al.
\newblock Multivariate time series imputation with generative adversarial networks.
\newblock \emph{Advances in neural information processing systems}, 31, 2018.

\bibitem[Luo et~al.(2019)Luo, Zhang, Cai, and Yuan]{luo2019e2gan}
Y.~Luo, Y.~Zhang, X.~Cai, and X.~Yuan.
\newblock E2gan: End-to-end generative adversarial network for multivariate time series imputation.
\newblock In \emph{Proceedings of the 28th international joint conference on artificial intelligence}, pages 3094--3100. AAAI Press, 2019.

\bibitem[Mildenhall et~al.(2021)Mildenhall, Srinivasan, Tancik, Barron, Ramamoorthi, and Ng]{mildenhall2021nerf}
B.~Mildenhall, P.~P. Srinivasan, M.~Tancik, J.~T. Barron, R.~Ramamoorthi, and R.~Ng.
\newblock Nerf: Representing scenes as neural radiance fields for view synthesis.
\newblock \emph{Communications of the ACM}, 65\penalty0 (1):\penalty0 99--106, 2021.

\bibitem[Nichol et~al.(2018)Nichol, Achiam, and Schulman]{nichol2018first}
A.~Nichol, J.~Achiam, and J.~Schulman.
\newblock On first-order meta-learning algorithms.
\newblock \emph{arXiv preprint arXiv:1803.02999}, 2018.

\bibitem[Nie et~al.(2022)Nie, Nguyen, Sinthong, and Kalagnanam]{PatchTST}
Y.~Nie, N.~H. Nguyen, P.~Sinthong, and J.~Kalagnanam.
\newblock A time series is worth 64 words: Long-term forecasting with transformers.
\newblock \emph{CoRR}, abs/2211.14730, 2022.

\bibitem[Rasmussen and Williams(2006)]{RasmussenW06}
C.~E. Rasmussen and C.~K.~I. Williams.
\newblock \emph{Gaussian processes for machine learning}.
\newblock Adaptive computation and machine learning. {MIT} Press, 2006.

\bibitem[Rubanova et~al.(2019)Rubanova, Chen, and Duvenaud]{corr/abs-1907-03907}
Y.~Rubanova, R.~T.~Q. Chen, and D.~Duvenaud.
\newblock Latent odes for irregularly-sampled time series.
\newblock \emph{CoRR}, abs/1907.03907, 2019.

\bibitem[Schulz and Stattegger(1997)]{schulz1997spectrum}
M.~Schulz and K.~Stattegger.
\newblock Spectrum: Spectral analysis of unevenly spaced paleoclimatic time series.
\newblock \emph{Computers \& Geosciences}, 23\penalty0 (9):\penalty0 929--945, 1997.

\bibitem[Shukla and Marlin(2021)]{ShuklaM21}
S.~N. Shukla and B.~M. Marlin.
\newblock Multi-time attention networks for irregularly sampled time series.
\newblock In \emph{9th International Conference on Learning Representations, {ICLR} 2021, Virtual Event, Austria, May 3-7, 2021}, 2021.

\bibitem[Sitzmann et~al.(2020)Sitzmann, Martel, Bergman, Lindell, and Wetzstein]{SIREN}
V.~Sitzmann, J.~N.~P. Martel, A.~W. Bergman, D.~B. Lindell, and G.~Wetzstein.
\newblock Implicit neural representations with periodic activation functions.
\newblock In \emph{Advances in Neural Information Processing Systems 33: Annual Conference on Neural Information Processing Systems 2020, NeurIPS 2020, December 6-12, 2020, virtual}, 2020.

\bibitem[Tancik et~al.(2020)Tancik, Srinivasan, Mildenhall, Fridovich-Keil, Raghavan, Singhal, Ramamoorthi, Barron, and Ng]{tancik2020fourier}
M.~Tancik, P.~Srinivasan, B.~Mildenhall, S.~Fridovich-Keil, N.~Raghavan, U.~Singhal, R.~Ramamoorthi, J.~Barron, and R.~Ng.
\newblock Fourier features let networks learn high frequency functions in low dimensional domains.
\newblock \emph{Advances in Neural Information Processing Systems}, 33:\penalty0 7537--7547, 2020.

\bibitem[Tang et~al.(2020)Tang, Yao, Sun, Aggarwal, Mitra, and Wang]{DBLP:conf/aaai/TangYSAMW20}
X.~Tang, H.~Yao, Y.~Sun, C.~C. Aggarwal, P.~Mitra, and S.~Wang.
\newblock Joint modeling of local and global temporal dynamics for multivariate time series forecasting with missing values.
\newblock In \emph{The Thirty-Fourth {AAAI} Conference on Artificial Intelligence, {AAAI} 2020, The Thirty-Second Innovative Applications of Artificial Intelligence Conference, {IAAI} 2020, The Tenth {AAAI} Symposium on Educational Advances in Artificial Intelligence, {EAAI}}, pages 5956--5963. {AAAI} Press, 2020.

\bibitem[Tashiro et~al.(2021)Tashiro, Song, Song, and Ermon]{tashiro2021csdi}
Y.~Tashiro, J.~Song, Y.~Song, and S.~Ermon.
\newblock Csdi: Conditional score-based diffusion models for probabilistic time series imputation.
\newblock \emph{Advances in Neural Information Processing Systems}, 34:\penalty0 24804--24816, 2021.

\bibitem[Taylor and Letham(2018)]{taylor2018forecasting}
S.~J. Taylor and B.~Letham.
\newblock Forecasting at scale.
\newblock \emph{The American Statistician}, 72\penalty0 (1):\penalty0 37--45, 2018.

\bibitem[Woo et~al.(2022)Woo, Liu, Sahoo, Kumar, and Hoi]{DeepTime}
G.~Woo, C.~Liu, D.~Sahoo, A.~Kumar, and S.~C.~H. Hoi.
\newblock Deeptime: Deep time-index meta-learning for non-stationary time-series forecasting.
\newblock \emph{CoRR}, abs/2207.06046, 2022.

\bibitem[Wu et~al.(2021)Wu, Xu, Wang, and Long]{AutoFormer}
H.~Wu, J.~Xu, J.~Wang, and M.~Long.
\newblock Autoformer: Decomposition transformers with auto-correlation for long-term series forecasting.
\newblock In M.~Ranzato, A.~Beygelzimer, Y.~N. Dauphin, P.~Liang, and J.~W. Vaughan, editors, \emph{Advances in Neural Information Processing Systems 34: Annual Conference on Neural Information Processing Systems 2021, NeurIPS 2021, December 6-14, 2021, virtual}, pages 22419--22430, 2021.

\bibitem[Yin et~al.(2023)Yin, Kirchmeyer, Franceschi, Rakotomamonjy, and Gallinari]{DiNO}
Y.~Yin, M.~Kirchmeyer, J.-Y. Franceschi, A.~Rakotomamonjy, and P.~Gallinari.
\newblock Continuous pde dynamics forecasting with implicit neural representations.
\newblock In \emph{International Conference on Learning Representations, ICLR}, 2023.

\bibitem[Zeng et~al.(2022)Zeng, Chen, Zhang, and Xu]{Dlinear}
A.~Zeng, M.~Chen, L.~Zhang, and Q.~Xu.
\newblock Are transformers effective for time series forecasting?
\newblock \emph{CoRR}, abs/2205.13504, 2022.

\bibitem[Zhou et~al.(2021)Zhou, Zhang, Peng, Zhang, Li, Xiong, and Zhang]{Informer}
H.~Zhou, S.~Zhang, J.~Peng, S.~Zhang, J.~Li, H.~Xiong, and W.~Zhang.
\newblock Informer: Beyond efficient transformer for long sequence time-series forecasting.
\newblock In \emph{Proceedings of the AAAI conference on artificial intelligence}, pages 11106--11115, 2021.

\bibitem[Zhou et~al.(2022)Zhou, Ma, Wen, Wang, Sun, and Jin]{FedFormer}
T.~Zhou, Z.~Ma, Q.~Wen, X.~Wang, L.~Sun, and R.~Jin.
\newblock Fedformer: Frequency enhanced decomposed transformer for long-term series forecasting.
\newblock In K.~Chaudhuri, S.~Jegelka, L.~Song, C.~Szepesv{\'{a}}ri, G.~Niu, and S.~Sabato, editors, \emph{International Conference on Machine Learning, {ICML} 2022, 17-23 July 2022, Baltimore, Maryland, {USA}}, volume 162 of \emph{Proceedings of Machine Learning Research}, pages 27268--27286. {PMLR}, 2022.

\bibitem[Zintgraf et~al.(2019)Zintgraf, Shiarli, Kurin, Hofmann, and Whiteson]{CAVIA}
L.~Zintgraf, K.~Shiarli, V.~Kurin, K.~Hofmann, and S.~Whiteson.
\newblock Fast context adaptation via meta-learning.
\newblock In \emph{International Conference on Machine Learning}, pages 7693--7702. PMLR, 2019.

\end{thebibliography}

\clearpage
\appendix

\section{Reproductiblity statement}
Our work is entirely reproducible, and all the references to the information in order to reproduce it are in this section.

\paragraph{Code.} The code for all our experiments is available at \href{https://github.com/EtienneLnr/TimeFlow.git}{this link}.

\paragraph{Data.} A subset of the processed data is available with the code at \href{https://github.com/EtienneLnr/TimeFlow.git}{this link}. The dataset description, processing and normalization are presented in \cref{sec:dataset_prez_norm}.

\paragraph{Model.} The model and the training details are presented in \cref{sec:method} and the hyperparameter selection is available in \cref{sec:hyperparameters}.

\paragraph{GPU.} We used NVIDIA TITAN RTX 24Go single GPU to conduct all the experiments for our method, which is coded in PyTorch (Python 3.9.2).

\section{Architecture details and ablation studies}

\subsection{Architecture details} \label{sec:hyperparameters}

For all imputation and forecasting experiments we choose the following hyperparameters :
\begin{itemize}
    \setlength\itemsep{-0.1em}
    \item $z$ dimension: 128
    \item Number of layers: 5
    \item Hidden layers dimension: 256
    \item $\gamma(t) \in \mathbb{R}^{2 \times 64}$ 
    \item $z$ code learning rate ($\alpha$ in \cref{alg:TimeFlow}): $10^{-2}$
    \item Hypernetwork and INR learning rate: $5 \times 10^{-4}$ 
    \item Number of steps in inner loop: $K=3$
    \item Number of epochs: $4 \times 10^{4}$
    \item Batch size: 64
\end{itemize}

It is worth noting that the hyperparameters mentioned above remain consistent across all experiments conducted in the paper. We chose to maintain a fixed set of hyperparameters for our model, while other imputation and forecasting approaches commonly fine-tune hyperparameters based on a validation dataset. The obtained results exhibit high robustness across various settings, suggesting that the selected hyperparameters are already effective in achieving reliable outcomes.

\subsection{Ablation studies}

\subsubsection{Fourier features vs SIREN on imputation task}
\label{section:fourier-vs-siren}

\paragraph{Baseline} The SIREN network differs from the Fourier features network because it does not explicitly incorporate frequencies as input. Instead, it is a multi-layer perceptron network that utilizes sine activation functions. An adjustable parameter, denoted $\omega_0$, is multiplied with the input matrices of the preceding layers to capture a broader range of frequencies. For this comparison, we adopt the same hyperparameters described in \cref{sec:hyperparameters}, selecting $\omega_0=30$ to align with \cite{SIREN}. Furthermore, we set the learning rate of both the hypernetwork and the INR to $5 \times 10^{-5}$ to enhance training stability. In Table \ref{tab:sirenVSfourier}, we compare the imputation results obtained by the Fourier features network and the SIREN network, specifically focusing on the first time window from the \textit{Electricity}, \textit{Traffic} and \textit{Solar} datasets.

\begin{table}[H]
\caption{MAE imputation errors on the first time window of each dataset. Best results are bold.}
\label{tab:sirenVSfourier}
\begin{center}
\scalebox{0.65}{%
\begin{tabular}{cccc}
\toprule
    & $\tau$ & TimeFlow  & TimeFlow w SIREN\\ 

        \midrule

        \multirow{5}{*}{Electricity} & 0.05 & \textbf{0.323} & 0.466 \\

        & 0.10 & \textbf{0.252} & 0.350 \\

        & 0.20 & \textbf{0.224} & 0.242 \\

        & 0.30 & \textbf{0.211} & 0.222\\

        & 0.50 & \textbf{0.194}& 0.209\\

        \midrule

        \multirow{5}{*}{Solar} & 0.05 & \textbf{0.105} & 0.114 \\

        & 0.10 & \textbf{0.083} & 0.094 \\

        & 0.20 & \textbf{0.065} & 0.079 \\

        & 0.30 & \textbf{0.061} & 0.072 \\

        & 0.50 & \textbf{0.056} & 0.066 \\

        \midrule
        
        \multirow{5}{*}{Traffic} & 0.05 & \textbf{0.292} & 0.333 \\

        & 0.10 & \textbf{0.220} & 0.252 \\

        & 0.20 & \textbf{0.168} & 0.191 \\

        & 0.30 & \textbf{0.152} & 0.163 \\

        & 0.50 & \textbf{0.141} & 0.154 \\

        \bottomrule

\end{tabular}}
\end{center}
\end{table}

\paragraph{Results}{According to the results presented in \cref{tab:sirenVSfourier}, the Fourier features network outperforms the SIREN network in the imputation task on these datasets. Notably, the performance gap between the two network architectures are more pronounced at low sampling rates. This disparity can be attributed to the SIREN network's difficulty in accurately capturing high frequencies when the time series is sparsely observed. We hypothesize that the MLP with ReLU activations correctly learns the different frequencies of time series with multi-temporal patterns by switching on or off the Fourier embedding frequencies. }

\subsubsection{Influence of the latent code dimension}
\label{section:dim-z}

The dimension of the latent code $z$ is a crucial parameter in our architecture. If it is too small, it underfits the time series. Consequently, this adversely affects the performance of both the imputation and forecasting tasks. On the other hand, if the dimension of $z$ is too large, it can lead to overfitting, hindering the model's ability to generalize to new data points.

\paragraph{Baselines}{To investigate the impact of $z$ dimensionality on the performance of TimeFlow, we conducted experiments on the three considered datasets, specifically focusing on the forecasting task. We varied the sizes of $z$ within $\{32, 64, 128, 256\}$. The other hyperparameters are set as presented in \cref{sec:hyperparameters}. The obtained results for each $z$ dimension are summarized in \cref{tab:inference_dim_z}.}

\begin{table}[!ht]
\caption{MAE error for different $z$ dimension.}
\label{tab:inference_dim_z}
\centering
\scalebox{0.65}{%
\begin{tabular}{cccccc}
    \toprule
     & H  &  32 & 64 & 128 & 256 \\
    \midrule

    \multirow{4}{*}{Electricity} & \multirow{1}{*}{96} & 0.232 $\pm$ 0.016 & 0.222 $\pm$ 0.017  & 0.222 $\pm$ 0.018 &  \textbf{0.215 $\pm$ 0.019} \\

     & \multirow{1}{*}{192} & 0.245 $\pm$ 0.020 & 0.239 $\pm$ 0.018 & \textbf{0.230 $\pm$ 0.026} &  0.233 $\pm$ 0.017 \\

     & \multirow{1}{*}{336} & 0.254 $\pm$ 0.029 & 0.244 $\pm$ 0.028 & 0.262 $\pm$ 0.031 & \textbf{0.243 $\pm$ 0.032} \\

    & \multirow{1}{*}{720} & 0.295 $\pm$ 0.027 & 0.284 $\pm$ 0.028 & 0.303 $\pm$ 0.041  &  \textbf{0.283 $\pm$ 0.029} \\

    \cmidrule(r){1-6}

        \multirow{4}{*}{SolarH} & \multirow{1}{*}{96} & 0.182 $\pm$ 0.009 & 0.181 $\pm$ 0.012 & \textbf{0.179 $\pm$ 0.003} & 0.225 $\pm$ 0.047  \\

     & \multirow{1}{*}{192} & 0.195 $\pm$ 0.014 & 0.195 $\pm$ 0.016 & \textbf{0.193 $\pm$ 0.015}  & 0.197 $\pm$ 0.029 \\

     & \multirow{1}{*}{336} & \textbf{0.181 $\pm$ 0.011} & 0.182 $\pm$ 0.011 & 0.189 $\pm$ 0.013 & 0.183 $\pm$ 0.012 \\

     & \multirow{1}{*}{720} & 0.201 $\pm$ 0.027 & \textbf{0.199 $\pm$ 0.025} & 0.209 $\pm$ 0.029 & 0.200 $\pm$ 0.030 \\

    \cmidrule(r){1-6}

    \multirow{4}{*}{Traffic} & \multirow{1}{*}{96} & 0.223 $\pm$ 0.024 & 0.215 $\pm$ 0.028 & 0.215 $\pm$ 0.037 &  \textbf{0.210 $\pm$ 0.033} \\

     & \multirow{1}{*}{192} & 0.214 $\pm$ 0.018 & 0.217 $\pm$ 0.025 & 0.206 $\pm$ 0.023 & \textbf{0.203 $\pm$ 0.024} \\

     & \multirow{1}{*}{336} & 0.238 $\pm$ 0.029 & 0.231 $\pm$ 0.029 & \textbf{0.226 $\pm$ 0.030} & 0.229 $\pm$ 0.029 \\

     & \multirow{1}{*}{720} & 0.272 $\pm$ 0.040 & 0.269 $\pm$ 0.035 & \textbf{0.259 $\pm$ 0.038}  & 0.262 $\pm$ 0.040 \\

    \bottomrule

\end{tabular}
}
\end{table}

\paragraph{Results}{The results presented in \cref{tab:inference_dim_z} suggest that a $z$ dimension of 128 is a reasonable compromise but only optimal for some settings. Moreover, even though the choice of $z$ dimension seems important, it doesn't critically impact the MAE error for the forecasting task.}

\subsubsection{Influence of the number of gradient steps}
\label{section:gradient_steps}

As can be seen in \cref{tab:inference_time_gradient_steps}, using three gradient steps at inference yield an inference time of less than 0.2 seconds. The latter can still be reduced by doing only one step at the cost of an increase in the forecasting error. As observed in \cref{tab:inference_time_gradient_steps}, increasing the number of gradient steps above 3 steps during inference does not improve forecasting performance.

\begin{table}[!ht]
\caption{Inference time (in seconds) and MAE error on the forecasting task on the \textit{Electricity} dataset for a horizon of length 720, a look-back window of length 512, and a varying number of adaptation gradient steps. The statistics are computed over 10 runs using an NVIDIA TITAN RTX GPU.}
\label{tab:inference_time_gradient_steps}
\centering
\scalebox{0.65}{%
\begin{tabular}{ccccccc}
    \toprule
    Gradient descent steps & 1 & 3 & 10 & 50 & 500 & 5000 \\
    \midrule

    Inference time (s) & 0.109 $\pm$ 0.003 &  0.176 $\pm$ 0.009 & 0.427  $\pm$ 0.031 & 3.547  $\pm$ 0.135 & 17.722 $\pm$ 0.536 & 189.487 $\pm$ 8.060 \\
    
    MAE & 0.351 $\pm$ 0.038 &  0.303 $\pm$ 0.041 &  0.300 $\pm$ 0.040  &  0.299 $\pm$ 0.039 &   0.302 $\pm$ 0.038 & 0.308  $\pm$ 0.037 \\

    \bottomrule

\end{tabular}
}
\end{table}

\begin{table}[!ht]
\caption{MAE error on the forecasting task using 1 inner-step during training and a varying number of adaptation gradient steps at inference. Best results are in bold and $/$ symbol means that the MAE score is very high ($\geq 1$).}
\label{tab:inference_score_gradient_steps_1}
\centering
\scalebox{0.65}{%
\begin{tabular}{cccccc}
    \toprule
     & H  &  1 & 3 & 10 & 50 \\
    \midrule

    \multirow{4}{*}{Electricity} & \multirow{1}{*}{96} & \textbf{0.244 $\pm$ 0.017} & 0.246 $\pm$ 0.017 & 0.261 $\pm$ 0.016 & / \\

     & \multirow{1}{*}{192} & \textbf{0.253 $\pm$ 0.024} & \textbf{0.253 $\pm$ 0.022} & 0.261 $\pm$ 0.020 & 0.265 $\pm$ 0.019 \\

     & \multirow{1}{*}{336} & \textbf{0.267 $\pm$ 0.032} & 0.268 $\pm$ 0.030 & 0.277 $\pm$ 0.028 & 0.281 $\pm$ 0.027\\

    & \multirow{1}{*}{720} & 0.302 $\pm$ 0.030 & 0.306 $\pm$ 0.029 & 0.310 $\pm$ 0.028 & \textbf{0.301 $\pm$ 0.029} \\

    \cmidrule(r){1-6}

        \multirow{4}{*}{SolarH} & \multirow{1}{*}{96} & \textbf{0.192 $\pm$ 0.023} & 0.623 $\pm$ 0.397 & / & /\\

     & \multirow{1}{*}{192} & \textbf{0.175 $\pm$ 0.006} & 0.252 $\pm$ 0.068 & / & / \\

     & \multirow{1}{*}{336} & \textbf{0.192 $\pm$ 0.016} & 0.471 $\pm$ 0.029 & / & / \\

     & \multirow{1}{*}{720} & \textbf{0.216 $\pm$ 0.034}  & 0.465 $\pm$ 0.063 & / & 0.550 $\pm$ 0.187 \\

    \cmidrule(r){1-6}

    \multirow{4}{*}{Traffic} & \multirow{1}{*}{96} & \textbf{0.215 $\pm$ 0.029} & 0.329 $\pm$ 0.039 & / & / \\

     & \multirow{1}{*}{192} & \textbf{0.208 $\pm$ 0.019} & 0.310 $\pm$ 0.033 & 0.312 $\pm$ 0.032 & / \\

     & \multirow{1}{*}{336} & \textbf{0.237 $\pm$ 0.028} & 0.307 $\pm$ 0.038 & / & / \\

     & \multirow{1}{*}{720} & \textbf{0.263 $\pm$ 0.038} & 0.320 $\pm$ 0.040 & / & / \\

    \bottomrule

\end{tabular}
}
\end{table}

\begin{table}[!ht]
\caption{MAE error on the forecasting task using 3 inner-steps during training and a varying number of adaptation gradient steps at inference. Best results are in bold.}
\label{tab:inference_score_gradient_steps}
\centering
\scalebox{0.65}{%
\begin{tabular}{ccccccc}
    \toprule
     & H  &  & 1 & 3 & 10 & 50\\
    \midrule

    \multirow{4}{*}{Electricity} & \multirow{1}{*}{96} & & 0.259 $\pm$ 0.020 &  \textbf{0.222 $\pm$ 0.018} &  \textbf{0.222 $\pm$ 0.017}  &   0.228 $\pm$ 0.019 \\

     & \multirow{1}{*}{192} & & 0.269 $\pm$ 0.020 &  \textbf{0.230 $\pm$ 0.026}  & 0.232 $\pm$ 0.020  &   0.233 $\pm$ 0.026 \\

     & \multirow{1}{*}{336} &  & 0.273 $\pm$ 0.033 & \textbf{0.262 $\pm$ 0.031}  &  0.264 $\pm$ 0.032  &  0.268 $\pm$ 0.032   \\

    & \multirow{1}{*}{720} & & 0.351 $\pm$ 0.038 &  0.303 $\pm$ 0.041 &  0.300 $\pm$ 0.040  &  \textbf{0.299 $\pm$ 0.039} \\

    \cmidrule(r){1-7}

        \multirow{4}{*}{SolarH} & \multirow{1}{*}{96} &  &  0.487 $\pm$ 0.196  &   \textbf{0.179 $\pm$ 0.003}  &  0.181 $\pm$ 0.003 & 0.186 $\pm$ 0.003 \\

     & \multirow{1}{*}{192} &  &  0.411 $\pm$ 0.088  & \textbf{0.193 $\pm$ 0.015} &   0.195 $\pm$ 0.014  &  0.199 $\pm$ 0.013 \\

     & \multirow{1}{*}{336} & &  0.435 $\pm$ 0.153  &  \textbf{0.189 $\pm$ 0.013} &   0.203 $\pm$ 0.006 &  0.223 $\pm$ 0.012 \\

     & \multirow{1}{*}{720} &   &  0.394 $\pm$ 0.173  &   0.209 $\pm$ 0.029  &   \textbf{0.203 $\pm$ 0.006}  &  0.209 $\pm$ 0.027  \\

    \cmidrule(r){1-7}

    \multirow{4}{*}{Traffic} & \multirow{1}{*}{96} & &  0.320 $\pm$ 0.038  &   \textbf{0.215 $\pm$ 0.037}  &   0.219 $\pm$ 0.043  &  0.226 $\pm$ 0.046  \\

     & \multirow{1}{*}{192} &  &  0.299 $\pm$ 0.023 & \textbf{0.206 $\pm$ 0.023}  & 0.209 $\pm$ 0.026 & 0.214 $\pm$ 0.027\\

     & \multirow{1}{*}{336} &  &  0.345 $\pm$ 0.038  &  \textbf{0.226 $\pm$ 0.030} &   0.228 $\pm$ 0.031 & 0.233 $\pm$ 0.032\\

     & \multirow{1}{*}{720} &  &  0.321 $\pm$ 0.034  &  \textbf{0.259 $\pm$ 0.038}  &   0.260 $\pm$ 0.038 &  0.266 $\pm$ 0.039 \\

    \bottomrule

\end{tabular}
}
\end{table}

\begin{table}[!ht]
\caption{MAE error on the forecasting task using 10 inner-steps during training and a varying number of adaptation gradient steps at inference. Best results are in bold.}
\label{tab:inference_score_gradient_steps_10}
\centering
\scalebox{0.65}{%
\begin{tabular}{ccccccc}
    \toprule
     & H  & 1 & 3 & 10 & 50\\
    \midrule

    \multirow{4}{*}{Electricity} & \multirow{1}{*}{96} & 0.381 $\pm$ 0.030 & 0.249 $\pm$ 0.024 & \textbf{0.236 $\pm$ 0.024} & 0.238 $\pm$ 0.024\\

     & \multirow{1}{*}{192} & 0.448 $\pm$ 0.045 & 0.273 $\pm$ 0.019 & \textbf{0.244 $\pm$ 0.014} & \textbf{0.244 $\pm$ 0.013}\\

     & \multirow{1}{*}{336} & 0.514 $\pm$ 0.053 & 0.283 $\pm$ 0.033 & \textbf{0.241 $\pm$ 0.025} & 0.242 $\pm$ 0.024 \\

    & \multirow{1}{*}{720} & 0.647 $\pm$ 0.068 & 0.400 $\pm$ 0.051 & \textbf{0.286 $\pm$ 0.023} & 0.287 $\pm$ 0.021\\

    \cmidrule(r){1-7}

        \multirow{4}{*}{SolarH} & \multirow{1}{*}{96} & 0.605 $\pm$ 0.029 & 0.380 $\pm$ 0.018 & \textbf{0.188 $\pm$ 0.012} & 0.199 $\pm$ 0.015\\

     & \multirow{1}{*}{192} & 0.382 $\pm$ 0.072 & 0.250 $\pm$ 0.012 & \textbf{0.202 $\pm$ 0.034} & 0.204 $\pm$ 0.035\\

     & \multirow{1}{*}{336} & 0.745 $\pm$ 0.105 & 0.431 $\pm$ 0.221 & \textbf{0.201 $\pm$ 0.033} & 0.208 $\pm$ 0.032 \\

     & \multirow{1}{*}{720} & 0.745 $\pm$ 0.082  & 0.477 $\pm$ 0.039 & \textbf{0.205 $\pm$ 0.030} & \textbf{0.205 $\pm$ 0.029} \\

    \cmidrule(r){1-7}

    \multirow{4}{*}{Traffic} & \multirow{1}{*}{96} & 0.450 $\pm$ 0.023 & 0.273 $\pm$ 0.026 & \textbf{0.225 $\pm$ 0.028} & 0.230 $\pm$ 0.034 &\\

     & \multirow{1}{*}{192} & 0.506 $\pm$ 0.028 & 0.318 $\pm$ 0.021 & \textbf{0.233 $\pm$ 0.022} & 0.236 $\pm$ 0.026 & \\

     & \multirow{1}{*}{336} & 0.500 $\pm$ 0.042 & 0.320 $\pm$ 0.021 & \textbf{0.247 $\pm$ 0.028} & 0.249 $\pm$ 0.031 \\

     & \multirow{1}{*}{720} & 0.511 $\pm$ 0.035 & 0.323 $\pm$ 0.022 & \textbf{0.266 $\pm$ 0.027} & 0.272 $\pm$ 0.024 \\

    \bottomrule

\end{tabular}
}
\end{table}

\paragraph{Results}{We conduct more extensive experiments in \cref{tab:inference_score_gradient_steps_1}, \cref{tab:inference_score_gradient_steps}, \cref{tab:inference_score_gradient_steps_10} to quantify the MAE score variation according to different number of gradient steps during training and inference. The tables show that using the same number of steps in training and inference leads to better results. This is expected since using different gradient steps for training and inference makes the inference model slightly different from the training model. In addition, using 3 gradient steps instead of 1 clearly improves the performances, but using 10 instead of 3 does not. Indeed, it usually leads to overall better results for longer horizon, but the gain is not clear for smaller horizons. Hence using 3 gradient steps is a suitable choice.}

\subsubsection{TimeFlow variants with other meta-learning techniques}

\label{section:timeflow-other-metalearning}

\paragraph{Baselines} Before converging to the current architecture and optimization of TimeFlow, we explored different options to condition the INR with the observations. The first one was inspired by the neural process architecture, which uses a set encoder to transform a set of observations $(t_i, x_{t_i})_{i \in \mathcal{I}}$ into a latent code $z$ by applying a pooling layer after a feed forward network. We observed that this encoder in combination with the modulated fourier features network was able to achieve relatively good results on the forecasting task but suffered of underfitting on more complex datasets such as \textit{Electricity}. 

This led us to consider auto-decoding methods instead, \ie encoder-less, architectures for conditioning the weights of the coordinate-based network. We trained TimeFlow with the REPTILE algorithm \citep{nichol2018first}, which is a first-order meta-learning technique that adapts the code in a few steps of gradient descent. In contrast with a second-order method, we observed that REPTILE was less costly to train but struggled to escape sub optimal minima, which led to unstable training and underfitting. 

From an implementation point of view, the only difference between second order and first order, is that in the latter the code is detached from the computation graph before taking the outer-loop parameter update. When the code is not detached, it remains a function of the common parameters $z = z(\theta, w)$, which means that the computation graph for the outer-loop also includes the inner-loop updates to the codes. Therefore the outer-loop gradient update involves a gradient through a gradient and requires an additional backward pass through the INR to compute the Hessian. Please refer to \cite{finn2017model} for more technical details.

\begin{table}[H]
\setlength{\tabcolsep}{4pt}
\caption{Comparison of second-order and first-order (REPTILE) meta learning for TimeFlow on the imputation task. Mean MAE results on the missing grid over five different time windows. $\tau$ stands for the subsampling rate. Bold results are best.}

\label{tab:impu_abla_first_order_meta_learning}
\begin{center}
\scalebox{0.60}{
\begin{tabular}{cccccccccc}
\toprule
  & $\tau$ &  TimeFlow & TimeFlow w REPTILE \\
\midrule

        & 0.05 &  \textbf{0.324 $\pm$ 0.013}    &   0.363 $\pm$ 0.062  \\

        & 0.10 &  \textbf{0.250 $\pm$ 0.010}    &  0.343 $\pm$ 0.036  \\

        Electricity & 0.20 &  \textbf{0.225 $\pm$ 0.008}    &   0.312 $\pm$ 0.043  \\

        & 0.30 &  \textbf{0.212 $\pm$ 0.007}    &    0.308 $\pm$ 0.035 \\ 

        & 0.50 &  \textbf{0.194 $\pm$ 0.007}   &  0.305 $\pm$  0.046\\ 
        
        \midrule

        & 0.05 &  \textbf{0.095 $\pm$ 0.015}    &  0.125 $\pm$ 0.025   \\

        & 0.10 &  \textbf{0.083 $\pm$ 0.015}    &  0.123 $\pm$ 0.032  \\

        Solar & 0.20 &  \textbf{0.072 $\pm$ 0.015}   &   0.108 $\pm$ 0.021 \\

        & 0.30 &   \textbf{0.061 $\pm$ 0.012}   &   0.105 $\pm$ 0.027 \\

        & 0.50 &   \textbf{0.054 $\pm$ 0.013}   &  0.102 $\pm$ 0.021  \\

        \midrule

        & 0.05 &  \textbf{0.283 $\pm$ 0.016}    &  0.304 $\pm$ 0.026 \\

        & 0.10 &  \textbf{0.211 $\pm$ 0.012}    &  0.264 $\pm$ 0.009 \\

        Traffic & 0.20 &  \textbf{0.168 $\pm$ 0.006}    &   0.242 $\pm$  0.019  \\

        & 0.30 &  \textbf{0.151 $\pm$ 0.007}    & 0.218 $\pm$ 0.020 \\

        & 0.50 &  \textbf{0.139 $\pm$ 0.007}    &  0.216 $\pm$ 0.017 \\ 
        \bottomrule
\end{tabular}}
\end{center}
\end{table}

\paragraph{Results} In \Cref{tab:impu_abla_first_order_meta_learning}, we show the performance of first-order TimeFlow on the imputation task. In low sampling regimes the difference with TimeFlow is less perceptive, but its performance plateaus when the number of points increases. This is not surprising. Indeed, as though the task is actually simpler when $\tau$ increases, the optimization is made more difficult with the increased number of observations. We provide the performance of TimeFlow with a set encoder on the Forecasting task in \Cref{tab:forecast_abla_set_encoder}. We observed that this version failed to generalize well for complex datasets.

\begin{table}[H]
\caption{Comparison of optimization-based and set-encoder-based meta learning for TimeFlow on the forecasting task. Mean MAE forecast results over different time windows. $H$ stands for the horizon. Bold results are best.}

\label{tab:forecast_abla_set_encoder}
\begin{center}
\scalebox{0.65}{
\begin{tabular}{cccccccccc}
\toprule
  & $H$ &  TimeFlow & TimeFlow w set encoder \\
\midrule

        & 96 &  \textbf{0.228 $\pm$ 0.026}    &  0.362 $\pm$ 0.032  \\

        & 192 &  \textbf{0.238 $\pm$ 0.020}    &  0.360  $\pm$ 0.028   \\

        Electricity & 336 &  \textbf{0.270 $\pm$ 0.031}  & 0.382    $\pm$ 0.038 \\

        & 720 &  \textbf{0.316 $\pm$ 0.055}    &   0.431 $\pm$ 0.059 \\ 
        
        \midrule

        & 96 &  \textbf{0.190 $\pm$ 0.013}    &  0.251 $\pm$  0.071 \\

        & 192 &  \textbf{0.202 $\pm$ 0.020}    &  0.239 $\pm$ 0.058\\

        SolarH & 336 &  \textbf{0.209 $\pm$ 0.017}   &  0.235  $\pm$ 0.040 \\

        & 720 &   \textbf{0.218 $\pm$ 0.048}   &  0.231 $\pm$ 0.032\\

        \midrule

        & 96 &  \textbf{0.217 $\pm$ 0.036}    &  0.276 $\pm$ 0.031 \\

        & 192 &  \textbf{0.212 $\pm$ 0.028}    & 0.281 $\pm$ 0.034 \\

        Traffic & 336 &  \textbf{0.238 $\pm$ 0.034}    & 0.297  $\pm$  0.042 \\

        & 720 &  \textbf{0.279 $\pm$ 0.050}    &  0.333 $\pm$ 0.048 \\
        \bottomrule

\end{tabular}}
\end{center}
\end{table}

\subsubsection{Influence of the modulation}
\label{section:modulation-choice}
In TimeFlow, we apply shift modulations to the parameters of the INR, \ie for each layer $l$ we only modify the biases of the network with an extra bias term $\phi_l^{(j)}$. We generate these bias terms with a linear hypernetwork that maps the code $z^{(j)}$ to the modulations.  The output of the $l$-th layer of the modulated INR is thus given by $\phi_{l+1} = \text{ReLU}(\theta_l \phi_{l-1} + b_l + \psi_l^{(j)})$, where $\psi_l^{(j)} = W_l z^{(j)}$ and $(W_l)_{l=1}^{L}$ are parameters of the hypernetwork. However, another common modulation is the combination of the scale and shift modulation, which leads to the output of the $l$-th layer of the modulated INR being given by 
$\phi_{l+1} = \text{ReLU}((S_l z^{(j)}) \circ (\theta_l \phi_{l-1} + b_l) + \psi_l^{(j)})$, where $\psi_l^{(j)} = W_l z^{(j)}$, and $(W_l)_{l=1}^{L}$ and $(S_l)_{l=1}^{L}$ are parameters of the hypernetwork and $\circ$ is the Hadamard product.

In \cref{tab:ablation_modulations}, we conduct additional experiments on the \textit{Electricity} dataset in the forecasting setting with different time horizons. In these experiments, we compare two scenarios: one where the INR is modulated only by a shift factor and the other where the INR is modulated by both a shift and a scale factor. We kept the architecture and hyperparameters consistent with those described in \Cref{sec:hyperparameters}. The experiments shown in \cref{tab:ablation_modulations} indicate that the INR is longer to train with shift and scale modulations due to the increased number of parameters involved. Furthermore, we observe that the shift and scale modulated INR performed similarly or even worse than the INR with only shift modulation. These two drawbacks, namely an increased computational time and similar or worse performances, motivate modulating the INR only by a shift factor.

\begin{table}[H]
\setlength{\tabcolsep}{4pt}
\caption{Ablation on modulations for the forecasting task on \textit{Electricity} dataset for different horizons. Models are trained on a given time window and tested on four new time windows. Models are trained on a single NVIDIA TITAN RTX GPU.}
\label{tab:ablation_modulations}
\begin{center}
\scalebox{0.65}{%
\begin{tabular}{ccccccccc}
    \toprule
     & \multicolumn{2}{c}{96} & \multicolumn{2}{c}{192} & \multicolumn{2}{c}{336} & \multicolumn{2}{c}{720} \\
     \cmidrule(r){2-3} \cmidrule(r){4-5}\cmidrule(r){6-7} \cmidrule(r){8-9}
     & \begin{tabular}[x]{@{}c@{}}MAE\end{tabular} & \begin{tabular}[x]{@{}c@{}}Training time\end{tabular} & \begin{tabular}[x]{@{}c@{}}MAE\end{tabular} & \begin{tabular}[x]{@{}c@{}}Training time\end{tabular} & \begin{tabular}[x]{@{}c@{}}MAE\end{tabular} & \begin{tabular}[x]{@{}c@{}}Training time\end{tabular} & \begin{tabular}[x]{@{}c@{}}MAE\end{tabular} & \begin{tabular}[x]{@{}c@{}}Training time\end{tabular} \\
    \midrule

    \begin{tabular}[x]{@{}c@{}}Shift\end{tabular} & 0.233 $\pm$ 0.014 & 2h30 & 0.245 $\pm$ 0.016 & 2h31 & 0.264 $\pm$ 0.020 & 2h33 & 0.303 $\pm$ 0.041 & 2h46  \\

    \midrule
    
    \begin{tabular}[x]{@{}c@{}}Shift and scale\end{tabular} & 0.257 $\pm$ 0.019 & 3h29 & 0.263 $\pm$ 0.014 & 3h32 & 0.268 $\pm$ 0.025 & 3h45 & 0.308 $\pm$ 0.037 & 4h14\\

    \bottomrule

\end{tabular}}
\end{center}
\end{table}

\subsubsection{Discussion on other hyperparameters}

While the dimension of $z$ is indeed a crucial hyperparameter, it is important to note that other hyperparameters also play a significant role in the performance of the INR. For example, the number of layers in the FFN directly affects the ability of the model to fit the time series. In our experiments, we have observed that using five or more layers yields good performance, and including additional layers can lead to slight improvements in the generalization settings.

Similarly, the number of frequencies used in the frequency embedding is another important hyperparameter. Using too few frequencies can limit the network's ability to capture patterns, while using too many frequencies can hinder its ability to generalize accurately.

The choice of learning rate is critical for achieving stable convergence during training. 
Therefore, in practice, we use a low learning rate combined with a cosine annealing scheduler to ensure stable and effective training.

\section{Datasets, scores and normalization} \label{sec:dataset_prez_norm}

For the complete datasets, \textit{Electricity} dataset is available \href{https://archive.ics.uci.edu/dataset/321/electricityloaddiagrams20112014}{here}, \textit{Traffic} dataset \href{https://pems.dot.ca.gov/}{here} and \textit{Solar} data set \href{https://zenodo.org/record/3889974}{here}. \cref{tab:datasets infos} provides a concise overview of the main information about the datasets used for forecasting and imputation tasks.

\begin{table}[H]
\renewcommand*{\arraystretch}{1.0}
\setlength{\tabcolsep}{3pt}
\caption{Summary of datasets information}
\label{tab:datasets infos}
\begin{center}
\scalebox{0.65}{%
\begin{tabular}{cccccc}
\toprule
    Dataset name & Number of samples & Number of time steps & Sampling frequency & Location & Years\\
        \midrule
    Electricity &  321               & 26 304  & hourly     & Portugal          & $2012-2014$ \\
    Traffic     &  862               & 17 544  & hourly     & San Francisco bay & $2015-2016$ \\
    Solar       &  137               & 52 560  & 10 minutes & Alabama           & $2006$\\
    SolarH      &  137               & 8 760   & hourly     & Alabama           & $2006$\\
        
        \bottomrule

\end{tabular}}
\end{center}
\end{table}

\paragraph{How TimeFlow relative improvement score is computed?}{In many paper tables, the TimeFlow improvement score appears in the last row of the table. Its purpose is to quantify the average marginal gain of TimeFlow over the method under consideration. It is computed as follows:}
\begin{equation*}
    \text{TimeFlow improvement} = \frac{1}{L} \sum_{l=1}^{L} \frac{{s}_{l}(\text{baseline})  - {s}_{l}(\text{TimeFlow})}{{s}_{l}(\text{baseline})}
\end{equation*}
\begin{itemize}
    \item s stands for the Mean Absolute Error score of the considered method against the ground truth at line l (for instance in \cref{tab:impu_classic}, ${s}_{1}$(TimeFlow)$=0.324$, ${s}_{2}$(TimeFlow)$=0.250$ etc).
    \item L stands for the number of line in the table (for instance 15 in \cref{tab:impu_classic}).
\end{itemize}

\paragraph{z-normalization}{To preprocess each dataset, we apply the widely used z-normalization technique per-sample $j$ on the entire series:
$x^{(j)}_{norm} = \frac{x^{(j)} - \text{mean}(x^{(j)})}{\text{std}(x^{(j)})}.$ 
}

\section{Imputation experiments}

\subsection{Baselines details}

\subsubsection{Baselines training and hyperparameters}
\label{baseline-training-sec}

The baselines underwent meticulous training and extensive testing, involving thorough exploration of hyperparameters. 
We used SAITS repository (\href{https://github.com/WenjieDu/SAITS}{code}) for BRITS and SAITS. The adopted setting results from a hyperparameters search, which yields marginally superior results for both methods compared to the default settings. The marginal difference in scores underscores the robustness of BRITS and SAITS.
In addition, mTAN and TIDER did not perform optimally with the recommended configurations, requiring an extensive search. Details of the parameters explored are provided in \Cref{tab:mtan_param_search} and \Cref{tab:tider_param_search}. While the hyperparameter search led to performance improvements, overall results remained sub-optimal.
For CSDI, the recommended settings proved inadequate for the considered datasets, prompting a comprehensive search. Among various parameters, the number of diffusion steps emerged as crucial, significantly enhancing performance, particularly at higher draw ratios. However, superior performance was attained with more diffusion steps, albeit at increased computational cost. The chosen parameters are detailed in \Cref{tab:CSDI_param}. 
In addition, the original DeepTime implementation did not perform well on imputation. Hence, we adapted the DeepTime training procedure (see \cref{sec:deeptime-adaptation}). Lastly, the vanilla Neural Process baseline underperformed, so we customized its architecture to conduct a fair comparison with TimeFlow. We used the INR and hypernetwork from TimeFlow to align the Neural Process with our temporal frequency bias and shift modulation technique. 

\begin{table}[h!]
    \centering
    \caption{mTAN hyperparameter search.}
    \label{tab:mtan_param_search}
    \begin{center}
        \scalebox{0.65}{
            \begin{tabular}{cccccccc}
                \toprule
                & Dimension size & $\gamma$ linear scheduler & lr & NumRefPoints & k-iwae & Target ratio \\
                \midrule
                 & 50 & 1 & $1 \times 10^{-5}$ & 32 & 5 & 0.2 \\
                 & 100 & 0.95 & 0.0001 & 64 & 10 & 0.8 \\
                 & - & 0.5 & 0.001 & 128 & - & - \\
                 & - & 0.1 & 0.005 & - & - & - \\
                \bottomrule
            \end{tabular}}
    \end{center}
\end{table}

\begin{table}[h!]
    \centering
    \caption{TIDER hyperparameter search.}
    \label{tab:tider_param_search}
    \begin{center}
        \scalebox{0.65}{
            \begin{tabular}{cccccccc}
                \toprule
                & Dimension size & $\lambda_{ar}$ & $\lambda_{trend}$ & lr & Season number \\
                \midrule
                 & 50 & 0.1 & 0.01 & 0.0001 & 2 \\
                 & 100 & 0.2 & 0.1 & 0.001 & 10 \\
                 & - & - & - & 0.005 & 15 \\
                 & - & - & - & - & 20 \\
                \bottomrule
            \end{tabular}
            }
    \end{center}
\end{table}

\begin{table}[h!]
    \centering
    \caption{CSDI chosen hyperparameters.}
    \label{tab:CSDI_param}
    \begin{center}
        \scalebox{0.65}{
            \begin{tabular}{ccccccccccc}
                \toprule
                & Epochs & lr & Layers & Channels & Nheads & Diffusion embedding dimension & NSteps & Schedule & Time embedding & Feature embedding \\
                \midrule
                 & 5000 & 0.001 & 4 & 64 & 8 & 128 & 100 & Quad & 128 & 16 \\
                \bottomrule
            \end{tabular}
            }
    \end{center}
\end{table}

\subsubsection{Models complexity}

We can see in \cref{tab:weight_comparison_imputation} that our method has 10 times less parameters than BRITS and 20 times less than SAITS. It is mainly due to their modelisation of interaction between samples. SAITS, which is based on transformers has the highest number of parameters when mTAN has the lowest number of parameters.

\begin{table}[h!]

\caption{Number of parameters for each DL methods on the imputation task on the \textit{Electricity} dataset. }
\label{tab:weight_comparison_imputation}
\begin{center}
\scalebox{0.65}{
\begin{tabular}{cccccccc}
    \toprule
    & TimeFlow & DeepTime & NeuralProcess & mTAN & SAITS & BRITS & TIDER \\
    \midrule

    Number of parameters & 602k & 1315k & 248k & 113k &11 137k  & 6 220k  & 1 034k  \\



    \bottomrule

\end{tabular}}
\end{center}
\end{table}

\subsection{Imputation for previously unseen time series}
\label{imputation_unseen_series}

\paragraph{Setting}{In this section we analyze in details the imputations results for previously unseen time series described in \cref{sec:imputation}. Specifically, TimeFlow is trained on a given set of time series within a defined time window and then used for inference on new time series. We train TimeFlow on 50 \% of the samples and consider the remaining 50 \% as the new time series.

We compare in \cref{tab:imputation_gene_train_test} observed grid fit scores and missing grid inference scores for time series known at training and time series unknown at training.}

\begin{table}[h!]
\caption{TimeFlow MAE imputation errors results for imputation previsouly unseen time series.}
\label{tab:imputation_gene_train_test}
\begin{center}
\scalebox{0.65}{
\begin{tabular}{cccccc}
\toprule
  &        &  \multicolumn{2}{c}{Known time series} & \multicolumn{2}{c}{New time series}  \\
  
  \cmidrule(r){3-4} \cmidrule(r){5-6}   
  
  & $\tau$ & Fit  & Inference  & Fit & Inference \\
        \midrule
        \multirow{5}{*}{Electricity} & 0.05 & 0.060 $\pm$ 0.010 & 0.402 $\pm$ 0.021 & 0.142 $\pm$ 0.083 & 0.413 $\pm$ 0.026 \\

        & 0.10 & 0.046 $\pm$ 0.006 & 0.302 $\pm$ 0.010 & 0.144 $\pm$ 0.098 & 0.309 $\pm$ 0.016 \\

        & 0.20 & 0.067 $\pm$ 0.015 & 0.285 $\pm$ 0.014 & 0.154 $\pm$ 0.089 & 0.291 $\pm$ 0.022 \\

        & 0.30 & 0.093 $\pm$ 0.022 & 0.266 $\pm$ 0.010 & 0.163 $\pm$ 0.073 & 0.271 $\pm$ 0.017 \\

        & 0.50 & 0.108 $\pm$ 0.012 & 0.236 $\pm$ 0.010 & 0.167 $\pm$ 0.061 & 0.245 $\pm$ 0.017\\

        \midrule
        \multirow{5}{*}{Solar}& 0.05 & 0.014 $\pm$ 0.002 & 0.104 $\pm$ 0.015 & 0.050 $\pm$ 0.037 & 0.109 $\pm$ 0.016 \\

        & 0.10 & 0.017 $\pm$ 0.002 & 0.092 $\pm$ 0.015 & 0.052 $\pm$ 0.036 & 0.099 $\pm$ 0.017\\

        & 0.20 & 0.028 $\pm$ 0.008 & 0.078 $\pm$ 0.014 & 0.058 $\pm$ 0.031 & 0.089 $\pm$ 0.017 \\

        & 0.30 & 0.038 $\pm$ 0.009 & 0.072 $\pm$ 0.013 & 0.063 $\pm$ 0.028 & 0.084 $\pm$ 0.018 \\

        & 0.50 & 0.045 $\pm$ 0.011 & 0.066 $\pm$ 0.013 & 0.067 $\pm$ 0.025 & 0.080 $\pm$ 0.019 \\

        \midrule
        \multirow{5}{*}{Traffic}& 0.05 & 0.044 $\pm$ 0.003 & 0.291 $\pm$ 0.013 & 094 $\pm$ 0.051 & 0.291 $\pm$ 0.012\\

        & 0.10 & 0.033 $\pm$ 0.001 & 0.209 $\pm$ 0.010 & 0.093 $\pm$ 0.060 & 0.216 $\pm$ 0.012 \\

        & 0.20 & 0.037 $\pm$ 0.006 & 0.175 $\pm$ 0.008 & 0.095 $\pm$ 0.058 & 0.186 $\pm$ 0.013 \\

        & 0.30 & 0.048 $\pm$ 0.005 & 0.164 $\pm$ 0.006 & 0.098 $\pm$ 0.051 & 0.175 $\pm$ 0.013 \\

        & 0.50 & 0.068 $\pm$ 0.004 & 0.159 $\pm$ 0.007 & 0.110 $\pm$ 0.042 & 0.169 $\pm$ 0.012 \\
    
        \bottomrule

\end{tabular}}
\end{center}
\end{table}

\paragraph{Results}{The results presented in \cref{tab:imputation_gene_train_test} indicate that the inference MAE for missing grids shows consistency between known and new samples, regardless of the data or sampling rate. 
However, it is worth noting that there is a slight drop in performance compared to the results in table \cref{tab:impu_classic}. This decrease is because in \cref{tab:imputation_gene_train_test}, the shared architecture is trained on only half the samples, affecting its overall performance.}

\subsection{Details on DeepTime adaptation for imputation}
\label{sec:deeptime-adaptation}

As DeepTime was proposed to address the forecasting task with a deeptime-index model, the authors did not tackle the task of imputation and left it out for future work. Given the success of this method and the motivation of our work, we wanted to explore its capabilities to impute time series with several subsampling rates. Following our current framework, we first tried to train the model in a self-supervised way, \ie trying to reconstruct observations $x^{(j)} \in \mathcal{T}^{(j)}$ after the INR has been conditioned with the Ridge Regressor on the same set of observations, but discovered failure cases for $\tau \leq 0.20$. To be faithful to the original supervised training of DeepTime, we therefore randomly mask out 50\% of the observations that we use as context for the Ridge Regressor and try to infer the other 50\% (the targets) to train the INR. 

\begin{figure}[!htb]
    \centering    \includegraphics[width=0.98\linewidth]{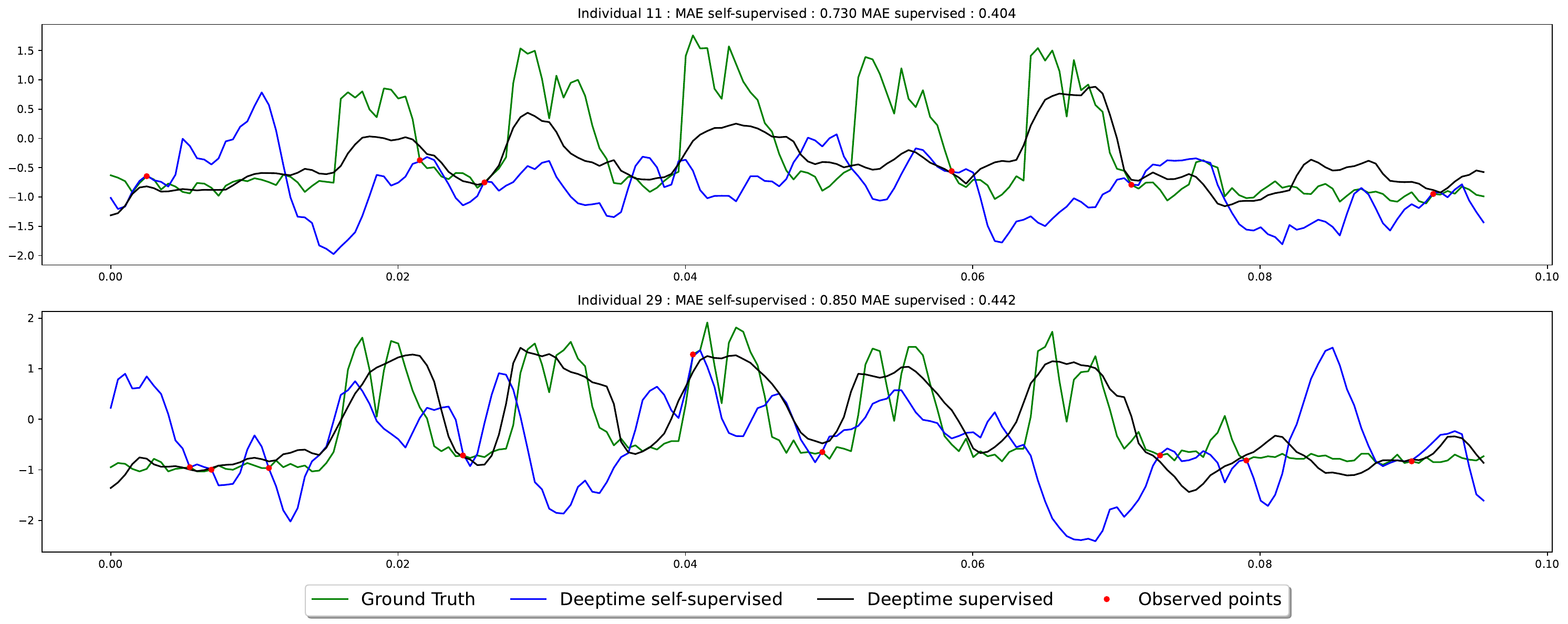}
    \caption{\textit{Electricity dataset}. Self supervised DeepTime imputation (blue line) and supervised DeepTime imputation (black line) with 5$\%$ of known point (red points) on the eight first days of samples 11 (top) and 29 (bottom).}
    \label{fig:deeptime-supervised-vs-self-supervised}
\end{figure}

We provide a qualitative comparison of the model's performance with these two different training procedures in \Cref{fig:deeptime-supervised-vs-self-supervised}. We can notice that the model that results from the self-supervised training perfectly fits the observations but completely misses the important patterns of the series. On the other hand, when DeepTime is trained to infer target values based on observations, it is able to capture the general trends. We think that in the small subsampling regime ($\tau \leq 0.20$), the Ridge Regressor easily fits very well all the observations which hinders the training of the INR's basis. 

\subsection{Imputation against non deep learning methods}
\label{sec:extra_results_imputation}

\paragraph{Setting}{In addition to the deep learning methods presented in \cref{tab:impu_classic}, we evalute TimeFlow against two classic machine learning baselines, K-Nearest Neighbours (KNN) and linear interpolation, which are valuable for getting an idea of the complexity of the problem.}

\begin{table}[H]
\renewcommand*{\arraystretch}{1.2}
\setlength{\tabcolsep}{4pt}
\caption{Mean MAE imputation results on the missing grid only over five different time window. $\tau$ stands for the subsampling rate. Bold results are best, underline results are second best.}
\label{tab:against_linear_knn}
\begin{center}
\scalebox{0.65}{%
\begin{tabular}{ccccc}
\toprule
  & $\tau$ &  TimeFlow & Linear interpolation & KNN (k=3)\\
\midrule

        & 0.05 &  \textbf{0.324 $\pm$ 0.013} & 0.828 $\pm$ 0.045 & \underline{0.531 $\pm$ 0.033}\\

        & 0.10 &  \textbf{0.250 $\pm$ 0.010} & 0.716 $\pm$ 0.039 & \underline{0.416 $\pm$ 0.020}\\

        Electricity & 0.20 &  \textbf{0.225 $\pm$ 0.008}  & 0.518 $\pm$ 0.029 &  \underline{0.363 $\pm$ 0.019}\\

        & 0.30 &  \textbf{0.212 $\pm$ 0.007} & 0.396 $\pm$ 0.022 & \underline{0.342 $\pm$ 0.017} \\

        & 0.50 &  \textbf{0.194 $\pm$ 0.007} & \underline{0.275 $\pm$ 0.015} &  0.323 $\pm$ 0.016 \\
        
        \midrule

        & 0.05 &  \textbf{0.095 $\pm$ 0.015} & 0.339 $\pm$ 0.031 & \underline{0.151 $\pm$ 0.017}\\

        & 0.10 &  \textbf{0.083 $\pm$ 0.015}  & 0.170 $\pm$ 0.014 & \underline{0.128 $\pm$ 0.017}\\

        Solar & 0.20 &   \textbf{0.072 $\pm$ 0.015}   &  \underline{0.088 $\pm$ 0.010} & 0.110 $\pm$ 0.016 \\

        & 0.30 &   \textbf{0.061 $\pm$ 0.012} &  \underline{0.063 $\pm$ 0.009} & 0.103 $\pm$ 0.017\\

        & 0.50 &   \underline{0.054 $\pm$ 0.013} & \textbf{0.044 $\pm$ 0.008} & 0.096 $\pm$ 0.016 \\

        \midrule

        & 0.05 &  \textbf{0.283 $\pm$ 0.016} & 0.813 $\pm$ 0.027 &  \underline{0.387 $\pm$ 0.014}\\

        & 0.10 &  \textbf{0.211 $\pm$ 0.012} & 0.701 $\pm$ 0.026 &  \underline{0.293 $\pm$ 0.012}\\

        Traffic & 0.20 &  \textbf{0.168 $\pm$ 0.006} & 0.508 $\pm$ 0.022 & \underline{0.249 $\pm$ 0.010}\\

        & 0.30 &  \textbf{0.151 $\pm$ 0.007} & 0.387 $\pm$ 0.018 &  \underline{0.228 $\pm$ 0.009}\\

        & 0.50 &  \textbf{0.139 $\pm$ 0.007} & 0.263 $\pm$ 0.013 &  \underline{0.204 $\pm$ 0.009}\\

        \midrule
        TimeFlow improvement & & / &  49.06 $\%$ & 35.95 $\%$ \\
        
        \bottomrule

\end{tabular}}
\end{center}
\end{table}

\paragraph{Results}{KNN imputation uses information from other individuals and gives satisfactory results at all sampling rates. On the other hand, the purely univariate approach of linear interpolation struggles at low sampling rates but performs well at high sampling rates. TimeFlow significantly outperforms both baselines by a large margin.}


\section{Forecasting experiments}

\subsection{Distinction between adjacent time windows and new time windows during inference} \label{sec:diff_adjacent_new_time_windows_appendix}

\begin{figure*}[!htb]
    \centering
    \includegraphics[width=0.80\linewidth]{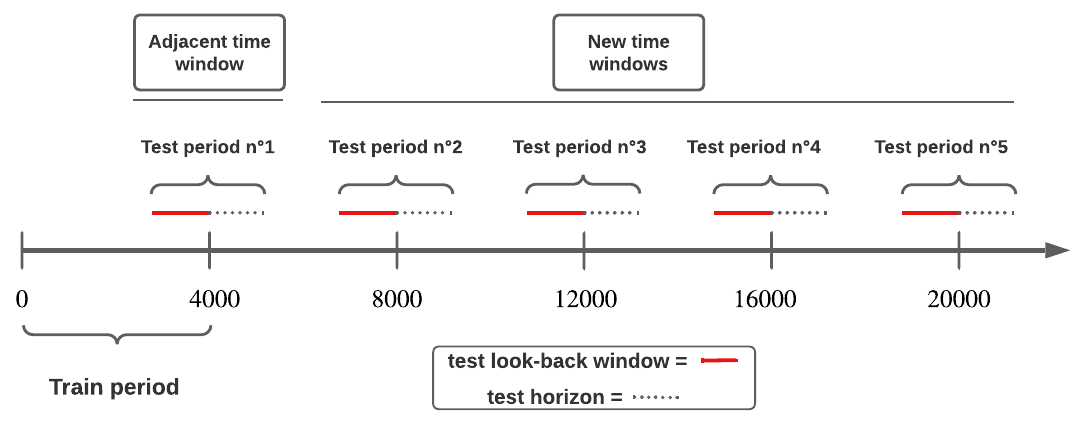}
    \caption{Distinction between adjacent time windows and new time windows during inference for the \textit{Electricity} dataset}
    \label{fig:test_periods_forecast}
\end{figure*}

In \cref{sec:expe_forecasting}, we presented the forecasting results for periods outside the training period. These periods can be classified into two types: adjacent to or disjoint from the training period. \cref{fig:test_periods_forecast} illustrates these distinct test periods for the \textit{Electricity} dataset. The same principle applies to the \textit{Traffic} and \textit{SolarH} datasets, with one notable difference: the number of test periods is smaller in these datasets compared to \textit{Electricity} dataset due to the fewer time steps available.

In \cref{tab:forecast_time_shift_electricity}, we presented the results indistinctly for the two types of test periods: adjacent to and disjoint from the training window. Here, we aim to differentiate the results for these two types of window and emphasize their significant impact on Informer and AutoFormer results. Specifically, \cref{tab:forecast_adjacent_period} showcases the results for the test periods adjacent to the training window. In contrast, \cref{tab:forecast_new_period} displays the results for the test periods disjointed from the training window

\paragraph{Results}{TimeFlow, PatchTST, DLinear and DeepTime maintain consistent forecasting results whether tested on the period adjacent to the training period or on a disjoint period. However, AutoFormer and Informer show a significant drop in performance when tested on new disjoint periods.}

\begin{table}[H]
\renewcommand*{\arraystretch}{1.2}
\setlength{\tabcolsep}{4pt}
\caption{Mean MAE forecast results for adjacent time windows. H stands for the horizon. Bold results are best, underline results are second best.}
\label{tab:forecast_adjacent_period}
\begin{center}
\scalebox{0.65}{%
\begin{tabular}{ccccccccc}

\toprule
  &        & \multicolumn{3}{c}{Continuous methods} & \multicolumn{4}{c}{Discrete methods} \\
  
  \cmidrule(r){3-5} \cmidrule(r){6-9} 
  & $H$ &  TimeFlow & DeepTime & Neural Process & Patch-TST & DLinear & AutoFormer & Informer\\
  
        \midrule

        \multirow{4}{*}{Electricity} & 96 & \underline{0.218 $\pm$ 0.017} & 0.240 $\pm$ 0.027 & 0.392 $\pm$ 0.045& \textbf{0.214 $\pm$ 0.020} & 0.236 $\pm$ 0.035 &  0.310 $\pm$ 0.031 & 0.293 $\pm$ 0.0184 \\
        & 192 & \underline{0.238 $\pm$ 0.012} & 0.251 $\pm$ 0.023 &  0.401 $\pm$ 0.046 & \textbf{0.225 $\pm$ 0.017} & 0.248 $\pm$ 0.032 &  0.322 $\pm$ 0.046 & 0.336 $\pm$ 0.032 \\
        & 336 & \underline{0.265 $\pm$ 0.036} & 0.290 $\pm$ 0.034 &  0.434 $\pm$ 0.075 & \textbf{0.242 $\pm$ 0.024} & 0.284 $\pm$ 0.043 & 0.330 $\pm$ 0.019 & 0.405 $\pm$ 0.044 \\
        & 720 & \underline{0.318 $\pm$ 0.073} & 0.356 $\pm$ 0.060 & 0.605 $\pm$ 0.149 & \textbf{0.291 $\pm$ 0.040} & 0.370 $\pm$ 0.086 &  0.456 $\pm$ 0.052 & 0.489 $\pm$ 0.072\\

        \midrule

        \multirow{4}{*}{SolarH} & 96 & \textbf{0.172 $\pm$ 0.017} & \underline{0.197 $\pm$ 0.002} &  0.221 $\pm$ 0.048 & 0.232 $\pm$ 0.008 & 0.204 $\pm$ 0.002 &  0.261 $\pm$ 0.053 & 0.273 $\pm$ 0.023\\
        & 192 & \textbf{0.198 $\pm$ 0.010} & \underline{0.202 $\pm$ 0.014} & 0.244 $\pm$ 0.048 & 0.231 $\pm$ 0.027 & 0.211 $\pm$ 0.012 &  0.312 $\pm$ 0.085 & 0.256 $\pm$ 0.026\\
        & 336 & \underline{0.207 $\pm$ 0.019} & \textbf{0.200 $\pm$ 0.012}  & 0.241 $\pm$ 0.005 & 0.254 $\pm$ 0.048 & 0.212 $\pm$ 0.019 & 0.341 $\pm$ 0.107 & 0.287 $\pm$ 0.006\\
        & 720 & \textbf{0.215 $\pm$ 0.016} & \underline{0.240 $\pm$ 0.011} & 0.403 $\pm$ 0.147 & 0.271 $\pm$ 0.036 & 0.246 $\pm$ 0.015 & 0.368 $\pm$ 0.006 & 0.341 $\pm$ 0.049\\
                
        \midrule

        \multirow{4}{*}{Traffic} & 96 & \underline{0.216 $\pm$ 0.033} & 0.229 $\pm$ 0.032 & 0.283 $\pm$ 0.028  & \textbf{0.201 $\pm$ 0.031} & 0.225 $\pm$ 0.034 & 0.299 $\pm$ 0.080 & 0.324 $\pm$ 0.113\\
        & 192 & \underline{0.208 $\pm$ 0.021} & 0.220 $\pm$ 0.020 & 0.292 $\pm$ 0.023 & \textbf{0.195 $\pm$ 0.024} & 0.215 $\pm$ 0.022 &  0.320 $\pm$ 0.036 & 0.321 $\pm$ 0.052\\
        & 336 & \underline{0.237 $\pm$ 0.040} & 0.247 $\pm$ 0.033 & 0.305 $\pm$ 0.039 & \textbf{0.220 $\pm$ 0.036} & 0.244 $\pm$ 0.035 & 0.450 $\pm$ 0.127 & 0.394 $\pm$ 0.066\\
        & 720 & \textbf{0.266 $\pm$ 0.048} & 0.290 $\pm$ 0.045 & 0.339 $\pm$ 0.037 & \underline{0.268 $\pm$ 0.050} & 0.290 $\pm$ 0.047 & 0.630 $\pm$ 0.043 & 0.441 $\pm$ 0.055 \\

        \midrule
          TimeFlow improvement & & / &  6.56 $\%$ &  30.79 $\%$ &  2.64 $\%$ &  7.30 $\%$ &  35.43 $\%$ &  33.07 $\%$ \\
        \bottomrule
\end{tabular}}
\end{center}
\end{table}









\begin{table}[H]
\renewcommand*{\arraystretch}{1.2}
\setlength{\tabcolsep}{4pt}
\caption{Mean MAE forecast results for new time windows. H stands for the horizon. Bold results are best, underline results are second best.}
\label{tab:forecast_new_period}
\begin{center}
\scalebox{0.65}{%
\begin{tabular}{ccccccccccc}

\toprule
  &        & \multicolumn{3}{c}{Continuous methods} & \multicolumn{4}{c}{Discrete methods} \\
  
  \cmidrule(r){3-5} \cmidrule(r){6-9} 
  & $H$ &  TimeFlow & DeepTime & Neural Process & Patch-TST & DLinear & AutoFormer & Informer\\
  
        \midrule

        \multirow{4}{*}{Electricity} & 96 & \underline{0.230 $\pm$ 0.012} & 0.245 $\pm$ 0.026 & 0.392 $\pm$ 0.045 & \textbf{0.222 $\pm$ 0.023} & 0.240 $\pm$ 0.025 & 0.606 $\pm$ 0.281 & 0.605 $\pm$ 0.227 \\
        & 192 & \underline{0.246 $\pm$ 0.025} & 0.252 $\pm$ 0.018 & 0.401 $\pm$ 0.046 & \textbf{0.231 $\pm$ 0.020} & 0.257 $\pm$ 0.027 & 0.545 $\pm$ 0.186 & 0.776 $\pm$ 0.257 \\
        & 336 & \underline{0.271 $\pm$ 0.029} & 0.285 $\pm$ 0.034 & 0.434 $\pm$ 0.076 & \textbf{0.253 $\pm$ 0.027} & 0.298 $\pm$ 0.051 & 0.571 $\pm$ 0.181 & 0.823 $\pm$ 0.241 \\
        & 720 & \underline{0.316 $\pm$ 0.051} & 0.359 $\pm$ 0.048  & 0.607 $\pm$ 0.15 & \textbf{0.299 $\pm$ 0.038} & 0.373 $\pm$ 0.075 & 0.674 $\pm$ 0.245 & 0.811 $\pm$ 0.257 \\
        
        \midrule

        \multirow{4}{*}{SolarH} & 96 & \underline{0.208 $\pm$ 0.005} & \textbf{0.206 $\pm$ 0.026} & 0.221 $\pm$ 0.048 & 0.293 $\pm$ 0.089 & 0.212 $\pm$ 0.019 & 0.228 $\pm$ 0.027 & 0.234 $\pm$ 0.011 \\
        & 192 & \textbf{0.206 $\pm$ 0.012} & \underline{0.207 $\pm$ 0.037} & 0.244 $\pm$ 0.048& 0.274 $\pm$ 0.060 & 0.223 $\pm$ 0.029 & 0.356 $\pm$ 0.122 & 0.280 $\pm$ 0.033 \\
        & 336 & \underline{0.211 $\pm$ 0.005} & \textbf{0.199 $\pm$ 0.035} & 0.240 $\pm$ 0.006 & 0.264 $\pm$ 0.088 & 0.223 $\pm$ 0.032 & 0.327 $\pm$ 0.029 & 0.366 $\pm$ 0.039 \\
        & 720 & \textbf{0.222 $\pm$ 0.020} & \underline{0.217 $\pm$ 0.028} & 0.403 $\pm$ 0.147 & 0.262 $\pm$ 0.083 & 0.251 $\pm$ 0.047 & 0.335 $\pm$ 0.075 & 0.333 $\pm$ 0.012\\
                
        \midrule

        \multirow{4}{*}{Traffic} & 96 & \underline{0.218 $\pm$ 0.042}  & 0.229 $\pm$ 0.032 & 0.283, 0.0275 & \textbf{0.204 $\pm$ 0.039} & 0.229 $\pm$ 0.032 & 0.326 $\pm$ 0.049 & 0.388 $\pm$ 0.055 \\
        & 192 & \underline{0.213 $\pm$ 0.028}  & 0.220 $\pm$ 0.023 & 0.292, 0.0236 & \textbf{0.198 $\pm$ 0.031} & 0.223 $\pm$ 0.023 & 0.575 $\pm$ 0.254 & 0.381 $\pm$ 0.049 \\
        & 336 & \underline{0.239 $\pm$ 0.035}  & 0.244 $\pm$ 0.040 & 0.305, 0.0392 & \textbf{0.223 $\pm$ 0.040} & 0.252 $\pm$ 0.042 & 0.598 $\pm$ 0.286 & 0.448 $\pm$ 0.055\\
        & 720 & \underline{0.280 $\pm$ 0.047} & 0.290 $\pm$ 0.055 & 0.339, 0.0375 & \textbf{0.270 $\pm$ 0.059} & 0.304 $\pm$ 0.061 & 0.641 $\pm$ 0.072 & 0.468 $\pm$ 0.064\\

        \midrule
         TimeFlow improvement & & / &  2.50 $\%$ &   27.75 $\%$ &  3.41 $\%$ &  6.80 $\%$ &   46.26  $\%$ &  45.53 $\%$ \\
        
        \bottomrule
\end{tabular}}
\end{center}
\end{table}

%






\subsection{Plots comparison: TimeFlow vs PatchTST}

\cref{tab:forecast_time_shift_electricity} demonstrates the similar forecasting performance of TimeFlow and PatchTST across all horizons. To visually represent their predictions, the figures below showcase the forecasted outcomes of these methods for two samples (24 and 38) and two horizons (96 and 192) on the \textit{Electricity}, \textit{SolarH}, and \textit{Traffic} datasets.

\begin{figure}[H]
    \centering
    \includegraphics[width=0.9\linewidth]{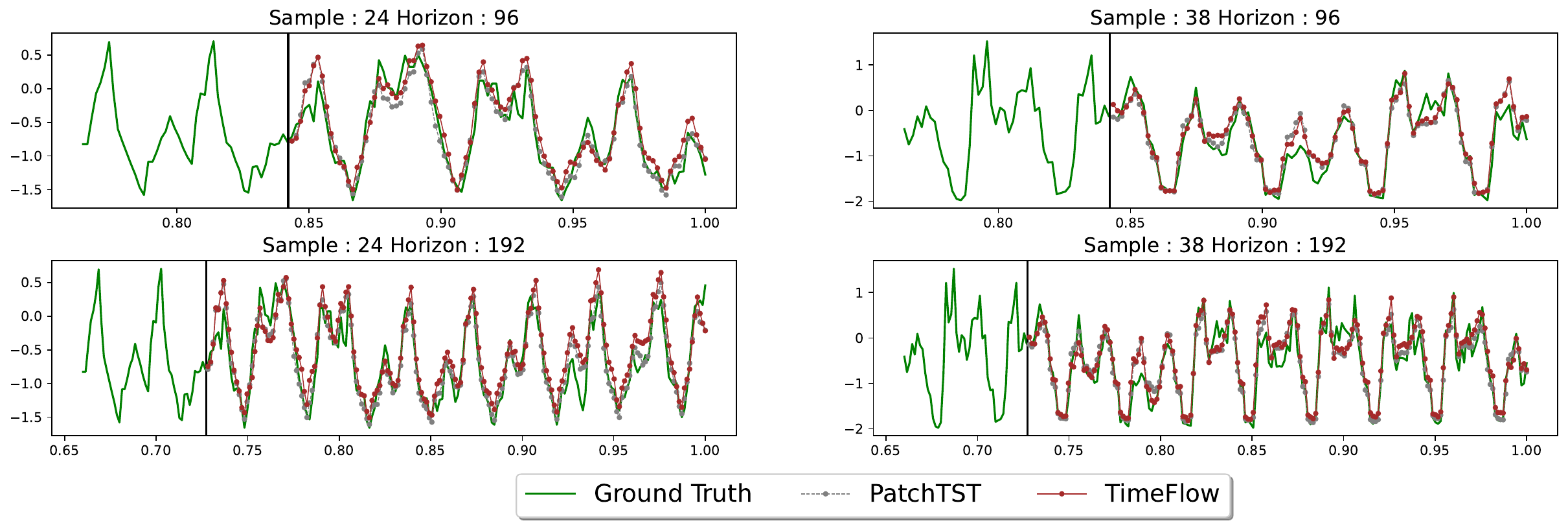}
    \caption{Qualitative comparisons of TimeFlow vs PatchTST on the \textit{Electricity} dataset for new time windows}
    \label{fig:forecast_elec_patchTST}
\end{figure}

\begin{figure}[H]
    \centering
    \includegraphics[width=0.9\linewidth]{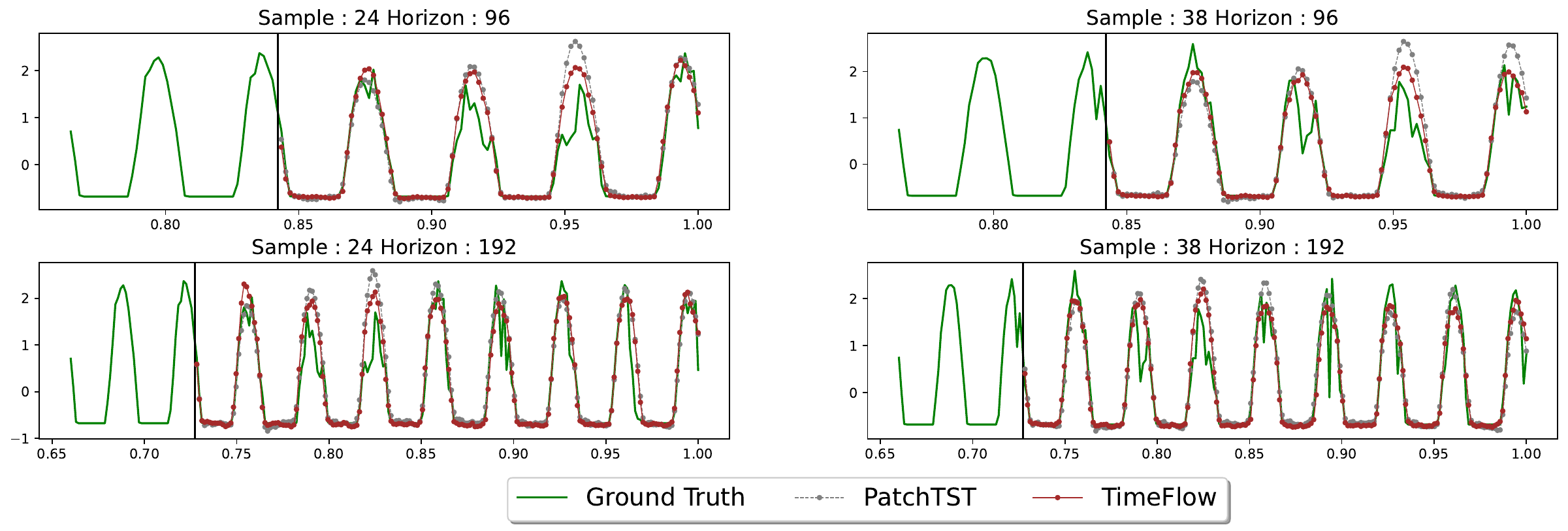}
    \caption{Qualitative comparisons of TimeFlow vs PatchTST on the \textit{SolarH} dataset for new time windows}
    \label{fig:forecast_elec_patchTST_solar}
\end{figure}

\begin{figure}[H]
    \centering
    \includegraphics[width=0.9\linewidth]{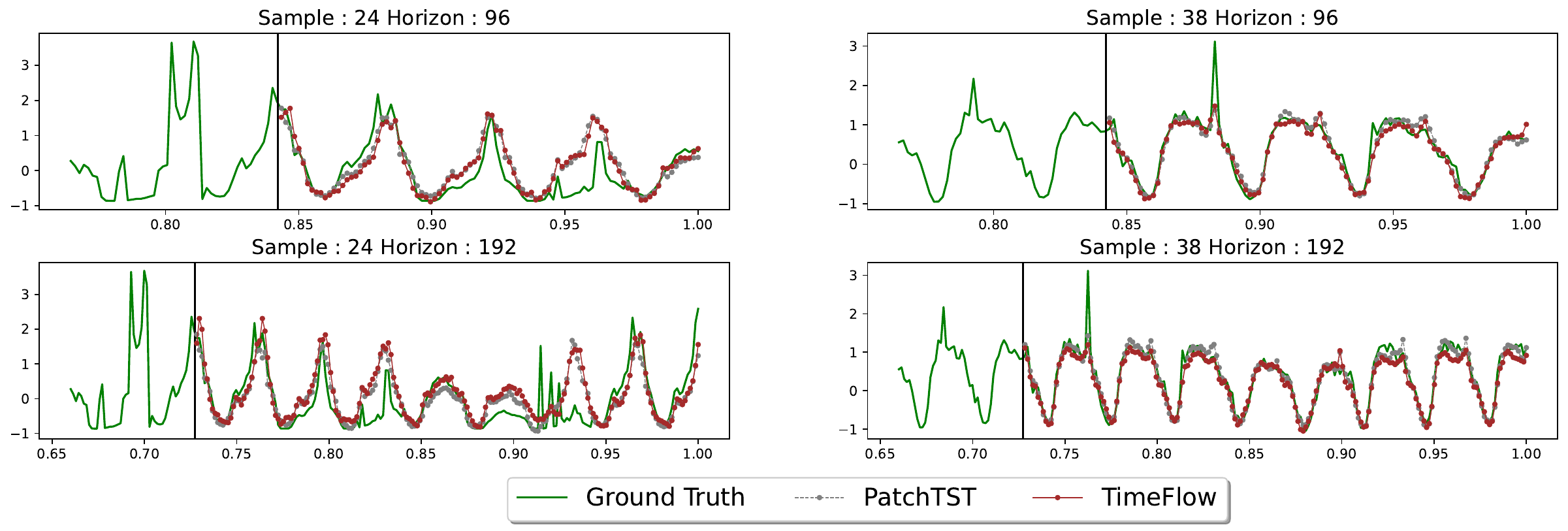}
    \caption{Qualitative comparisons of TimeFlow vs PatchTST on the \textit{Traffic} dataset for new time windows}
    \label{fig:forecast_elec_patchTST_traffic}
\end{figure}

\paragraph{Results}{The visual analysis of the figures above reveals that the predictions of TimeFlow and PatchTST are remarkably similar. For instance, when examining sample 24 and horizon 192 of the \textit{Traffic} dataset, both forecasters exhibit similar error patterns. The only noticeable distinction emerges in the \textit{SolarH} dataset, where PatchTST tends to overestimate certain peaks.}

\subsection{Baseline details}

\subsubsection{Baselines training and hyperparameters}
\label{baseline-forecast-training}

We provide a detailed breakdown of the hyperparameters and our training approach for the forecasting baselines. We took an in-depth approach, testing each method under a range of configurations to ensure they were well-suited to the unique characteristics of the datasets and tasks at hand.
For DLinear and transformer baselines, including PatchTST, AutoFormer, and Informer, we utilized the implementations detailed in the PatchTST baselines (\href{https://github.com/yuqinie98/PatchTST}{code}) and adhered to the best practices recommended for our particular tasks. Notably, our implementation of PatchTST was combined with ReVIN, enhancing the robustness of the results. Regarding DeepTime, we followed the recommended hyperparameters, opting for a structure with 5 layers, each 256 units wide, and using 4096 Fourier features spanning a diverse set of scales. As for the Neural Process, the standard model did not train as expected. So, we customized its architecture to conduct a fair comparison with TimeFlow. We used the INR and hypernetwork from TimeFlow to align the Neural Process with our temporal frequency bias and shift modulation technique. We also meticulously searched for the optimal hyperparameters, like the Kullback Leibler (KL) divergence weight and learning rate. Moreover, we extended the training duration to ensure thorough convergence.

\subsubsection{Models complexity}

In this section, we present the parameter counts and the inference time for the main forecasting baselines.
Except for TimeFlow and DeepTime, the number of parameters varies with the number of samples, the look-back window, and the horizon. Thus, we report the number of parameters for two specific configurations, including a fixed dataset, a fixed look-back window, and a fixed horizon.
In \cref{tab:weight_comparison_forecasting}, we see that for PatchTST and DLinear, the larger the horizon, the more the number of parameters increases. 
In \cref{tab:inference_time}, it is shown that all methods' computational time increases with the horizon, which is expected. Moreover, TimeFlow is slower than the baselines that use forward computations only. Still, on the Electricity dataset, for example, the method can infer for 321 samples a horizon of 720 values with a look-back window of 512 timestamps in less than 0.2s, which does not look prohibitive for many real-world usages. This is mainly due to the small number of gradient steps at inference.

\begin{table}[H]
\renewcommand*{\arraystretch}{1.2}
\setlength{\tabcolsep}{4pt}
\caption{The number of parameters for main baselines on the forecasting task on the \textit{Electricity} dataset for horizons 96 and 720. The look-back window size is 512. }
\label{tab:weight_comparison_forecasting}
\begin{center}
\scalebox{0.65}{%
\begin{tabular}{ccccccccc}
    \toprule[1.5pt]
    & TimeFlow & DeepTime & Neural Process & Patch-TST & DLinear & Informer & Autoformer  \\
    \midrule

    96 & 602k & 1 315k & 480k & 1 194k & 98k & 984k & 1 005k\\
    720 & 602k & 1 315k & 480k & 6 306k & 739k & 984k & 1 005k \\

    \bottomrule[1.5pt]

\end{tabular}}
\end{center}
\end{table}

\begin{table}[H]
\renewcommand*{\arraystretch}{1.2}
\setlength{\tabcolsep}{4pt}
\caption{Inference time (in seconds) for the forecasting task on the \textit{Electricity} dataset with horizons 96 and 720 and a look-back window of length 512. The statistics are computed over 10 runs using an NVIDIA TITAN RTX GPU.}
\label{tab:inference_time}
\begin{center}
\scalebox{0.65}{%
\begin{tabular}{ccccccc}
    \toprule[1.5pt]
    & TimeFlow & Patch-TST & DLinear & DeepTime & AutoFormer & Informer \\
    \midrule

    96 & 0.147 $\pm$ 0.007 & 0.016 $\pm$ 0.002 & 0.007 $\pm$ 0.003 & 0.006 $\pm$ 0.002 & 0.027 $\pm$ 0.001& 0.0191 $\pm$ 0.002\\
    720 & 0.176 $\pm$ 0.009  & 0.020 $\pm$ 0.001 & 0.009 $\pm$ 0.001  & 0.010 $\pm$ 0.002 & 0.034$\pm$ 0.001& 0.0251 $\pm$ 0.002\\

    \bottomrule[1.5pt]

\end{tabular}}
\end{center}
\end{table}

\subsection{Sparsely observed look-back window: comparison with Patch-TST}

\paragraph{Setting and baseline.}{Let's consider a setting where at inference time, the look-back window is sparsely observed. Models such as PatchTST must proceed in two steps: (i) completing the look-back window on a dense regular grid using imputation; (ii) apply the model on the completed window to predict the future. We compared TimeFlow with the following two-step processing baseline: linear interpolation handling the missing values within the partially observed look-back window, and PatchTST handling the forecasting task. We conducted experiments on the \textit{Traffic} and \textit{Electricity} datasets, focusing on the 96 and 192 horizons.  In \cref{tab:PatchTST_sparse}, we present the results at different sampling rates $\tau \in \{0.5, 0.2, 0.1\}$ within the look-back window.}

\begin{table}[htb]
\renewcommand*{\arraystretch}{1.1}
\setlength{\tabcolsep}{5pt}
\caption{MAE results for forecasting on new samples and new period with missing values in the look-back window. Best results are in bold.}
\label{tab:PatchTST_sparse}
\begin{center}
\scalebox{0.65}{%
\begin{tabular}{ccccccc}
\toprule[1.5pt]
  &   &        & \multicolumn{2}{c}{TimeFlow} & \multicolumn{2}{c}{Linear interpo $+$ PatchTST} \\
  
  \cmidrule(r){4-5} \cmidrule(r){6-7}                                       
  
  & H & $\tau$ & Imputation error & Forecast error & Imputation error & Forecast error\\
  
        \midrule
         \multirow{8}{*}{Electricity} & \multirow{4}{*}{96} & 1.  & 0.000 $\pm$ 0.000  & 0.228 $\pm$ 0.028 & 0.000 $\pm$ 0.000 & \textbf{0.221 $\pm$ 0.023} \\
         &    & 0.5 & \textbf{0.151 $\pm$ 0.003} & \textbf{0.239 $\pm$ 0.013} &  0.257 $\pm$ 0.008 & 0.279 $\pm$ 0.026 \\
                                      &                      & 0.2 & \textbf{0.208 $\pm$ 0.006} & \textbf{0.260 $\pm$ 0.015} & 0.482 $\pm$ 0.019 & 0.451 $\pm$ 0.042 \\
                                      &                      & 0.1 & \textbf{0.272 $\pm$ 0.006} & \textbf{0.295 $\pm$ 0.016} & 0.663 $\pm$ 0.029  & 0.634 $\pm$ 0.053\\

                            \cmidrule(r){2-7}
                    
                                        & \multirow{4}{*}{192} & 1. & 0.000 $\pm$ 0.000 & 0.238 $\pm$ 0.020 & 0.000 $\pm$ 0.000 &  \textbf{0.229 $\pm$ 0.020}\\
                                       &  & 0.5 & \textbf{0.149 $\pm$ 0.004} & \textbf{0.235 $\pm$ 0.011} & 0.258 $\pm$ 0.006 & 0.280 $\pm$ 0.032\\
                                       &                      & 0.2 & \textbf{0.209 $\pm$ 0.006} & \textbf{0.257 $\pm$ 0.013} & 0.481 $\pm$ 0.021 & 0.450 $\pm$ 0.054 \\
                                       &                      & 0.1 & \textbf{0.274 $\pm$ 0.010} & \textbf{0.289 $\pm$ 0.016} & 0.669 $\pm$ 0.030 & 0.650 $\pm$ 0.060 \\

        \midrule

        \multirow{8}{*}{Traffic} & \multirow{4}{*}{96} & 1.  & 0.000 $\pm$ 0.000 & 0.217 $\pm$ 0.032 & 0.000 $\pm$ 0.000 & \textbf{0.203 $\pm$ 0.037}\\
         &  & 0.5 & \textbf{0.219 $\pm$ 0.017} & \textbf{0.224 $\pm$ 0.033} & 0.276 $\pm$ 0.012 & 0.255 $\pm$ 0.041 \\
                                    &          & 0.2 & \textbf{0.278 $\pm$ 0.017} & \textbf{0.252 $\pm$ 0.029} & 0.532 $\pm$ 0.017 & 0.483 $\pm$ 0.040 \\
                                          &    & 0.1 & \textbf{0.418 $\pm$ 0.019} & \textbf{0.382 $\pm$ 0.014} & 0.738 $\pm$ 0.023 & 0.721 $\pm$ 0.073 \\

                    \cmidrule(r){2-7}
                    
                                       & \multirow{4}{*}{192} & 1. & 0.000 $\pm$ 0.000 & 0.212 $\pm$ 0.028 & 0.000 $\pm$ 0.000 & \textbf{0.197 $\pm$ 0.030} \\
              &   & 0.5 & \textbf{0.176 $\pm$ 0.014} & \textbf{0.217 $\pm$ 0.017} & 0.276 $\pm$ 0.011 & 0.245 $\pm$ 0.029 \\
                                  &     & 0.2 & \textbf{0.233 $\pm$ 0.017} & \textbf{0.236 $\pm$ 0.021} & 0.532 $\pm$ 0.020 & 0.480 $\pm$ 0.050 \\
                                  &     & 0.1 & \textbf{0.304 $\pm$ 0.019} & \textbf{0.277 $\pm$ 0.021} & 0.734 $\pm$ 0.022 & 0.787 $\pm$ 0.172 \\
        
        \bottomrule[1.5pt]

\end{tabular}}
\end{center}
\end{table}

\paragraph{Results.}{Although PatchTST performs slightly better with a dense look-back window, its performance significantly deteriorates as the value of $\tau$ decreases. In contrast, the performance of TimeFlow is only minimally affected by the reduction in the sampling rate.}

\subsection{Influence of the look-back window for forecasting}

\begin{figure}[H]
    \centering
    \includegraphics[width=0.70\linewidth]{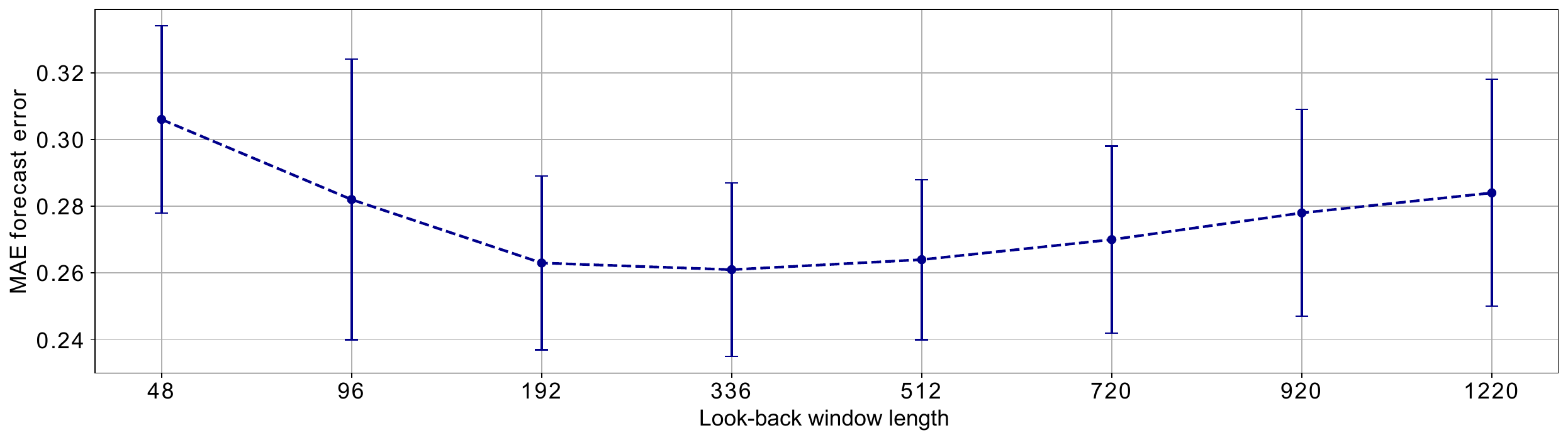}
    \caption{MAE forecast error per look-back windows length for the \textit{Electricity} dataset (horizon window length is 336). The model is trained on a given time window and tested on four new time windows.}
    \label{fig:mae_per_lbw}
\end{figure}

In \cref{fig:mae_per_lbw}, it is shown that both excessively short and overly long look-back windows can harm TimeFlow forecasting performance. More precisely, the performances increases with the look-back window size up to a certain size, where the performances then drop slowly.

\subsection{Influence of the horizon length for forecasting}

\begin{figure}[H]
    \centering
    \includegraphics[width=0.70\linewidth]{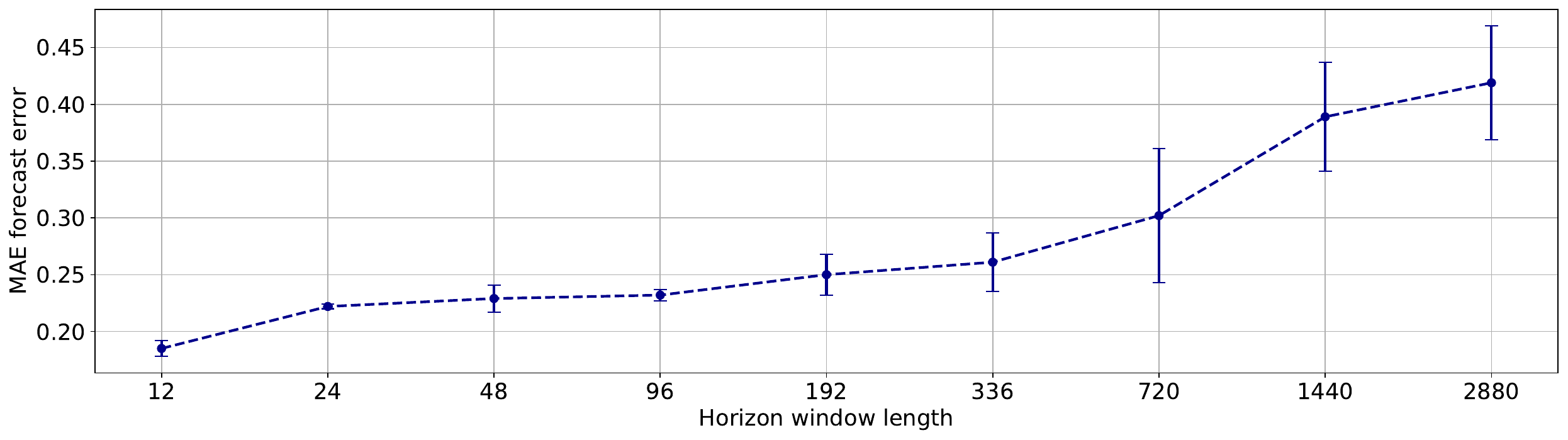}
    \caption{MAE forecast error per horizons length for the \textit{Electricity} dataset (look-back window length is 512). The model is trained on a given time window and tested on four new time windows.}
    \label{fig:mae_per_horizons}
\end{figure}

In \cref{fig:mae_per_horizons}, it is shown that the performances decrease with the length of the horizon. This is to be expected, since the longer the horizon, the harder the task. 

\clearpage

\section{Discussion on using frequency embedding as input to regression models}
\label{sec:discussion_ff_embeddings}

\subsection{Related work}
In forecasting univariate time series, several models have explored the integration of frequency embedding of timestamps for regression purposes. \cite{taylor2018forecasting} approached the time series as a continuous function of time using a general additive model \citep{hastie2017generalized}. They represented seasonality components as learnable Fourier series while explicitly specifying the ground truth seasonalities (e.g., weekly, monthly). Similarly, \cite{hyndman2018forecasting} proposed embedding the timestamp $t$ with the ground truth frequencies and applying a regression model to predict the series value at the timestamp $t$. Both methods rely on the explicit specification of seasonalities and are tailored for purely univariate time series, where information is not shared between samples.

In contrast, other models, such as TimeFlow or DeepTime \citep{DeepTime}, based on deep learning techniques, offer more flexibility. These approaches can autonomously learn relevant frequencies and effectively share information between samples through backpropagation. This enables a more dynamic and adaptable approach to time series forecasting, particularly in scenarios with complex temporal patterns and inter-sample dependencies.

\subsection{Experiments}

Given the seasonal patterns observed in the \textit{Solar}, \textit{Electricity}, and \textit{Traffic} datasets, an alternative to deep learning forecasting methods is to individually regress a timestamp embedding on the corresponding value using a robust regressor. This approach exploits the inherent periodicity in the data, using timestamp embeddings to capture temporal dependencies and accurately predict the target values.

\paragraph{Baselines.}{We compare TimeFlow against two regression baselines}:

\begin{itemize}
    \item \textbf{TimeFlow frequencies embedding $+$ XGBoost}: This baseline uses the same frequency embedding as TimeFlow and applies an XGBoost regressor \citep{chen2016xgboost} on top. The aim is to assess whether the XGBoost regressor can effectively identify the correct frequencies, filter out irrelevant ones, and establish the appropriate mapping between timestamps and values.
    \item \textbf{XGB Explicit Seasonal encoding + XGboost}: in this baseline, we give \textit{explicitly} the right frequencies of each datasets to the model. It is important to highlight that \textbf{this method uses information that other baselines don't}. For instance for the \textit{Traffic} dataset, the Explicit Seasonal encoding is 
    $\gamma(t) = \left( t, \cos\left(\frac{2 \pi t}{24}\right), \sin\left(\frac{2 \pi t}{24}\right), \cos\left(\frac{2 \pi t}{24 \times 7}\right), \sin\left(\frac{2 \pi t}{24 \times 7}\right) \right)$. It allows to explicitly integers trend, daily frequencies and weekly frequencies. We apply the same type of frequencies embedding for $Solar$ and $Electricity$ with the appropriate seasonalities.
 \end{itemize}

\paragraph{Experimental Setup.}{For each dataset and sample, we applied the frequency embedding individually and then trained an XGBoost regressor on each observed timestamp (the look-back window in forecasting and the observed grid in imputation). This approach results in one model per sample. The XGBoost regressor is configured with the following hyperparameters:}
\begin{itemize}
    \item n estimators: 500
    \item max depth: 4
    \item learning rate: 0.1
    \item lambda: [0.1, 1, 10]
\end{itemize}

The optimal regularization parameter lambda is determined through cross-validation. The imputation and forecasting results are presented in detail in \cref{tab:imputation_fourier_features} and \cref{tab:forecast_fourier_adjacent_period}, respectively.

\begin{table}[H]
\renewcommand*{\arraystretch}{1.2}
\setlength{\tabcolsep}{4pt}
\caption{Mean MAE imputation results on the missing grid only. In
the table, $\tau$ stands for the subsampling rate, i.e. the proportion of observed points considered for each samples. Bold results are best, underlined results are second best.}
\label{tab:imputation_fourier_features}
\begin{center}
\scalebox{0.65}{
\begin{tabular}{ccccc}

        \toprule

        & $\tau$ &  TimeFlow & TimeFlow frequencies embedding+ XGB & Explicit Seasonal encoding + XGB \\

        \midrule 

        & 0.05 &  \textbf{0.324 $\pm$ 0.013} & 0.834 $\pm$ 0.092 & \underline{0.365 $\pm$ 0.051}\\

        & 0.10 &  \textbf{0.250 $\pm$ 0.010} &  0.761 $\pm$ 0.074 & \underline{0.318 $\pm$ 0.049} \\

        Electricity & 0.20 &  \textbf{0.225 $\pm$ 0.008} & 0.632 $\pm$ 0.066 & \underline{0.278 $\pm$ 0.044}\\

        & 0.30 &  \textbf{0.212 $\pm$ 0.007} & 0.536 $\pm$ 0.041 & \underline{0.259 $\pm$ 0.048}\\

        & 0.50 &  \textbf{0.194 $\pm$ 0.007}  & 0.418 $\pm$ 0.042 & \underline{0.238 $\pm$ 0.022}\\
        
        \midrule

        & 0.05 &  \textbf{0.095 $\pm$ 0.015} & 0.603 $\pm$ 0.035 & \underline{0.234 $\pm$ 0.021} \\

        & 0.10 &  \textbf{0.083 $\pm$ 0.015} & 0.478 $\pm$ 0.024 & \underline{0.190 $\pm$ 0.022}\\

        Solar & 0.20 &   \textbf{0.072 $\pm$ 0.015} & 0.350 $\pm$ 0.022 & \underline{0.150 $\pm$ 0.019}\\

        & 0.30 &   \textbf{0.061 $\pm$ 0.012} & 0.286 $\pm$ 0.018 & \underline{0.134 $\pm$ 0.011}\\

        & 0.50 &   \textbf{0.054 $\pm$ 0.013} & 0.227 $\pm$ 0.015 & \underline{0.123 $\pm$ 0.015}\\

        \midrule

        & 0.05 &  \textbf{0.283 $\pm$ 0.016} & 0.739 $\pm$ 0.140 & \underline{0.344 $\pm$ 0.036}\\

        & 0.10 &  \textbf{0.211 $\pm$ 0.012} & 0.676 $\pm$ 0.129 &  \underline{0.290 $\pm$ 0.029}\\

        Traffic & 0.20 &  \textbf{0.168 $\pm$ 0.006} & 0.562 $\pm$ 0.108 & \underline{0.245 $\pm$ 0.027}\\

        & 0.30 &  \textbf{0.151 $\pm$ 0.007} & 0.487 $\pm$ 0.095 & \underline{0.223 $\pm$ 0.015}\\

        & 0.50 &  \textbf{0.139 $\pm$ 0.007} & 0.393 $\pm$ 0.083 & \underline{0.198 $\pm$ 0.021}\\

        \midrule
        
         TimeFlow improvement & & / & 69.5 $\%$ & 33.7 $\%$\\

        \bottomrule

\end{tabular}}
\end{center}
\end{table}

\paragraph{Imputation results.}{TimeFlow performs better than the other two methods. Although the second baseline explicitly incorporates ground truth frequencies and provides decent results, its inability to share information between samples leads to the loss of valuable insights that TimeFlow effectively exploits. In addition, the first baseline struggles to learn the correct frequencies and overfits observed data points, resulting in excessive high-frequency noise. As a result, its performance degrades significantly compared to the second baseline, where frequencies are explicitly provided. These findings underscore TimeFlow's ability to identify the underlying frequencies, and leverage shared information across samples to improve accuracy during imputation.}

\begin{table}[H]
\renewcommand*{\arraystretch}{1.2}
\setlength{\tabcolsep}{4pt}
\caption{Mean MAE forecast results for adjacent time windows averaged over different time windows. Each time, the model is trained on one time window and tested on the others (there are 2 windows for SolarH and 5 for Electricity and Traffic). $H$ stands for the horizon. Bold results are best, and underlined results are second best}
\label{tab:forecast_fourier_adjacent_period}
\begin{center}
\scalebox{0.65}{
\begin{tabular}{cccccc}

        \toprule
        & $H$ &  TimeFlow & TimeFlow frequencies embedding + XGB  & Explicit Seasonnal encoding + XGB\\
          
        \midrule

        \multirow{4}{*}{Electricity} & 96 & \textbf{0.218 $\pm$ 0.017} & 0.662 $\pm$ 0.102  & \underline{0.282 $\pm$ 0.020}\\
        & 192 & \textbf{0.238 $\pm$ 0.012}  & 0.750 $\pm$ 0.128 & \underline{0.279 $\pm$ 0.021}\\
        & 336 & \textbf{0.265 $\pm$ 0.036}  & 0.809 $\pm$ 0.136 & \underline{0.294 $\pm$ 0.041}\\
        & 720 & \textbf{0.318 $\pm$ 0.073}  & 0.852 $\pm$ 0.144 & \underline{0.357 $\pm$ 0.092}\\

        \midrule

        \multirow{4}{*}{SolarH} & 96 & \textbf{0.172 $\pm$ 0.017}  & 0.792 $\pm$ 0.062 & \underline{0.244 $\pm$ 0.023} \\
        & 192 & \textbf{0.198 $\pm$ 0.010}  & 0.933 $\pm$ 0.055 & \underline{0.236 $\pm$ 0.018}\\
        & 336 & \textbf{0.207 $\pm$ 0.019}  & 1.033 $\pm$ 0.052 & \underline{0.229 $\pm$ 0.022}\\
        & 720 & \textbf{0.215 $\pm$ 0.016}  & 1.116 $\pm$ 0.057 & \underline{0.262 $\pm$ 0.021}\\
                
        \midrule

        \multirow{4}{*}{Traffic} & 96  & \textbf{0.216 $\pm$ 0.033} & 0.655 $\pm$ 0.156  & \underline{0.288 $\pm$ 0.052}\\
        & 192 & \textbf{0.208 $\pm$ 0.021}  & 0.678 $\pm$ 0.139 & \underline{0.246 $\pm$ 0.033}\\
        & 336 & \textbf{0.237 $\pm$ 0.040}  & 0.719 $\pm$ 0.143 & \underline{0.262 $\pm$ 0.044}\\
        & 720 & \textbf{0.266 $\pm$ 0.048}  & 0.741 $\pm$ 0.140 & \underline{0.288 $\pm$ 0.063}\\

        \midrule
        TimeFlow improvement & & / & 70.8 \% & 15.7 \% \\
        \bottomrule
\end{tabular}}
\end{center}
\end{table}

\paragraph{Forecasting results.}{TimeFlow also outperforms the other two baselines in forecasting. However, the improvement over the second baseline, where the correct frequencies are explicitly provided, is more modest compared to the gains observed in imputation. Nevertheless, the relative improvement achieved by TimeFlow remains significant (exceeding 15 $\%$). Similar to the imputation scenario, the XGBoost baseline with TimeFlow timestamps encoding, which attempts to learn the correct frequencies, fails to discern them accurately and introduces excessive high-frequency components.}

\section{Latent space exploration}

\subsection{Latent space interpolation between two learned codes}

\begin{figure}
    \centering
    \includegraphics[width=0.7\linewidth]{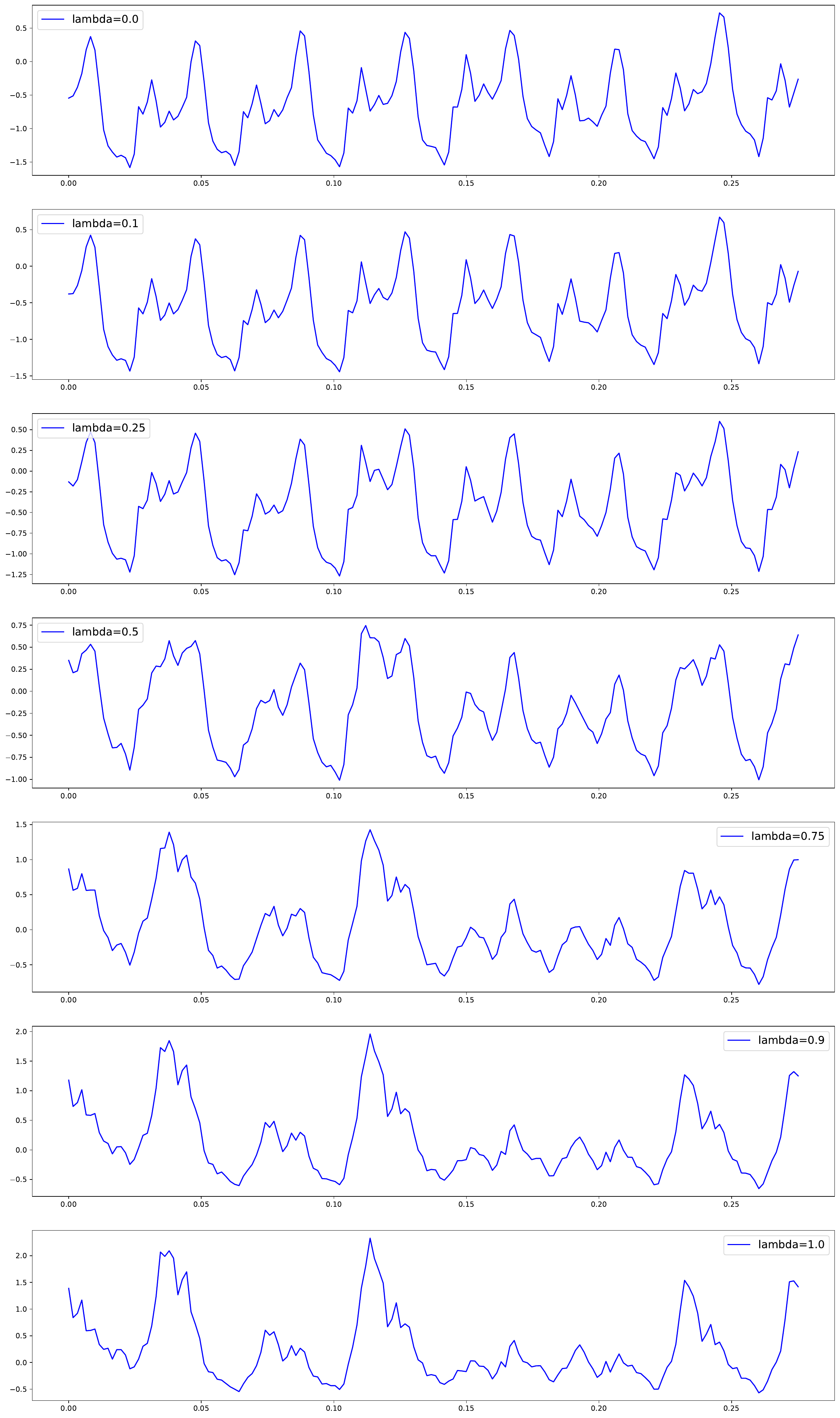}
    \caption{Visualization of the reconstructed time series for different linear interpolation of the two codes $z_1$ and $z_2$ learned from the \textit{Electricity} dataset.}
    \label{fig:latent_space_interpo}
\end{figure}

It is interesting to understand how the latent space behaves between two learned codes, $z_1$ and $z_2$, which are representations of $x_1$ and $x_2$. We propose in \cref{fig:latent_space_interpo} to visualize how new $z_{\lambda} = \lambda z_1 + (1 - \lambda) z_2$  are decoded in the time-series domain ($f_{\theta, h_w({z_{\lambda}})}(t)$ values).

\paragraph{Setting.}{We choose two $z_1$ and $z_2$ learned from the \textit{Electricity} dataset and interpolate these two latent codes for $\lambda \in \{0.0, 0.1, 0.25, 0.50, 0.75, 0.9, 1. \}.$}

\paragraph{Results.}{In \cref{fig:latent_space_interpo}, we observe that the interpolation path between two codes yields a smooth transition in the time series domain. This suggests that the latent space is smooth and well-structured, and that the learned representations captured meaningful features of the time series, which could explain TimeFlow's generalization property.}


\subsection{TimeFlow sensitivity to modulations perturbation}

In the preceding section, we observed the smoothness of the latent space. A crucial question arises: can we interpret each dimension of the latent space independently ? 

\begin{figure}[h!]
    \centering
    \includegraphics[width=0.6\linewidth]{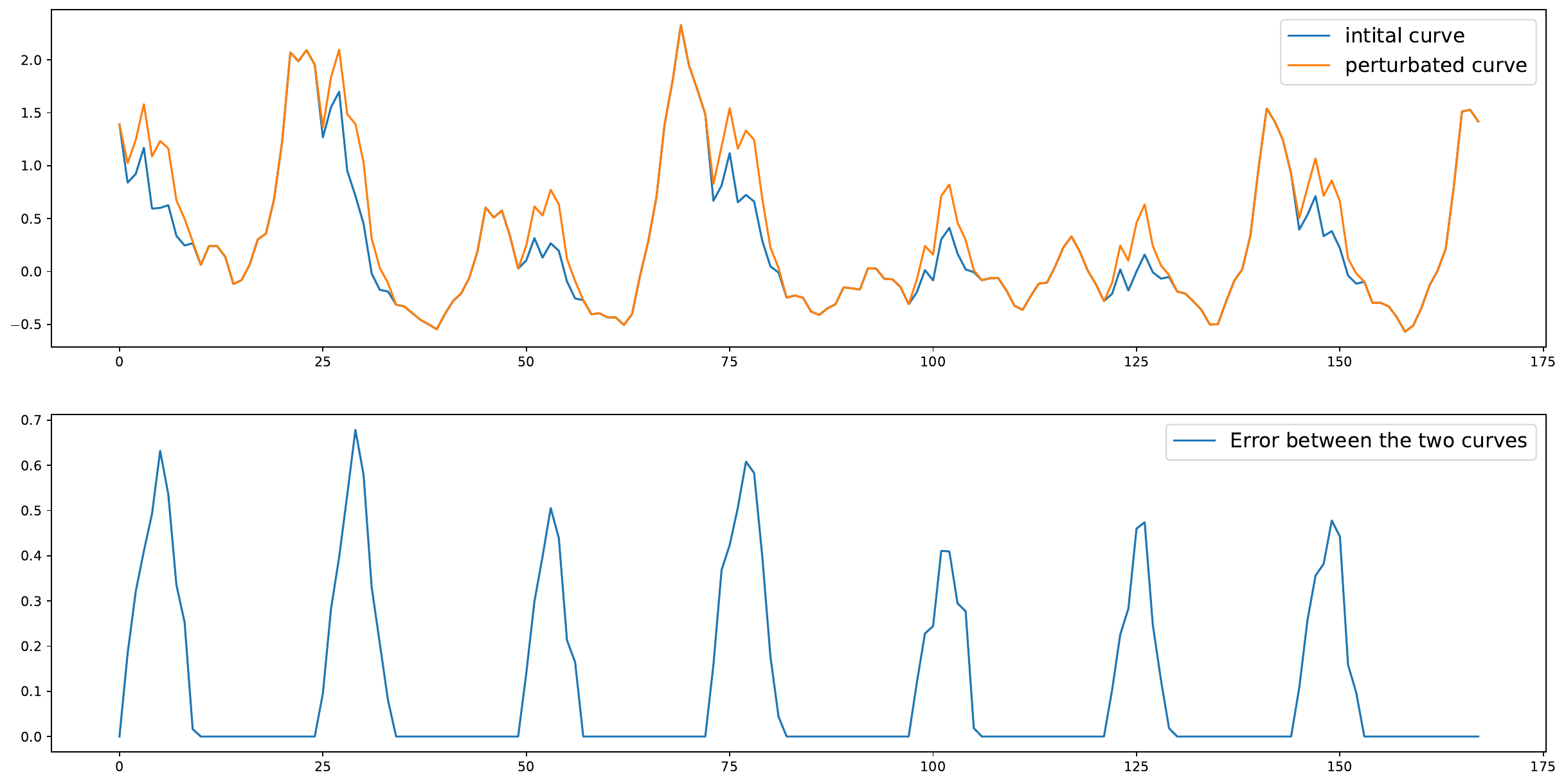}
    \caption{Effect of adding a small perturbation to the modulation shift of the third layer and 50th channel.}
    \label{fig:modulations_50}
\end{figure}

\paragraph{Setting}{In this setting, we perturbed a specific dimension of the modulation (by adding Gaussian noise) for only one particular layer of the INR for the \textit{Electricity} dataset. Then, we observe the difference in the time domain between the non-perturbated TimeFlow and the perturbated one. For instance in \cref{fig:modulations_50}, we add noise only for the third layer of the INR and the 50th channel.}

\begin{figure}[h!]
    \centering
    \includegraphics[width=0.6\linewidth]{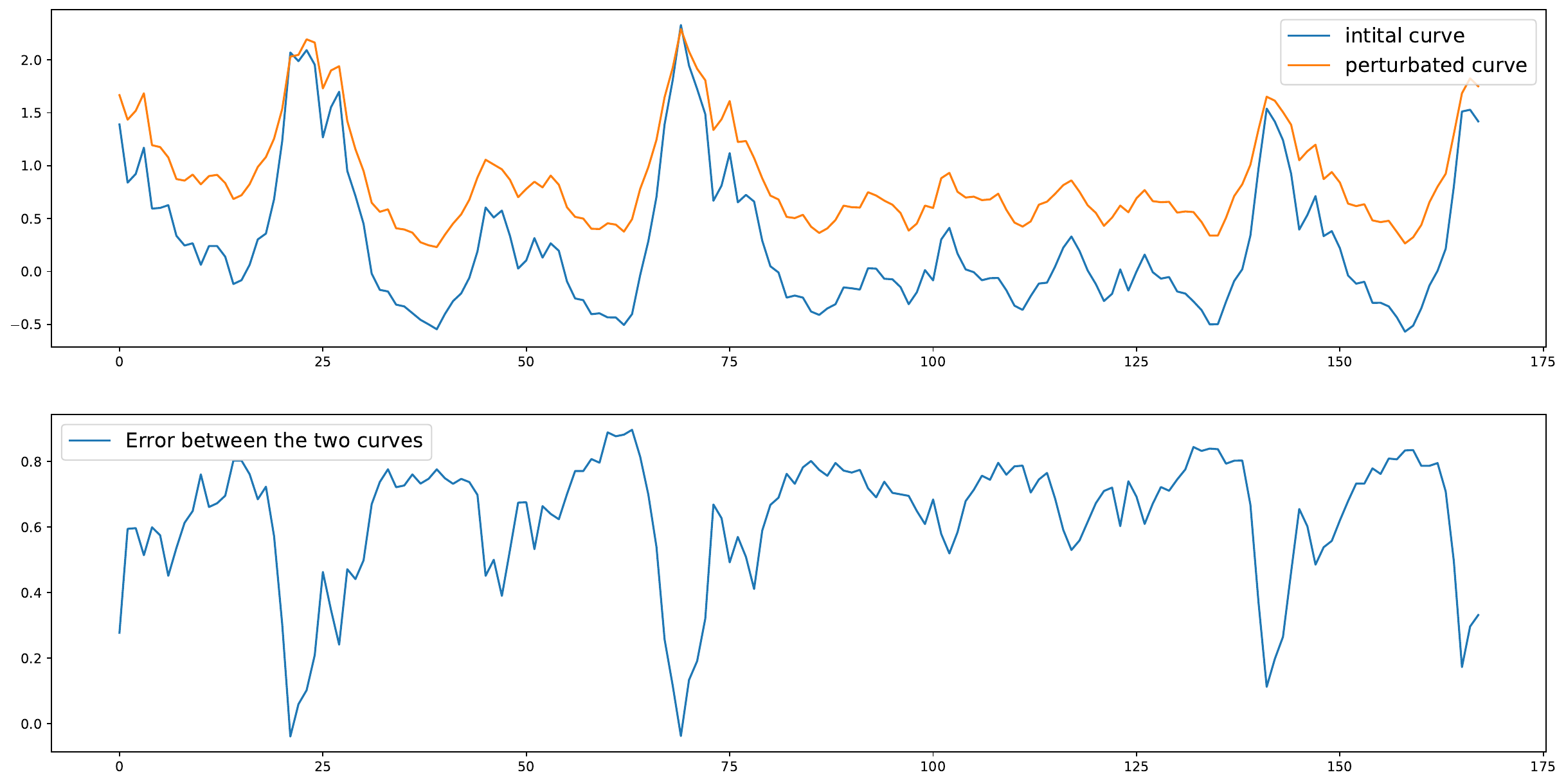}
    \caption{Effect of adding a small perturbation to the modulation shift of the third layer and 51th channel.}
    \label{fig:modulations_51}
\end{figure}

\begin{figure}[h!]
    \centering
    \includegraphics[width=0.6\linewidth]{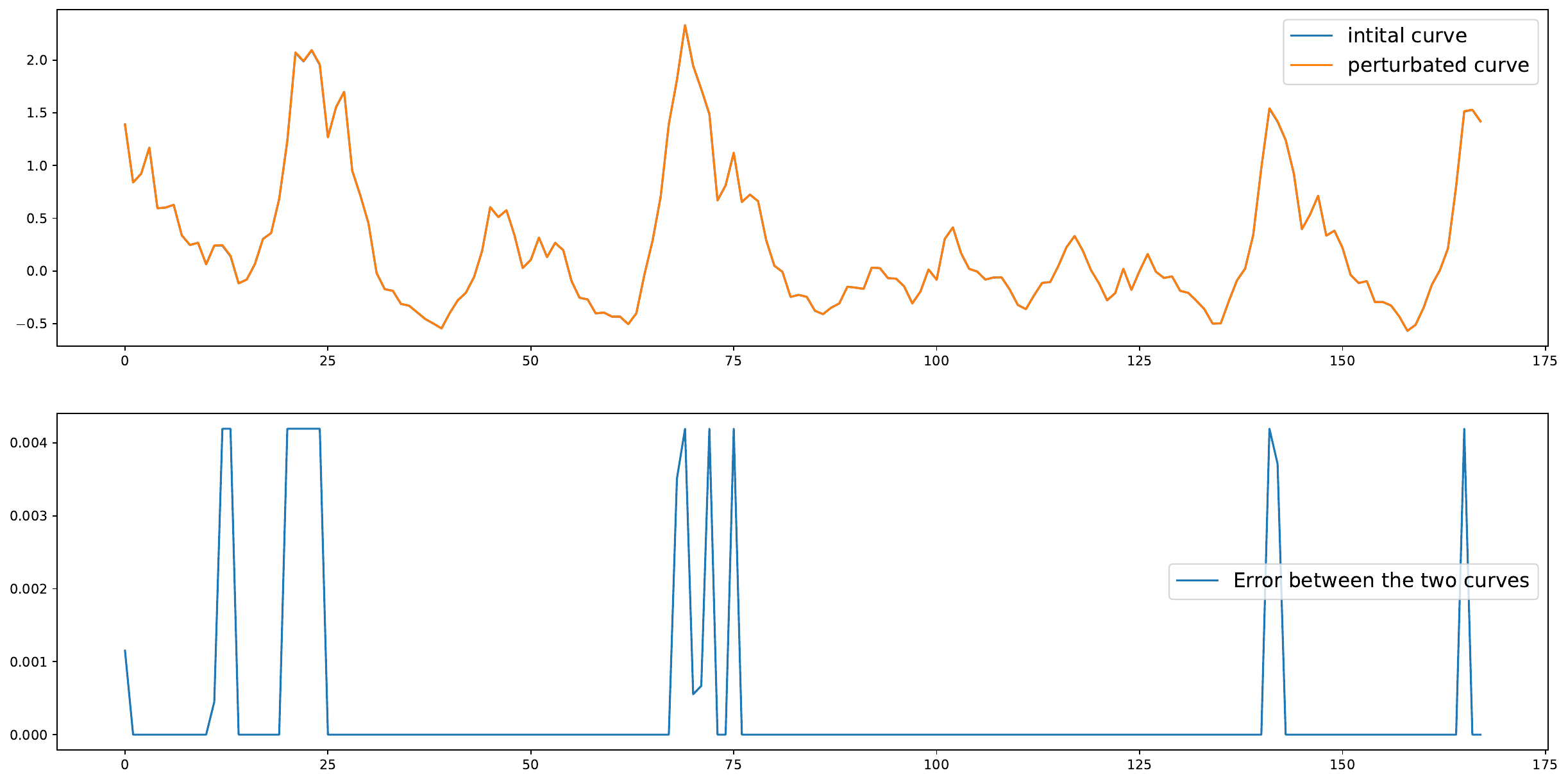}
    \caption{Effect of adding a small perturbation to the modulation shift of the fourth layer and 50th channel.}
    \label{fig:modulations_legere}
\end{figure}

\paragraph{Results.}{In \cref{fig:modulations_50}, we observed that adding a small perturbation added a smooth daily frequency pattern. In \cref{fig:modulations_51}, we observed that adding a small perturbation induces a bias that impacts the high frequencies but does not affect the low frequencies. Finally, in \cref{fig:modulations_legere}, adding a small perturbation induces a very local and slight bias (the effect is almost null). In conclusion, the impact of the small perturbation depends on the channel and the layer, but it is hard to interpret each dimension independently.}

\subsection{Visualization of two code distributions in the latent space}

Examining the behavior of the latent space at the instance level is of particular interest. It provides insights into how individual time series evolve within the latent space. However, exploring the latent space between two time series distributions is also crucial.

\begin{figure}[h!]
  \begin{subfigure}{0.47\textwidth}
    \centering
    \includegraphics[width=\linewidth]{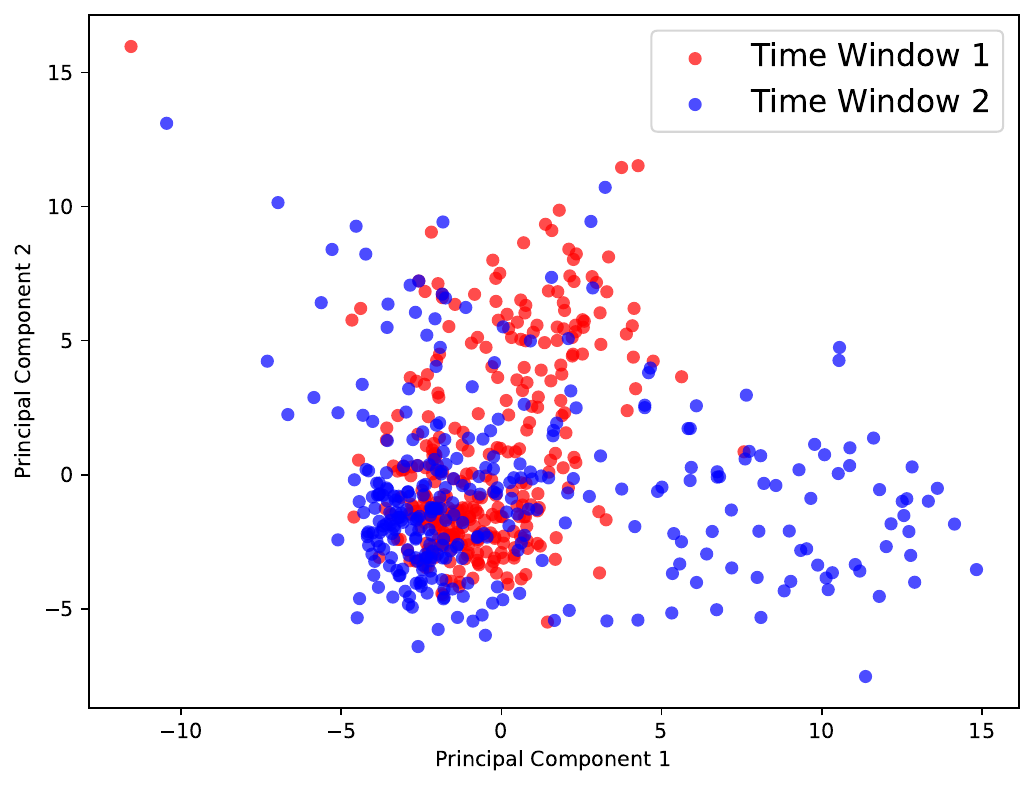}  
    \caption{The temporal shift between the two codes distribution is 3 months.}
    \label{fig:subfig1}
  \end{subfigure}%
  \begin{subfigure}{0.47\textwidth}
    \centering
    \includegraphics[width=\linewidth]{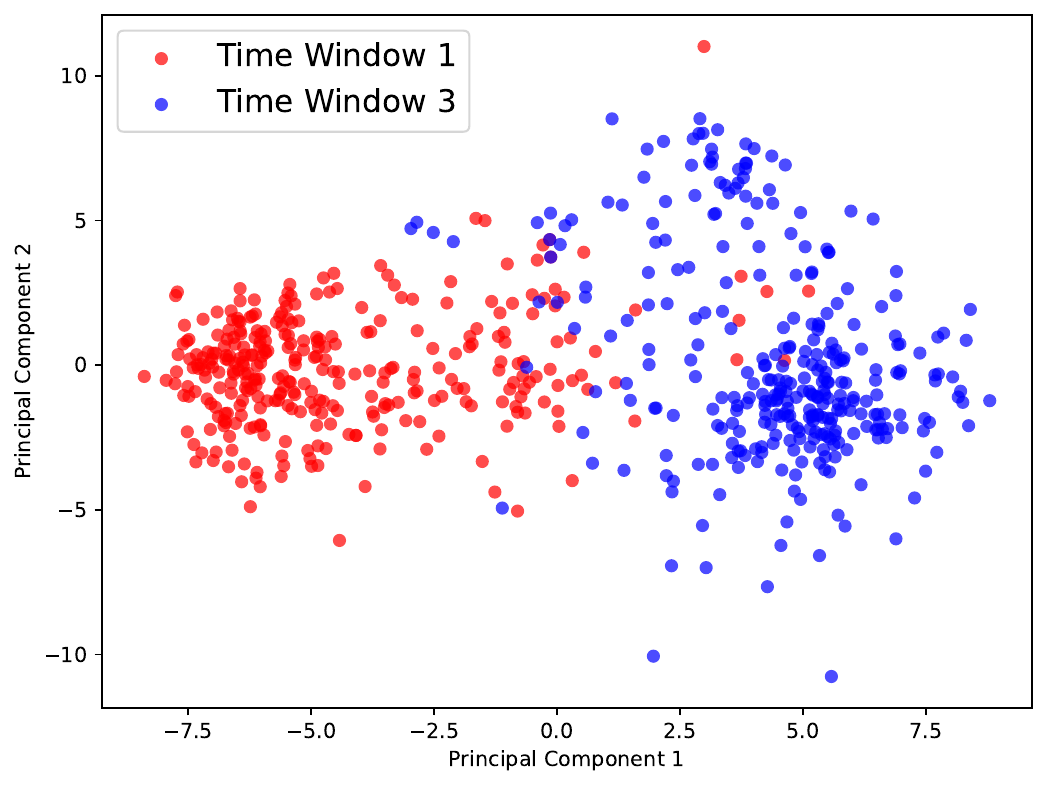}  
    \caption{The temporal shift between the two codes distribution is 6 months.}
    \label{fig:subfig2}
  \end{subfigure}
  \caption{Visualization of the two first PCA axes for two distributions of the latent codes (temporal distribution shift).}
  \label{fig:pca_analysis}
\end{figure}

\paragraph{Setting.}{We propose encoding all samples (321 samples) from the \textit{Electricity} dataset for two distinct time periods (each period is about 25 days $\approx$ 600 timestamps). This results in two distributions of latent codes, each representing different temporal support. Then, we employ Principal Component Analysis (PCA) to visualize these two latent code distributions in a two-dimensional space, as illustrated in \cref{fig:pca_analysis}. This visualization allows us to explore the structural differences, similarities, and temporal variations in the latent space representations across the specified time intervals. In \cref{fig:subfig1}, the two compared time period are separated by approximately 3 months ($\approx$ 2000 timestamps). In \cref{fig:subfig2}, the two compared time period are separated by approximately 6 months ($\approx$ 4000 timestamps).}

\paragraph{Results.}{\cref{fig:subfig1} shows that when the temporal periods are not too far from each other, the distributions of codes can largely overlap in the 2D visualization from the PCA. Conversely, as illustrated in \cref{fig:subfig2}, when the temporal periods are far from each other, the distributions of codes between the two periods become more distinct in the 2D visualization. This observation suggests that the proximity or disparity in temporal distribution shift influences the separability of latent representations in the 2D PCA space. However, this presumed separability in the latent space does not seem to significantly impact the generalization performance of TimeFlow across time, as evidenced by the results presented in \cref{tab:forecast_time_shift_electricity} and \cref{tab:forecast_new_period}. This suggests that our INR can handle relatively diverse code distributions.}

\clearpage

\section{The intuition behind the meta-learning optimization in time series forecasting}
\label{intuition_metalearning}

The concept of inner and outer loops is to be reformulated within the broader general framework of model-agnostic meta-learning \citep{finn2017model}, where the authors seek to enable rapid adaptation of the model to unseen tasks. In TimeFlow, we adapt the general idea of agnostic meta-learning to our tasks. We propose an efficient way to achieve this goal by splitting the parameters into two parts: context parameters (learned in the inner loop, responsible for the adaptive part of the model) and meta-parameters (or 'parameters shared across tasks') (learned in the outer loop, responsible for the generic part of the model).

\paragraph{In the context of time series forecasting.} TimeFlow aims to have a subset of parameters that adapts to specific contextual factors (e.g., the look-back window of a particular sample) and another subset that performs the forecasting task according to this learned context (e.g., forecasting any point within the forecast horizon as well as within the look-back window). To achieve this, we seek to adjust the codes $z^{(j)}$ exclusively on the contexts (e.g., the look-back window) for each sample $j$, while the shared parameters between samples $\theta$ and $w$ characterize a shared function capable of forecasting based on a given $z^{(j)}$. This function could be represented as $f_{\theta, w}(t, z(x^{(j)})$). This concept entails adapting  $z^{(j)}$ by sample $j$ and training $\theta$, $w$  by batch.

\clearpage

\end{document}